\documentclass[journal]{IEEEtai}

\usepackage{array}
\usepackage{cite}
\usepackage{amsmath,amssymb,amsfonts}
\usepackage{algorithmic}
\usepackage{graphicx}
\usepackage{textcomp}
\usepackage{multirow}
\usepackage[most]{tcolorbox}
\usepackage{xcolor}
\usepackage{pgf}
\usepackage{multicol}
\usepackage{forest}

{\begin{myalgo}[#1]
\centering
\begin{minipage}{7.4cm}
\begin{algorithm}[H]}%
{\end{algorithm}
\end{minipage}
\end{myalgo}}

\def\BibTeX{{\rm B\kern-.05em{\sc i\kern-.025em b}\kern-.08em
    T\kern-.1667em\lower.7ex\hbox{E}\kern-.125emX}}

\usetikzlibrary{arrows, calc, positioning}

\usepackage{amssymb}
\usepackage{pifont}
\newcommand{\cmark}{\ding{51}}%
\newcommand{\xmark}{\ding{55}}%
\usepackage{pgfplots}
\pgfplotsset{compat=1.17}
\usepackage{tikz}

\usepackage{booktabs}
\usepackage{lscape}

\usepackage{array}                
\usepackage{colortbl}             
\usepackage{xcolor}               
\usepackage{multirow}             
\usepackage{adjustbox}                 
\usepackage{float}   
\usepackage{array}
\usepackage{colortbl}
\usepackage{xcolor}
\usepackage{float}
\usepackage{multicol}
\usepackage{caption}
\usepackage{graphicx}

\usepackage[hidelinks]{hyperref}

\usetikzlibrary{arrows, calc, positioning}
\usepackage{pgfplots}
\pgfplotsset{compat=1.17}
\usepackage{tikz}
\usepackage{booktabs}
\usepackage{pifont}
\usepackage{hyperref}

\usetikzlibrary{positioning}

\usepgfplotslibrary{external}

\usepackage{tikz}
\usetikzlibrary{shapes.multipart, positioning, arrows.meta, calc}

\tikzset{
  baseSplit/.style={
    rectangle split,
    draw=blue!50!black,
    thick,
    rounded corners,
    minimum width=4cm,
    font=\normalsize,
    align=center
  },
  arrow/.style={draw, thick, ->, >=Stealth}
}

\usepackage{acronym}

\renewcommand{\IEEEiedlistdecl}{\IEEEsetlabelwidth{ABCDEFG}}

\begin{document}






\title{Teaching LLMs to Think Mathematically: A Critical Study of Decision-Making via Optimization}

\author{Mohammad~J.~Abdel-Rahman, \IEEEmembership{Senior~Member,~IEEE}, Yasmeen~Alslman, Dania~Refai, Amro~Saleh, Malik~A.~Abu~Loha, and Mohammad~Yahya~Hamed
\thanks{M.~J.~Abdel-Rahman is with the Data Science Department, Princess Sumaya University for Technology, Amman 11941, Jordan. He is also with the Electrical and Computer Engineering Department, Virginia Tech, Blacksburg, VA 24061 USA.}
\thanks{Y.~Alslman and A.~Saleh are with the Computer Science Department, Princess Sumaya University for Technology, Amman 11941, Jordan.}
\thanks{D.~Refai is with the Computer Science Department, King Fahd University of Petroleum and Minerals, Dhahran 31261, Saudi Arabia.}
\thanks{M.~A.~Abu~Loha and M.~Yahya~Hamed are with the Data Science Department, Princess Sumaya University for Technology, Amman 11941, Jordan. They have equal contributions.}
}



\maketitle

\begin{abstract}
This paper investigates the capabilities of large language models (LLMs) in formulating and solving decision-making problems using mathematical programming. We first conduct a systematic review and meta-analysis of recent literature to assess how well LLMs understand, structure, and solve optimization problems across domains. The analysis is guided by critical review questions focusing on learning approaches, dataset designs, evaluation metrics, and prompting strategies. Our systematic evidence is complemented by targeted experiments designed to evaluate the performance of state-of-the-art LLMs in automatically generating optimization models for problems in computer networks. Using a newly constructed dataset, we apply three prompting strategies: Act-as-expert, chain-of-thought, and self-consistency, and evaluate the obtained outputs based on optimality gap, token-level F1 score, and compilation accuracy. Results show promising progress in LLMs' ability to parse natural language and represent symbolic formulations, but also reveal key limitations in accuracy, scalability, and interpretability. These empirical gaps motivate several future research directions, including structured datasets, domain-specific fine-tuning, hybrid neuro-symbolic approaches, modular multi-agent architectures, and dynamic retrieval via chain-of-RAGs. This paper contributes a structured roadmap for advancing LLM capabilities in mathematical programming. 
\end{abstract}



\IEEEpeerreviewmaketitle

\def\textsc#1{\textnormal{{\sc #1}}}%

\section*{Acronyms}
\begin{acronym}
\acro{llm}[LLM]{large language model}
\acro{nlp}[NLP]{natural language processing}
\acro{gpt}[GPT]{generative pre-trained transformer}
\acro{or}[OR]{operations research}
\acro{cot}[CoT]{chain-of-thought}
\acro{gnn}[GNN]{graph neural network}
\acro{tsp}[TSP]{traveling salesman problem}
\acro{rl}[RL]{reinforcement learning}
\acro{vrp}[VRP]{vehicle routing problem}
\acro{rag}[RAG]{retrieval-augmented generation}
\acro{lp}[LP]{linear programming}
\acro{milp}[MILP]{mixed-integer \acs{lp}}
\acro{uav}[UAV]{unmanned aerial vehicle}
\acro{lpwp}[LPWP]{\acs{lp} word problems}
\acro{lora}[LoRA]{low-rank adaptation}
\acro{tot}[ToT]{tree of thoughts}
\acro{smt}[SMT]{satisfiability modulo theories}
\acro{got}[GoT]{graph of thoughts}
\acro{jss}[JSS]{job shop scheduling}
\acro{coe}[CoE]{chain-of-experts}
\acro{moe}[MoE]{mixture of experts}
\acro{cp}[CP]{constraint programming}
\acro{qp}[QP]{quadratic programming}
\end{acronym}

\renewcommand{\IEEEiedlistdecl}{\relax}

\section{Introduction}\label{sec:introduction}

\IEEEPARstart{I}{n} recent years, \acp{llm}, such as \ac{gpt} and DeepSeek, have shown remarkable advancements in the field of \ac{nlp}. These sophisticated models are designed to understand and generate human language, enabling them to perform a wide range of tasks. As a result, \acp{llm} have been applied to various applications that enhance our interactions with technology, such as text generation, question answering, summarization, and other tasks~\cite{app}.  

The remarkable capability of \acp{llm} propels researchers to harness their potential in revolutionizing various domains, including healthcare~\cite{health} and mathematical reasoning~\cite{mathres}.
By training \acp{llm} on domain-specific tasks, their potential to tolerate various challenges continues to expand, driving innovation across different industries. By using sophisticated tuning, \acp{llm} leveraged the effectiveness of many domains, such as clinical settings, educational, and research work in healthcare systems. Additionally, \acp{llm} demonstrate a surge in code generation~\cite{code} and mathematical reasoning for solving math word problems and geometry, and even providing mathematical proofs. 

\Ac{or} problems arise in various domains, including supply chain, finance, networking, and many others. 
Mathematical modeling plays a crucial role across various industries by enabling companies to analyze complex systems, predict outcomes, and optimize decision-making processes. Mathematical models help improve efficiency, reduce costs, and increase competitiveness. For example, in supply chain management, models are used to optimize inventory levels and transportation routes; in finance, they support risk assessment and investment strategies. 

Recognizing its value, many companies invest heavily in developing and implementing optimization solutions, often allocating millions of dollars annually for research, software tools, and expert consultation. Finding optimal solutions through mathematical modeling is a complex process that requires deep expertise in the specific problem domain, a strong understanding of the underlying data, solid mathematical foundations, and effective deployment skills to ensure the model can be successfully integrated and utilized within the company. Consequently, global spending on \ac{or} and analytics is estimated to exceed several billion dollars per year, highlighting the importance of mathematical modeling in driving innovation and operational industries.

Unlike the common tasks that \acp{llm} are used in, \ac{or} problems involve more than text generation and information retrieval. The objective of \acp{llm} when approaching \ac{or} problems is to correctly formulate and find the optimal solution for a given problem. This includes dealing with both textual and numerical data. Additionally, optimality, feasibility, and computational efficiency are key factors to be measured in any generated solution, which pose different challenges compared to original measures, such as relevance, accuracy, and fluency.

\acp{llm} demonstrate potential proficiency in solving \ac{or} problems, enabling \ac{or} practitioners to streamline problem formulation, enhance decision-making, and make optimization techniques more intuitive and accessible. As a result, the research into the synergy between \acp{llm} and \ac{or} has accelerated. 
This paper aims to provide an in-depth review of utilizing \acp{llm} for mathematical modeling in terms of the type of mathematical problems, their domain, the main approach followed to leverage the \acp{llm}' capability in solving these mathematical problems, and the type of \acp{llm} utilized. Furthermore, in this paper, we conduct a meta-analysis on the key factors that affect the process of using \acp{llm} for generating mathematical models. Finally, this paper complements the provided systematic evidence with targeted experiments designed to evaluate the capabilities of state-of-the-art \acp{llm} in formulating optimization models for problems in computer networks, showing promising progress in \acp{llm}' ability to parse natural language and represent symbolic formulations, but also revealing key limitations in accuracy, scalability, and interpretability. These empirical gaps motivate several future research directions.

\subsection*{Main Contributions:}

Our main contributions can be summarized as follows:
\begin{itemize}
\item Conduct a review on utilizing \acp{llm} for mathematical modeling. Exclusion criteria are applied to papers collected from five digital libraries: IEEE, Springer, ACM, ScienceDirect, and arXiv. The resulting $51$ papers are reviewed comprehensively. Furthermore, we present an overview covering all stages of utilizing \acp{llm} for generating mathematical models, from identifying the domain of the targeted problem to assessing the \ac{llm}-generated solution.
\item Perform a meta-analysis based on a set of critical review questions to gain a deep and structured understanding of current trends in using \acp{llm} for mathematical modeling. The review questions are centered around the learning approaches adopted, the types of datasets used, the evaluation metrics applied, and the prompt engineering strategies employed. 
The generated visualizations are used to identify dominant gaps in the literature and potential synergies between techniques, thereby laying a data-driven foundation for selecting the most effective methods in our experimental framework.
\item Design and conduct experiments to examine the capabilities of \acp{llm} in formulating optimization models for problems in computer networks. 
Driven by our meta-analysis, we adopt the most prevalent approaches for employing \acp{llm} (specifically, DeepSeek and \ac{gpt}-4o) in automatic mathematical modeling: (i) Constructing a new domain-specific dataset, (ii) applying three widely used prompt engineering strategies: Act-as-expert, \ac{cot}, and self-consistency, and (iii) evaluating the results using the most commonly accepted performance metrics.
\item Shed light on the future directions and challenges facing \ac{llm}-based mathematical modeling. We identify two main areas for future research and provide actionable directions. Specifically, we propose directions for (i) enhancing \ac{llm} learning and mathematical reasoning capabilities, as well as directions for (ii) developing methodologies for better understanding and diagnosing \ac{llm} limitations. 
\end{itemize}

This paper not only helps researchers contribute toward leveraging the process of automating formulation of \ac{or} problems but also provides actionable insights on how to deploy \acp{llm} in diverse business domain-specific areas, supply chains, finance, energy, communications, and others.

\subsection*{Research Questions:}

The primary objective of this survey is to identify and analyze articles that employ \acp{llm} for generating mathematical models. In light of this objective, the following research questions are formulated:
\begin{itemize}
\item What are the most commonly used \acp{llm} for generating mathematical models?
\item To what extent are \acp{llm} capable of generating mathematical models?
\item What is the best approach to be used when utilizing an \ac{llm} for formulating mathematical models?
\item What are the current challenges and limitations of utilizing \acp{llm} in mathematical modeling?
\item What are the main future directions to enhance the \acp{llm}' capabilities in mathematical modeling?
\end{itemize}

\subsection*{Paper Organization:}

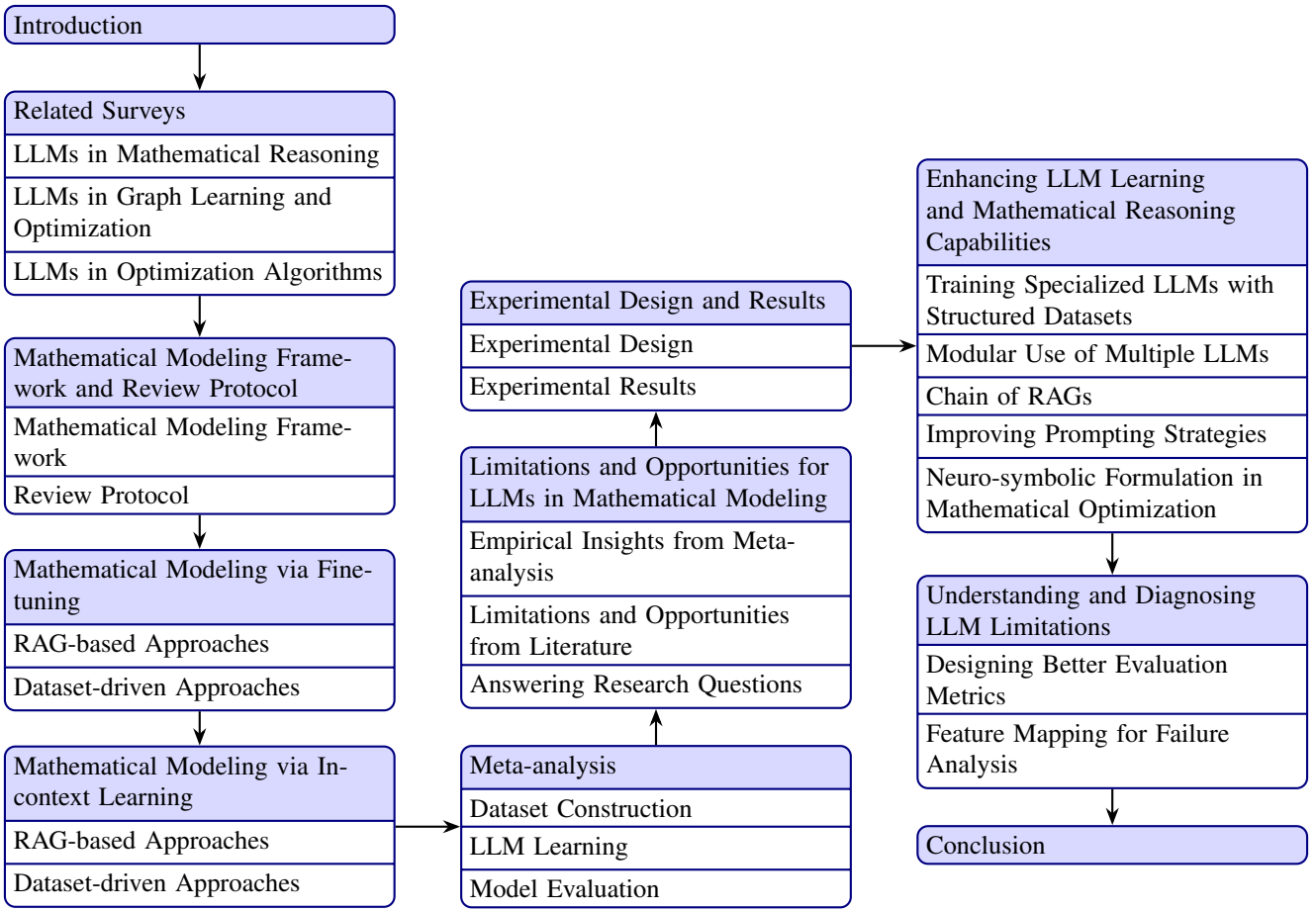
\begin{figure*}
\centering

  \noindent
  \begin{tikzpicture}
    %
    \node[baseSplit,
          rectangle split parts=1,
          rectangle split part fill={blue!15},
          text width=5cm, align=left]
      (s1) at (0,0)
      {\hyperref[sec:introduction]{Introduction}};
   %
    \node[baseSplit,
          rectangle split parts=4,
          rectangle split part fill={blue!15,white,white,white},
          text width=5cm, align=left]
      (s2) at ($(s1.south)+(0,-2cm)$)
      {\hyperref[sec:related work]{Related Surveys}
       \nodepart{two}\hyperref[subsec:LLMs in Mathematical Reasoning]{\acp{llm} in Mathematical Reasoning}
       \nodepart{three}\hyperref[subsec:LLMs in Graph Learning and Optimization]{\acp{llm} in Graph Learning and Optimization}
       \nodepart{four}\hyperref[subsec:LLMs in Algorithmic Optimization and Operations Research]{\acp{llm} in Optimization Algorithms}
      };

    %
    \node[baseSplit,
          rectangle split parts=3,
          rectangle split part fill={blue!15,white,white,white},
          text width=5cm, align=left]
      (s3) at ($(s2.south)+(0,-1.75cm)$)
      {\hyperref[sec:framework]{Mathematical Modeling Framework and Review Protocol}
       \nodepart{two}\hyperref[subsec:mathematical modeling framework]{Mathematical Modeling Framework}
       \nodepart{three}\hyperref[subsec:paper review protocol]{Review Protocol} 
      };

    %
    \node[baseSplit,
          rectangle split parts=3,
          rectangle split part fill={blue!15,white,white},
          text width=5cm, align=left]
      (s4) at ($(s3.south)+(0,-1.55cm)$)      {\hyperref[finetuning]{Mathematical Modeling via Fine-tuning}
       \nodepart{two}\hyperref[subsec:RAG finetuning]{RAG-based Approaches}
       \nodepart{three}\hyperref[subsec:dataset finetuning]{Dataset-driven Approaches}
      };

    %
    \node[baseSplit,
          rectangle split parts=3,
          rectangle split part fill={blue!15,white,white},
          text width=5cm, align=left]
      (s5) at ($(s4.south)+(0,-1.55cm)$)
      {\hyperref[incontext]{Mathematical Modeling via In-context Learning}
       \nodepart{two}\hyperref[subsec:RAG incontext]{RAG-based Approaches}
       \nodepart{three}\hyperref[subsec:dataset incontext]{Dataset-driven Approaches}
      };

    %
    \node[baseSplit,
          rectangle split parts=4,
          rectangle split part fill={blue!15,white,white,white},
          text width=5cm, align=left]
      (s6) at ($(s5.east)+(3.5cm,0)$)
      {\hyperref[meatanalysis]{Meta-analysis}
       \nodepart{two}\hyperref[subsec:dataset construction]{Dataset Construction}
       \nodepart{three}\hyperref[subsec:LLM training]{LLM Learning}
       \nodepart{four}\hyperref[subsec:model evaluation]{Model Evaluation}
      };
    %
    \node[baseSplit,
          rectangle split parts=4,
          rectangle split part fill={blue!15,white,white,white}
          ,
          text width=5cm, align=left]
      (s7) at ($(s6.north)+(0,2.25cm)$)
      {\hyperref[sec:discussion]{Limitations and Opportunities for LLMs in Mathematical Modeling}
      \nodepart{two}\hyperref[subsec:empirical insights]{Empirical Insights from Meta-analysis}
       \nodepart{three}\hyperref[subsec:limitations]{Limitations and Opportunities from Literature}
       \nodepart{four}\hyperref[subsec:RQ answers]{Answering Research Questions}
      };

    %
    \node[baseSplit,
          rectangle split parts=3,
          rectangle split part fill={blue!15,white,white},
          text width=5cm, align=left]
      (s8) at ($(s7.north)+(0,1.35cm)$)
      {\hyperref[sec:experiments_section]{Experimental Design and Results}
       \nodepart{two}\hyperref[sec:experimental_setup]{Experimental Design}
       \nodepart{three}\hyperref[sec:experiment_combined]{Experimental Results}
      };

    %
    \node[baseSplit,
          rectangle split parts=6,
          rectangle split part fill={blue!15,white,white,white,white,white},
          text width=5cm, align=left]
      (s9) at ($(s8.east)+(3.5cm,0cm)$)
      {\hyperref[sec:FutureDirection1]{Enhancing \ac{llm} Learning and Mathematical Reasoning Capabilities}
       \nodepart{two}\hyperref[sec:FutureDirection1.1]{Training Specialized \acp{llm} with Structured Datasets}
       \nodepart{three}\hyperref[sec:FutureDirection1.2]{Modular Use of Multiple \acp{llm}}
       \nodepart{four}\hyperref[sec:FutureDirection1.3]{Chain of RAGs}
       \nodepart{five}\hyperref[sec:FutureDirection1.4]{Improving Prompting Strategies}
       \nodepart{six}\hyperref[sec:FutureDirection1.5]{Neuro-symbolic Formulation in Mathematical Optimization}
      };

    %
    \node[baseSplit,
          rectangle split parts=3,
          rectangle split part fill={blue!15,white,white},
          text width=5cm, align=left]
      (s10) at ($(s9.south)+(0,-2cm)$)
      {\hyperref[sec:FutureDirection2]{Understanding and Diagnosing \ac{llm} Limitations}
       \nodepart{two}\hyperref[sec:FutureDirection2.1]{Designing Better Evaluation Metrics}
       \nodepart{three}\hyperref[sec:FutureDirection2.2]{Feature Mapping for Failure Analysis}
      };

    %
    \node[baseSplit,
          rectangle split parts=1,
          rectangle split part fill={blue!15},
          text width=5cm, align=left]
      (s11) at ($(s10.south)+(0,-0.75cm)$)
      {\hyperref[sec:conclusion]{Conclusion}
      };

    %
    \draw[arrow] (s1.south) -- (s2.north);
    \draw[arrow] (s2.south) -- (s3.north);
    \draw[arrow] (s3.south) -- (s4.north);
    \draw[arrow] (s4.south)  -- (s5.north);
    \draw[arrow] (s5.east) -- (s6.west);
    \draw[arrow] (s6.north) -- (s7.south);
    \draw[arrow] (s7.north) -- (s8.south);
    \draw[arrow] (s8.east)  -- (s9.west);
    \draw[arrow] (s9.south) -- (s10.north);
    \draw[arrow] (s10.south) -- (s11.north);
  \end{tikzpicture}

  \caption{Paper organization.}
  \label{fig:paper organization}
\end{figure*}

The rest of this paper is organized as follows: In Section~\ref{sec:related work}, we provide an overview of existing surveys on related areas. Section~\ref{sec:framework} outlines the general steps for utilizing \acp{llm} in generating mathematical models and presents our paper review protocol. Sections~\ref{finetuning} and~\ref{incontext} offer a review of the existing literature on \ac{llm}-based mathematical modeling. In Section~\ref{meatanalysis}, we conduct a meta-analysis based on critical review questions. Key limitations and opportunities for LLMs in mathematical modeling are discussed in Section~\ref{sec:discussion}. We examine the capabilities of \acp{llm} in addressing domain-specific problems in Section~\ref{sec:experiments_section}. Our future directions are elaborated in Sections~\ref{sec:FutureDirection1} and~\ref{sec:FutureDirection2}. Finally, in Section~\ref{sec:conclusion}, we present our concluding remarks for this paper. A visual representation of the structure of the paper is provided in Fig.~\ref{fig:paper organization} for more clarity.

\section{Related Surveys}\label{sec:related work}

The synergy of \acp{llm} in different domains has led to remarkable development. As a result, researchers have made notable efforts in providing several surveys that explore \ac{llm} applications in various fields.  To the best of our knowledge, no prior work has directly explored the use of \acp{llm} for solving mathematical optimization problems. However, related efforts can be found under broader categories such as mathematical reasoning and general problem-solving with \acp{llm}. 
Recently, the research on utilizing \acp{llm} in modeling mathematical optimization problems has been accelerated significantly, and there has been no dedicated study that offers a comprehensive overview of \acp{llm} utilization in formulating mathematical optimization problems. In this section, we aim to summarize surveys on mathematical reasoning and related subjects, including graph learning and optimization algorithms, to contextualize our study.

\subsection{LLMs in Mathematical Reasoning}\label{subsec:LLMs in Mathematical Reasoning}

Although mathematical reasoning and mathematical modeling are strongly interconnected, there are key differences between them. Mathematical reasoning is an abstract concept that focuses on logically analyzing and proving mathematical truths using proof techniques and algebraic manipulations. In contrast, mathematical modeling applies these mathematical concepts to represent real-world problems using equations. Our review is particularly concerned with mathematically modeling real-world, often constrained, decision-making problems, aiming to achieve one or multiple objectives, \textit{a.k.a.} mathematical programming. 

Several surveys have focused on the role of \acp{llm} in mathematical reasoning, showing their capabilities to solve arithmetic problems and prove theorems. Ahn~\textit{et al.}~\cite{ahn2024largelanguagemodelsmathematical} categorized mathematical problems faced by \acp{llm} into arithmetic, math word, geometry problems, and automated theorem proving. Additionally, the authors discussed the challenges of formal proof generation, logical reasoning, and handling complex symbolic representations. Liu \textit{et al.}~\cite{liu2025mathematicallanguagemodelssurvey} provided a broader taxonomy of mathematical \acp{llm}, including instruction-based learning, tool-assisted problem solving, and \ac{cot} reasoning approaches. Their survey highlighted over $60$ benchmark datasets used for evaluating \acp{llm} in mathematical tasks.

\subsection{LLMs in Graph Learning and Optimization}\label{subsec:LLMs in Graph Learning and Optimization}

Beyond symbolic mathematics, researchers have explored how \acp{llm} interact with graph-based problems, including graph learning and graph optimization. Graph learning is a field that aims to gain useful information and patterns using machine learning (ML) and other techniques from data that is represented as a graph. Node classification, edge prediction, and graph clustering are considered the main common tasks in graph learning. In graph optimization, the main objective is to find the best solution to problems modeled as graphs. 

Chen \textit{et al.}~\cite{chen2024exploringpotentiallargelanguage} focused on the synergy between \acp{llm} and \acp{gnn}; they explored the use of an \ac{llm} as an encoder and as a predictor for the node classification tasks. When utilizing \acp{llm} as an encoder, they are used to pre-process the text attributes, and then \acp{gnn} are trained on the enhanced attributes as predictors. However, they overlooked the importance of using \acp{llm} to solve pure graph problems, which refer to classical graph-theoretic tasks such as the \ac{tsp}, shortest path problem, and minimum cut problem, where the objective is to directly reason over graph structures without relying on external textual attributes or paired information. These problems are typically formulated as combinatorial optimization problems and often require rigorous mathematical modeling to identify optimal or near-optimal solutions. Ignoring this dimension limits the exploration of how \acp{llm} could contribute not only to graph representation and reasoning but also to bridging optimization techniques with mathematical formulations of complex graph problems.

Ren \textit{et al.}~\cite{Ren_2024} provided a comprehensive review of \acp{llm} in graph learning tasks, discussing four main aspects: \acp{gnn} as a prefix, \acp{llm} as a perfix, fusion models, and applying \acp{llm} directly on graph tasks. Using \acp{gnn} as a prefix refers to applying \ac{gnn} as a preliminary step for processing a specific graph, after which the structural tokens are input into \acp{llm}. Similar to~\cite{chen2024exploringpotentiallargelanguage}, the authors discussed \acp{llm} as a prefix for providing initial embeddings or labels for \acp{gnn}, which is the same as using \acp{llm} as encoders.
They also covered fusion models where \acp{llm} are combined with \acp{gnn}, as well as the direct use of \acp{llm} in tasks like node classification and spatio-temporal graph representation. Once again, the authors did not adequately cover the \ac{llm}'s capability in solving pure graph problems, which require mathematical modeling and optimization rather than representation learning alone.

Jin~\textit{et al.}~\cite{jin2024largelanguagemodelsgraphs} categorized graph scenarios where \acp{llm} are applicable into pure, text-attributed, and text-paired graphs. In a text-attributed graph, each node or edge is associated with textual attributes or labels, while the primary structure remains the graph itself. When graphs are combined with external textual data, such as documents, sentences, or paragraphs, this is referred to as text-paired graphs. Moreover, they classified the proposed three techniques for applying \acp{llm}: \ac{llm} as a predictor, \ac{llm} as an encoder, and \ac{llm} as an aligner. Although they offered a comprehensive taxonomy of pure graphs, which can be addressed as an optimization problem through mathematical modeling. They inadequately explore graph representations, such as edge verbalization and adjacency lists.

\subsection{LLMs in Optimization Algorithms}\label{subsec:LLMs in Algorithmic Optimization and Operations Research}

Ma~\textit{et al.}~\cite{ma2024automatedalgorithmdesignsurvey} provided a comprehensive review of meta-black-box optimization (MetaBBO), categorizing the role of \acp{llm} in algorithm selection, configuration, and automated problem formulation. They reviewed \ac{rl}, supervised learning, and in-context learning techniques to optimize computational efficiency and solve complex \ac{or} problems. These contributions indicate the growing role of \acp{llm} in automating and optimizing mathematical and decision-making problems, a core aspect of \ac{or}.

Tao \textit{et al.}~\cite{tao2025combinatorial} presented how deep learning (DL) and artificial intelligence (AI) in general have been deployed for combinatorial optimization. It traces the evolution from traditional methods, such as branch-and-bound and simulated annealing, which struggle with large-scale problems, to modern DL-based approaches that now rival or even outperform professional solvers in specific scenarios. The paper categorizes DL methods based on problem types, e.g., \ac{tsp}, \ac{vrp}, model architecture, e.g., \acp{gnn}, transformers, and training strategies, e.g., supervised, reinforcement, and unsupervised learning, highlighting recent innovations, such as chaotic backpropagation and prompt-based optimization using \acp{llm}.  

Table~\ref{tab:survey_comparison} presents a comparison of related surveys on \ac{llm} capabilities in mathematical modeling, focusing on their methodological approach, scope of applications, and the roles assigned to \acp{llm}. The last two columns highlight the key distinctions between their findings and the contributions of our work.

\begin{table*}
\centering
\renewcommand{\arraystretch}{1.3}
\caption{Overview of surveys that explore \ac{llm}
applications in various fields. To the best of our knowledge, our paper is the first to survey the synergy between \acp{llm} and mathemtical optimization.}
\label{tab:survey_comparison}
\small{
\begin{tabular}{|>{\centering\arraybackslash}m{1cm}
                ||>{\centering\arraybackslash}m{2.4cm}
                |>{\centering\arraybackslash}m{2.3cm}
                |>{\centering\arraybackslash}m{2cm}
                |>{\centering\arraybackslash}m{2cm}
                |>{\centering\arraybackslash}m{1.9cm}
                |
                }
\hline
\textbf{Survey} & \textbf{Focus area} & \textbf{Methodology used} & \textbf{Scope of application} & \textbf{Use of \acp{llm}} & \textbf{Relation to optimization} 
\\
\hline \hline
Ahn \textit{et al.}~\cite{ahn2024largelanguagemodelsmathematical} & Mathematical reasoning & Performance analysis & Theorem proving, problem-solving & Reasoning & Not discussed 
\\
\hline
Liu \textit{et al.}~\cite{liu2025mathematicallanguagemodelssurvey}  & Mathematical reasoning & Categorization & Datasets, problem-solving & Mathematical calculation and mathematical reasoning & Not discussed 
\\
\hline
Ren \textit{et al.}~\cite{Ren_2024} & Graph learning tasks & Literature review and novel taxonomy & How \acp{llm} are applied in graph-based learning & Integration with \acp{gnn} & Not discussed 
\\
\hline
Jin \textit{et al.}~\cite{jin2024largelanguagemodelsgraphs} & Graph-based \ac{llm} applications (\ac{llm} as predictor, \ac{llm} as encoder, and \ac{llm} as aligner) & Taxonomy creation and systematic review of scenarios and techniques of utilizing \acp{llm} with graphs & Graph-structured data & Classification & Not discussed
\\
\hline
Chen \textit{et al.}~\cite{chen2024exploringpotentiallargelanguage} & \acp{llm} in graph learning & Empirical study & Graph-based prediction & Attribute enhancement & Not discussed 
\\
\hline
Ma \textit{et al.}~\cite{ma2024automatedalgorithmdesignsurvey}  & Automated algorithm design & Meta-learning review & Evolutionary algorithms & Meta-learning for optimization & Mentioned but not primary 
\\
\hline
Peng \textit{et al.}~\cite{tao2025combinatorial} & Combinatorial optimization & Literature review & Combinatorial algorithms & Reasoning & Mentioned
\\
\hline
\end{tabular}
}
\end{table*}

\section{Mathematical Modeling Framework and Review Protocol}\label{sec:framework}

In this section, we first discuss the \ac{llm}-driven mathematical modeling framework, outlining the key steps involved in leveraging the capabilities of \acp{llm} for automating mathematical modeling. Then, we present our adopted protocol for collecting and reviewing papers related to mathematical modeling using \acp{llm}.

\subsection{Mathematical Modeling Framework}\label{subsec:mathematical modeling framework}

The first step in using \acp{llm} for solving mathematical problems is to establish the scope of the knowledge and the type of data that will be collected by determining the specific domain of interest (e.g., networking, healthcare, telecommunication, economics, etc.). The second step involves establishing a theoretical framework to address the domain-specific problem, enabling the capture of relationships between entities, concepts, or events. At this stage, the problem may be formulated using graph theory, game theory, or mathematical programming. Considerable research has explored the use of \acp{llm} for solving graph-based problems, as discussed in~\cite{jin2024largelanguagemodelsgraphs, chen2024exploringpotentiallargelanguage}. Game-based problems have received comparatively less attention, with only a few studies such as~\cite{game1,mao2023alympics,duan2024gtbench}. In this paper, we focus on the mathematical programming framework. 

\begin{figure*}
\centering
\includegraphics[width=0.9\linewidth]{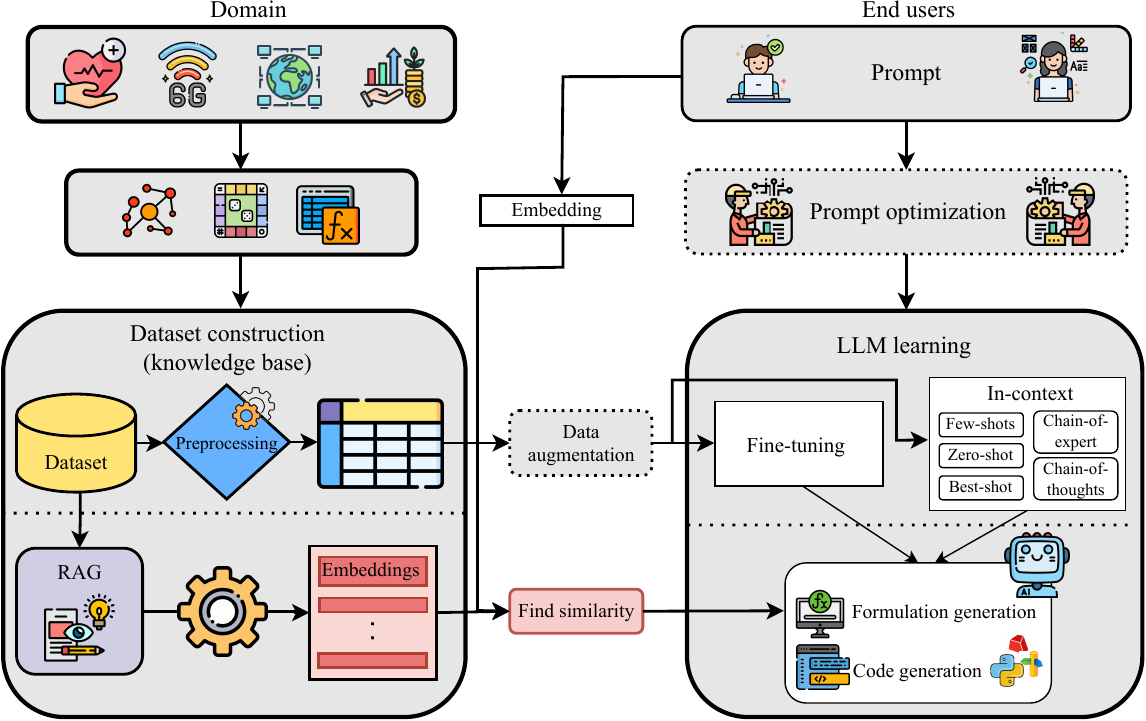}
\caption{General framework steps. The process begins by determining the main domain, followed by establishing a theoretical framework for problem representation. A dataset or knowledge base is then created to support \ac{llm} learning through in-context learning or fine-tuning. Finally, the \ac{llm} generates a mathematical formulation, possibly including solver code, using either a standard or optimized prompt.}
\label{fig:steps}
\end{figure*}

The third step is dataset construction, and it is a crucial step for building a knowledge base that can be used in \ac{llm} training. There are two main approaches for using the data: Either by constructing a typical dataset that requires preprocessing techniques or by using the \ac{rag} approach. To ensure that the collected data is ready to be used when considering a typical dataset, data preprocessing is considered to clean the data. On the other hand, when using \ac{rag}, the data is broken into chunks, making it ready for use by downstream tasks (calculating the embeddings). Data augmentation is an optional step that can be used to increase the size of the knowledge base when considering the use of a typical dataset. Data augmentation can be considered depending on the performance of \acp{llm}. The data is then fed to the \ac{llm} for fine-tuning or in-context learning. After training the \ac{llm}, the end user can interact with the \ac{llm} via prompts, and prompt optimization techniques are employed to optimize responses. The \ac{llm}-driven mathematical modeling framework is depicted in Fig.~\ref{fig:steps}.

\subsection{Review Protocol}\label{subsec:paper review protocol}

In this subsection, we present (i) the search queries and strings used to collect papers related to mathematical modeling using \acp{llm}, (ii) the paper exclusion criteria applied to exclude irrelevant papers, and (iii) the review questions utilized to comprehensively review the selected papers. We defined review questions related to each step of the \ac{llm}-driven mathematical modeling framework.

\subsubsection{Search Queries and Strings}

The keywords used for the initial search can be divided into two main groups: The first is related to the \acp{llm} (large language model, generative AI, natural language processing), whereas the second one is related to mathematical modeling (mathematical modeling, operations research, mathematical programming, linear programming, optimization modeling, mathematical optimization, auto formulation). 


The search queries shown in Table~\ref{tab:strings} were used to retrieve all $61$ papers from five digital libraries, IEEE, Springer, ACM, ScienceDirect, and arXiv. 
\begin{table}
\centering
\renewcommand{\arraystretch}{1.2}
\caption{Search strings.}
\label{tab:strings}
\begin{tabular}{|>{\centering\arraybackslash}m{0.1\columnwidth}
                ||>{\centering\arraybackslash}m{0.78\columnwidth}|}
\hline
S1 & ((large language model) OR (natural language processing)) AND ((mathematical modeling) OR (operations research) OR (mathematical programming) OR (optimization modeling)) \\
\hline
S2 & ((\ac{llm}) OR (large language model) OR (natural language processing)) AND ((mathematical optimization) OR (mathematical modeling) OR (operations research) OR (mathematical programming) OR (linear programming) OR (optimization modeling)) \\
\hline
S3 & ((\ac{llm}) OR (large language model) OR (generative AI) OR (natural language processing)) AND ((mathematical optimization) OR (mathematical modeling) OR (operations research) OR (mathematical programming) OR (linear programming) OR (optimization modeling)) \\
\hline
\end{tabular}
\end{table}

\subsubsection{Paper Exclusion Criteria}

The following exclusion criteria were then applied to exclude irrelevant papers:
\begin{itemize}
\item Papers that did not use \acp{llm} specifically for modeling mathematical optimization problems: Our main objective is to provide an in-depth review of papers that utilize \acp{llm} to tackle mathematical optimization problems.
\item Survey or review papers: Our objective is to
analyze primary research contributions rather than resummarizing existing reviews.
\item Short papers, letters, or comment papers: These formats often lack detailed methodological descriptions and thorough experimental evaluations, making them unsuitable for in-depth review. 
\end{itemize}

The following subsections outline sets of review questions aligned with the steps described in the previous subsection and depicted in Fig.~\ref{fig:steps}.

\subsubsection{Domain Determination and Problem Representation Review Questions}

As the first step is to determine the domain and the representation of the problem, the following review questions were used when reviewing the final set of papers.
\begin{enumerate}
    \item[\textbf{RQ1:}] What is the primary domain of the problem?
    \item[\textbf{A1:}]This question is essential to study the primary domain studied across the literature. The answer to this question allows for classifying the problem into one of the following domains: (i) \Ac{lp} problems, (ii) combinatorial problems, which include integer \ac{lp} (ILP) and \ac{milp} problems, and (iii) a combination of both linear and combinatorial problems. Identifying the domain provides clarity on the mathematical structures, methodologies, and algorithms most suitable for addressing the problem at hand.
    \item[\textbf{RQ2:}] What type of optimization problems is being addressed?
    \item[\textbf{A2:}] Identifying the type of optimization problem that is being addressed can help in determining the complexity of the problem. Additionally, the problems differ in their structure, constraints, and objectives, which influence the choice of solution techniques. The possible optimization problems include general optimization problems, wireless optimization problems, scheduling problems, resource allocation problems, \ac{uav} optimization problems, data center networking, and business optimization problems. 
\end{enumerate}

\subsubsection{Dataset Construction Review Questions}

This subsection presents the review questions related to the process of preparing the data for use by the \acp{llm}. 
\begin{enumerate}
    \item[\textbf{RQ3:}] What is the name of the used dataset?
    \item[\textbf{A3:}] The used datasets can directly affect the capability of the \acp{llm} to solve complex optimization problems. Many datasets can be used by \acp{llm}, including NL4Opt~\cite{nl4opt}, 
    \ac{lpwp}, Mamo~\cite{mamo}, 
    IndustryOR~\cite{tang2024orlm}, NLP4LP~\cite{ahmaditeshnizi2024optimus}, ComplexOR~\cite{complexORdataset}, 
    OptiBench~\cite{huang2024optibench}, 
    StarJob~\cite{anonymous2024starjob}, ReSocratic~\cite{huang2024optibench}, GSM8K~\cite{GSM8k}, Big-Bench Hard (BBH)~\cite{BBH}, 
    StructuredOR~\cite{wang2024bpp}, 
    Cycle Share~\cite{jiao2024city},
    DiDi operational~\cite{didi}, 
    Pyomo Cookbook~\cite{pyomoCookbook}, 
    AQUA~\cite{AQuA}, 
    Ner4Opt~\cite{kadiouglu2024ner4opt},
    Scientific variable extraction benchmark~\cite{liu2024variable},
    and MultiArith~\cite{MultiArith}. In some studies, the authors developed their own datasets.

    \item[\textbf{RQ4:}] What is the size of the used dataset?
    \item[\textbf{A4:}] The size of the dataset also has a direct impact on the \acp{llm} learning process. The answer to this question outlines the size used.
\end{enumerate}

\subsubsection{\ac{llm} and Solver Determination Review Questions}

This subsection presents the review questions related to \acp{llm} learning.
\begin{enumerate}
    \item[\textbf{RQ5:}] What is the name of the used \ac{llm}?
    \item[\textbf{A5:}] In recent years, there have been many \acp{llm} that can be used for different tasks, each with its benefits. The type of \acp{llm} can be determined by one of the following:  \ac{gpt}-2, \ac{gpt}-3, \ac{gpt}-3.5, \ac{gpt}-4, Mistral, Llama, DeepSeek, Qwen, PaLM, Zephyr, Phi, CodeRL, Mixtral, Falcon, and Claude. 

    \item[\textbf{RQ6:}] What types of solvers and modeling languages are used in the study?
    \item[\textbf{A6:}]In the literature, \acp{llm} are utilized not only for generating the mathematical formulations of specific problems but also for generating the corresponding code. This code is tailored for specific solvers, meaning the types of solvers adopted by the \acp{llm}, which include CPLEX~\cite{cplex_manual126}, Gurobi~\cite{gurobi_manual12}, Google's \ac{or}-Tools~\cite{or-tools_manual}, \ac{smt}~\cite{z3_manual}, OptVerse~\cite{optverse_manual}, COPT~\cite{copt_manual}, and SCIP~\cite{scip_manual}. In certain situations, the solver may not be specified. These solvers were adopted using different modeling languages, including MiniZinc, MAPL code, Pyomo, CPMpy, PuLP, Zimpl, and AMPL. 
    
    \item[\textbf{RQ7:}] Which approach does the study use to adapt \acp{llm}?
    \item[\textbf{A7:}] Adapting an \ac{llm} to a specific task or domain is a critical step in leveraging its capabilities effectively. The study may employ one of the following approaches:
    Using a dataset for in-context or fine-tuning, or using the \ac{rag} approach. Each approach has its trade-offs in terms of cost, complexity, and adaptability.
\end{enumerate}

\subsubsection{\ac{llm} Learning Review Questions}

This subsection outlines the review questions concerning the chosen learning paradigm for the \acp{llm}.
\begin{enumerate}
    \item[\textbf{RQ8:}] What are the approaches used to adapt \acp{llm} for new tasks?
    \item[\textbf{A8:}] Generally, there are two main approaches: Fine-tuning and in-context learning. 
\end{enumerate}

\subsubsection{Model Evaluation Review Questions}

This subsection presents the review questions related to the evaluation process of the models generated by \acp{llm}.  
\begin{enumerate}

\item[\textbf{RQ9:}] What evaluation metrics are used to assess the quality of \ac{llm}-generated optimization formulations?  

\item[\textbf{A9:}] The evaluation of \ac{llm}-generated formulations can be structured into several categories.  Firstly, the category of solution quality evaluates how far the obtained solution is from the best-known or optimal value. This is measured through \textbf{solution accuracy}, \textbf{optimality gap}, and \textbf{relative regret}, which all quantify the deviation between the obtained solution and the optimum, as well as the \textbf{average improvement ratio (AIR)}, among each benchmark it compares how far the LLM-generated formulation is from the best-known optimum relative to a human-designed heuristic; values below $1$ mean the generated formulation is closer to the optimum on average, around $1$ means similar performance, and above $1$ means worse. 
The second category, surface-form accuracy, examines how well the generated text matches the reference formulation. This is assessed at the token level using \textbf{precision} (the fraction of generated tokens that are correct), \textbf{recall} (the fraction of required tokens that appear in the output), and their harmonic mean, the \textbf{F1-score}. \\

The third category, buildability and runtime robustness, examines whether the generated code can run reliably in practice. Buildability is assessed through \textbf{compilation accuracy}, which checks whether the code parses correctly and is accepted by the solver without syntax or schema errors. Runtime robustness is then measured by the \textbf{execution rate}, defined as the proportion of compilable runs that complete successfully without runtime errors or crashes. 
A related category, \textbf{feasibility or model soundness}, goes one step further by verifying whether compiled runs produce solver-feasible outputs, quantified by the \textbf{feasibility pass rate}. \\

Efficiency and search effectiveness form another category, which captures different aspects of solver performance. Efficiency reflects how quickly the solver delivers high-quality results. It is measured by the \textbf{average solving time} or \textbf{running time}, where shorter times indicate better performance, and by \textbf{convergence performance}, which evaluates how rapidly solution quality improves, for example, the time required to reach a predefined quality level such as achieving an objective value within 1\% of the best-known solution. Search effectiveness, in contrast, measures the system’s ability to identify valid solutions. A common metric is \textbf{Valid@k within time $t$}, which checks whether at least one of the top-$k$ generated solutions is valid within a specified time limit. \\

Additional metrics capture broader aspects of quality. Domain utility outcomes measure real-world effectiveness using \textbf{utility improvement metric} in the problem’s native units (e.g., Mbps, dollars, or minutes), compared against a baseline or the best-known result. For multi-objective problems, \textbf{hypervolume (HV)} quantifies the extent of objective space covered by the produced Pareto set, whereas \textbf{inverted generational distance (IGD)} measures the closeness of the produced set to the reference front. \\  

Another category is \emph{mathematical fidelity}, which checks whether the LLM-generated formulation expresses the same mathematics as the original formulation. We use two measures. \textbf{Integrity gap (structure):} which shows how different the building blocks and their connections are compared with the original counts and types of variables and constraints, and which variables appear in which constraints. Report it so that $0$ means identical structure and larger values mean a worse match. \textbf{Semantic similarity} Evaluates whether the two formulations express the same meaning even if written differently; in practice, this can be computed via (i) embedding-based similarity between normalized formulations or (ii) an LLM-judge that decides whether predicted elements semantically match the original and then summarizes with precision, recall, and F1.
 \\

Finally, human-centered evaluation complements these quantitative metrics with \textbf{expert qualitative assessments}, focusing on aspects such as clarity, maintainability, and practical usefulness. 
\end{enumerate}

    

In the following two sections, we comprehensively review the fine-tuning and in-context learning based studies, respectively. 
    
\section{Mathematical Modeling via Fine-tuning}\label{finetuning}

Mathematical programming is essential in solving complex optimization problems across various domains. As mentioned earlier, \acp{llm} demonstrate exceptional capabilities across various domains due to their reasoning abilities. While their general-purpose capabilities are powerful, adapting \acp{llm} to the complexities of mathematical programming often requires refinement. In this context, fine-tuning can be used to refine pre-trained models such as \acp{llm} to specialize in mathematical programming tasks by learning from domain-specific data. Fine-tuning takes the model's previous knowledge as a starting point and refines its performance in specific tasks such as solving \ac{lp}, integer programming (IP), or even \ac{milp} problems. Fine-tuning can be very expensive in terms of computational resources, making it a less preferable approach. However, researchers have developed strategies to reduce the computational cost of fine-tuning by using lightweight adapters such as \acp{lora} and similar techniques. These methods allow models to adapt efficiently without retraining all parameters, thereby significantly lowering resource requirements.

\begin{figure*}
\centering
\scalebox{0.74}{
\begin{forest}
        for tree={
            draw, rounded corners, align=center, edge={->, thick}, text width=2.3cm, l sep+=1pt
        }
        [\ac{llm}-based optimization \\ modeling approaches, fill=cyan, text width=3.5cm
        [Optimization modeling via\\fine-tuning, fill=cyan, text width=4cm
               [\ac{rag}-based, text centered, fill=lightgray
                    [Combinatorial\\ \cite{liu2024generative}, text centered, fill=pink]
                ]
                [Data-driven, text centered, fill=lightgray
                    [Combinatorial\\ \cite{abgaryan2024llms,  huang2024optibench,ahmed2024lm4opt}
                    \\ \cite{anonymous2024starjob, zhou2024llmsolver,li2023synthesizing}\\
                    \cite{yu2024deep, wang2024leveraging, wang2024bpp}, text centered, fill=pink]
                    [Combinatorial \\ and linear \\ \cite{tang2024orlm, zhang2024solving,jiang2024llmopt}\\
                    \cite{chen2025solver}, text centered, fill=pink]
                    [Linear\\ \cite{amarasinghe2023aicopilotbusinessoptimisationframework, kadiouglu2024ner4opt}, text centered, fill=pink]
                ]
            ]
        [Optimization modeling via\\ in-context learning, fill=cyan, text width=4cm
        [\ac{rag}-based, text centered, fill=lightgray
                    [Combinatorial\\ \cite{anonymous2024droc, ahmaditeshnizi2024optimus, zhang2024generative}, text centered, fill=pink]
                ]
        [Data-driven, text centered, fill=lightgray
                    [Combinatorial\\ \cite{liu2024variable, ahmaditeshnizi2024optimus} \\
                    \cite{chen2024diagnosing, yao2024multi, wasserkrug2024large,yang2024large,hao2024planning,huang2025llms,wang2025ormind, nammouchi2024towards, luzzi2025chatgpt, dui2025generative, gemp2024steering}\\
                    \cite{huang2024optibench,sun2024generative,xiao2023chain}\\
                    \cite{wang2024large, li2024towards,astorga2024autoformulation, jiao2024city}, text centered, fill=pink]
                    [Combinatorial\\ and linear\\ \cite{li2023large, ahmaditeshnizi2024optimus, tang2024orlm}\\ \cite{ zhang2024solving, mostajabdaveh2024optimization, li2024nl2or, sidhu2024evaluation}, text centered, fill=pink]
                    [Linear\\ \cite{deng24cafa, zhang2025or, li2025abstract, kadiouglu2024ner4opt, zhang2025decision}, text centered, fill=pink]
         ]
 ]           
]
\end{forest}
}
\caption{Taxonomy of \ac{llm}-based optimization modeling approaches.}
\label{fig:incontexttax}
\end{figure*}
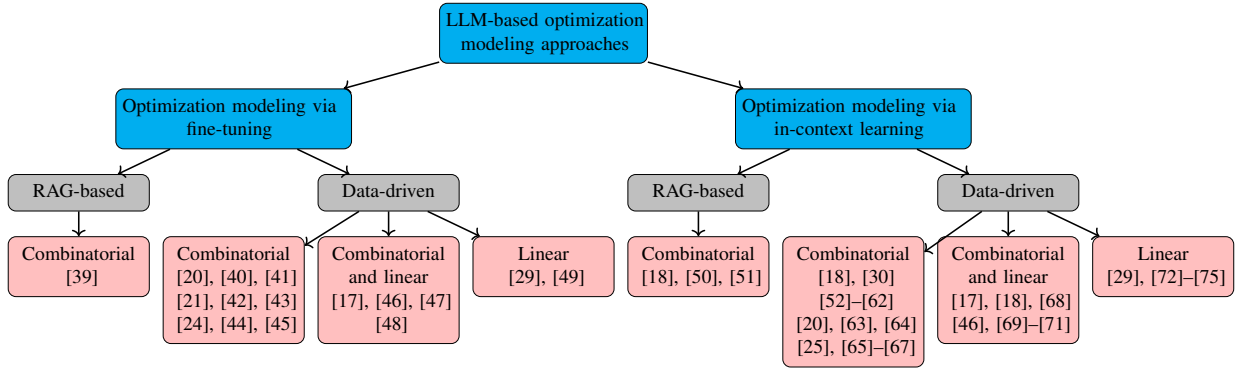

This section explores the role of fine-tuning in transforming \acp{llm} into tools for mathematical programming. It highlights how fine-tuning enhances the ability of \acp{llm} to solve optimization problems effectively. Across the literature, researchers have identified the need for a comprehensive knowledge base to fine-tune pre-trained models effectively. This knowledge base, tailored to the specific domain in which the model is intended to specialize, can be constructed through two primary approaches: \ac{rag} techniques or utilizing a supervised dataset. As a result, we categorized the literature based on how they build their knowledge base. \Ac{lp}, IP, and \ac{milp} differ significantly in terms of the computational complexity and the effort required for any model to solve them, primarily due to the nature of their constraints and variables. Thus, we further classify the literature based on the type of optimization problem that has been addressed, as shown in Fig.~\ref{fig:incontexttax}.

\subsection{RAG-based Approaches}\label{subsec:RAG finetuning}

\Ac{rag} serves as a powerful mechanism to enhance \ac{llm}'s ability to tackle M/ILP problems by retrieving relevant domain-specific knowledge, such as constraints, problem formulations, and solution techniques, from an external knowledge base. \Ac{rag} can be used with fine-tuning. The \ac{llm} is first fine-tuned for a specific domain. Then, \ac{rag} provides the context that fills gaps in the fine-tuned model's understanding, especially for uncommon or highly specific tasks.

To the best of our knowledge, one study has considered using \ac{rag} to leverage the capabilities of \acp{llm} in mathematical modeling~\cite{liu2024generative}. 
Liu \textit{et al.}~\cite{liu2024generative} used \ac{rag} for data center networking in solving and formulating optimization problems. Their study demonstrates the fundamentals of using \acp{llm} for automatic formulations and presents the data center networking problems as a case study. Table~\ref{tab:comp_fine_rag} summarizes the main findings of the related literature.
\begin{table*}
\centering
\renewcommand{\arraystretch}{1.3}
\caption{Combinatorial optimization modeling via fine-tuning using RAGs.}
\label{tab:comp_fine_rag}
\resizebox{0.95\linewidth}{!}{
\small{
\begin{tabular}{|>{\centering\arraybackslash}m{0.8cm}
                ||>{\centering\arraybackslash}m{1.5cm}
                |>{\centering\arraybackslash}m{1.5cm}
                |>{\centering\arraybackslash}m{1.5cm}
                |>{\centering\arraybackslash}m{1.5cm}
                |>{\centering\arraybackslash}m{1.5cm}
                |>{\centering\arraybackslash}m{6.1cm}|}
\hline
\textbf{Study} & \textbf{RQ2} & \textbf{RQ3} & \textbf{RQ4} & \textbf{RQ5} & \textbf{RQ9} & \textbf{Main findings} \\
\hline \hline
\cite{liu2024generative} & Data center networking & Raw data & -- & \ac{gpt}-4 & Optimality gap & Presented a case study composed of two modules: Automatic optimization formulation and diffusion-empowered optimization solving. \\
\hline
\end{tabular}
}
}
\end{table*}


\subsection{Dataset-driven Approaches}\label{subsec:dataset finetuning}



Recently, researchers have proposed several benchmarks and datasets designed to facilitate the ability of \acp{llm} to solve mathematical problems. Each of these datasets is characterized by distinct properties, as detailed in Table~\ref{tab:dataset}.
The most common dataset is NL4Opt, which is used for a competition aimed at automating the process of mathematical modeling. This dataset comprises $1,101$ annotated \acp{lpwp} sourced from six distinct domains: Sales, advertising, investment, production, transportation, and sciences. Each problem from the NL4Opt dataset has approximately two constraints and two decision variables. 

In~\cite{ahmaditeshnizi2024optimus}, the authors constructed a similar dataset, NLP4LP, a comprehensive open-source dataset containing $355$ optimization problems, designed to provide broad coverage of various problem types. It integrates problems from existing datasets alongside newly introduced ones, ensuring diversity in complexity and description length. The dataset includes real-world optimization problems that are significantly longer than those found in other \ac{milp} modeling datasets.

Tang \textit{et al.}~\cite{tang2024orlm} proposed a more challenging dataset for \acp{llm}. This dataset consists of $100$ real-world problems reflecting the complexity and diversity encountered in industrial settings. 
Similarly, Xiao \textit{et al.}~\cite{xiao2023chain} constructed a dataset consisting of $37$ problems from diversified sources, including academic papers, textbooks, and real-world industry scenarios. These problems cover a wide range of subjects, spanning from supply chain optimization and scheduling problems to warehousing logistics. The Mamo dataset proposed by Huang \textit{et al.}~\cite{mamo} consists of two main types of problems: $196$ ordinary differential equations (ODEs) and $863$ optimization problems. What makes this dataset stand out is its integration with mathematical solvers.


Unlike previous datasets, OptiBench developed in~\cite{huang2024optibench} provides both linear and nonlinear optimization instances.  Other datasets, such as StructuredOR~\cite{jiang2024llmopt}, DiDi operational~\cite{wang2024leveraging}, and Cycle share~\cite{jiao2024city}, are not as popular as the other datasets. It is worth noting that many researchers constructed and used their own datasets. These datasets are usually designed to solve problems related to a specific and specialized field.
\begin{table}
\centering
\renewcommand{\arraystretch}{1.2}
\caption{Datasets.}
\label{tab:dataset}
\small{
\begin{tabular}{|>{\centering\arraybackslash}m{1.6cm}
                ||>{\centering\arraybackslash}m{1cm}
                |>{\centering\arraybackslash}m{1cm}
                |>{\centering\arraybackslash}m{3cm}|}
\hline
\textbf{Dataset} & \textbf{Size} & \textbf{Year} & \textbf{List of Papers} \\
\hline \hline 
NLP4LP & $355$ & 2024 & \cite{jiang2024llmopt, yu2024deep, ahmaditeshnizi2024optimus} \\
\hline
NL4Opt & $1011$ & 2021 & \cite{ahmaditeshnizi20241, tang2024orlm, zhang2024solving, jiang2024llmopt, ahmaditeshnizi2024optimus, astorga2024autoformulation} \\
\hline
ComplexOR & $37$ & 2024 & \cite{xiao2023chain, jiang2024llmopt, yu2024deep, ahmaditeshnizi2024optimus, wang2024large} \\
\hline
IndustryOR & $100$ & 2024 & \cite{tang2024orlm, astorga2024autoformulation} \\
\hline
OptiBench & $605$ & – & \cite{huang2024optibench} \\
\hline
Mamo & $863$ & 2024 & \cite{tang2024orlm, jiang2024llmopt} \\
\hline
StructuredOR & $124$ & – & \cite{jiang2024llmopt} \\
\hline
DiDi operational &$ 12,500$ & 2016 & \cite{wang2024leveraging} \\
\hline
StarJob & $120,000$ & 2024 & \cite{anonymous2024starjob} \\
\hline
Cycle share & $283,143$ & 2024 & \cite{jiao2024city} \\
\hline
ReSocratic& $29,000$&2024& ~\cite{huang2024optibench} \\
\hline
 Pyomo Cookbook& $40$&2018 &\cite{chen2024diagnosing}\\
 \hline
  GSM8K & $8500$&2021 &\cite{cobbe2021gsm8k}\\
   \hline
MultiArith & $600$&2016 & \cite{yang2024large}\\
 \hline
AQuA &$100000$ &2017 &\cite{chen2024diagnosing}\\
 \hline
BBH& $23$& 2022&\cite{chen2024diagnosing}\\
\hline
Ner4Opt&$1101$&2023&\cite{kadiouglu2024ner4opt}\\
\hline
Their own & – & – & \cite{li2023large, amarasinghe2023aicopilotbusinessoptimisationframework, zhang2024solving, abgaryan2024llms, anonymous2024droc, li2023synthesizing} \\
\hline
\end{tabular}
}
\end{table}

\subsubsection{Combinatorial Optimization}

Recent studies have explored the capabilities of \acp{llm} in addressing combinatorial optimization. A common approach across these works involves fine-tuning \acp{llm} on domain-specific datasets, such as StrJob, ReSocratic~\cite{huang2024optibench}, OptiBench, and StructuredOR, which were specifically designed to convert structured optimization problems into natural language formats suitable for \acp{llm} learning. These models, including \ac{gpt}, Llama, Phi3, and Qwen, have been evaluated using various techniques such as zero-/few-shot prompting, \ac{cot} reasoning, and \ac{rl}. 

Findings consistently indicate that \acp{llm} can be used to solve simple tasks, such as job shop scheduling, binary packing, and automated model generation. Fine-tuning strategies, such as \ac{lora}, parameter-efficient tuning (PEFT), and progressive fine-tuning, were shown to significantly enhance performance, especially when combined with methods such as temperature scaling and \ac{tot} reasoning. 

Notably, \acp{llm} tend to perform better on simple linear problems, while challenges remain in solving highly non-linear complex tasks. Almost all studies leveraged \acp{llm} to generate executable algorithm code, highlighting their potential not just as solvers but as meta-reasoners capable of producing optimization strategies. Overall, these works underscore the growing promise of \acp{llm} in combinatorial optimization and the importance of tailored datasets and adaptive fine-tuning techniques. Table~\ref{comfinedata} presents a comprehensive summary of all studies undertaken to utilize a dataset for the fine-tuning of \acp{llm}, enabling them to address mathematical modeling in the context of the specified review questions. 

\begin{table*}
\centering
\renewcommand{\arraystretch}{1.3}
\caption{Combinatorial optimization modeling via fine-tuning using datasets.}
\label{comfinedata}
\resizebox{0.9\linewidth}{!}{
\small{
\begin{tabular}{|>{\centering\arraybackslash}m{0.7cm}
                ||>{\centering\arraybackslash}m{1cm}
                |>{\centering\arraybackslash}m{1.7cm}
                |>{\centering\arraybackslash}m{0.7cm}
                |>{\centering\arraybackslash}m{1.2cm}
                |>{\centering\arraybackslash}m{1.4cm}
                |>{\centering\arraybackslash}m{1.3cm}
                |>{\centering\arraybackslash}m{6cm}|}
\hline
\textbf{Study} & \textbf{RQ2} & \textbf{RQ3} & \textbf{RQ4} & \textbf{RQ5} & \textbf{RQ6} & \textbf{RQ9} & \textbf{Main findings} \\
\hline \hline

\cite{abgaryan2024llms} & \ac{jss} & Their own & $120$k & Phi-3-Mini & \Ac{or}-Tools & Optimality gap & 
\begin{itemize}
  \item Developed a supervised dataset specifically designed to train \acp{llm} for Job Shop Scheduling Problems (JSSPs).
  \item Utilized the \ac{lora} for fine-tuning, demonstrating high-quality schedules for small-scale \ac{jss} problems.
\end{itemize} \\
\hline

\cite{huang2024optibench} & General & OptiBench & $816$ & \ac{gpt}-3.5, \ac{gpt}-4, Llama, DeepSeek & SCIP & Solution accuracy & 
\begin{itemize}
  \item Presented a new dataset, OptBinch, reflecting the complexity of real-world optimization challenges.
  \item Introduced a reverse method for generating synthetic data called ReSocratic.
\end{itemize} \\
\hline

\cite{anonymous2024starjob} & \ac{jss} & StarJob & $120$k & Llama & \Ac{or}-Tools & Optimality gap & 
\begin{itemize}
  \item Provided a dataset of $120$k on \ac{jss}. Applied \ac{llm} to the dataset.
  \item Fine-tuned the Llama model on the proposed dataset using \ac{lora}.
\end{itemize} \\
\hline

\cite{zhou2024llmsolver} & General & MIPLIB & $1065$ & \ac{gpt}-3.5, \ac{gpt}-4, Claude & SCIP & AIR, optimality gap & 
\begin{itemize}
  \item Proposed using \acp{llm} to produce high-quality algorithms for combinatorial optimization solvers.
  \item Utilized a derivative-free evolutionary framework that allows for efficient exploration and optimization of the generated algorithms.
\end{itemize} \\
\hline

\cite{wang2024leveraging} & General & DiDi operational & $12500$ & Llama, Instruct & Gurobi, CPLEX, COPT & Optimality gap & 
\begin{itemize}
  \item Proposed a framework for solving complex and mixed-integer programming (MIP) challenges.
  \item Introduced a method that adjusts the ``temperature'' parameter during the solution generation process.
\end{itemize} \\
\hline

\cite{wang2024bpp} & General & StructuredOR & $30$ & \ac{gpt}-4o & Gurobi & Solution accuracy, recall, precision, F1-score & 
\begin{itemize}
  \item Introduced a new dataset called StructuredOR.
  \item Proposed new searching techniques that integrate \ac{rl} into a \ac{tot}, incorporating beam search, PRM, and pairwise preference algorithm for enhancing the decision making.
\end{itemize}\\
\hline
\cite{li2023synthesizing} & General & Extended NL4Opt & -- & \ac{gpt}-3.5, \ac{gpt}-3 & -- & Solution accuracy & 
\begin{itemize}
  \item Extended the NL4Opt dataset with more problem descriptions and constraint types.
  \item Proposed a three-phase framework for determining the decision variables, constraints, and objective function.
  \item Presented a structured approach to classifying constraints.
\end{itemize} \\
\hline
\cite{ahmed2024lm4opt} & General & NL4Opt & $67$ & Llama & -- & Solution accuracy & 
\begin{itemize}
  \item Introduced LM4OPT, a progressive fine-tuning framework designed to enhance the performance of smaller \acp{llm}, such as Llama.
\end{itemize} \\
\hline
\end{tabular}
}
}
\end{table*}


\subsubsection{Combinatorial and Linear Optimization}

Motivated by the urge to reduce the reliance on closed-source \acp{llm} like \acp{gpt}, Tang~\textit{et al.}~\cite{tang2024orlm} proposed \ac{or} language models (ORLMs) trained on the IndustryOR dataset, which is a dataset constructed by the authors using a new data synthesis method, \ac{or}-Instruct. Their framework not only provides mathematical modeling for a wide variety of \ac{or} problems but also constructs a solver code for finding the objective value. 

Alibaba Group also worked on automating the process of mathematical modeling using a supervised dataset to fine-tune Qwen, as stated in~\cite{zhang2024solving}. The authors proposed OptLLM, which supports multi-round dialog to learn the mathematical model correctly. It has been experimentally proven that fine-tuning the \ac{llm} in OptLLM achieves better accuracy compared to the prompt-based
models. Similar to~\cite{tang2024orlm}, OptLLM uses an external solver to ensure that the model is solved correctly and to help decision-makers make decisions.

Jiang~\textit{et al.}~\cite{jiang2024llmopt} proposed LLMOpt, a model that automates mathematical modeling using the five-step approach (sets, parameters, variables, objective function, and constraints). Chen~\textit{et al.}~\cite{chen2025solver} integrated \ac{rl} with \acp{llm} in a solver, outperforming the \acp{llm} baseline solver.  
Table~\ref{comandlinearfinedata} summarizes the main findings of studies that utilize datasets in solving both linear and combinatorial problems using fine-tuning in relation to the established review questions..
\begin{table*}
\centering
\renewcommand{\arraystretch}{1.3}
\caption{Combinatorial and linear optimization modeling via fine-tuning using datasets.}
\label{comandlinearfinedata}
\resizebox{0.9\linewidth}{!}{
\small{
\begin{tabular}{|>{\centering\arraybackslash}m{0.7cm}
                ||>{\centering\arraybackslash}m{1cm}
                |>{\centering\arraybackslash}m{1.4cm}
                |>{\centering\arraybackslash}m{1.1cm}
                |>{\centering\arraybackslash}m{1.3cm}
                |>{\centering\arraybackslash}m{1cm}
                |>{\centering\arraybackslash}m{1.2cm}
                |>{\centering\arraybackslash}m{6.4cm}|}
\hline
\textbf{Study} & \textbf{RQ2} & \textbf{RQ3} & \textbf{RQ4} & \textbf{RQ5} & \textbf{RQ6} & \textbf{RQ9} & \textbf{Main findings} \\
\hline \hline

\cite{tang2024orlm} & General & IndustryOR & 100 & Mistral, Llama, DeepSeek Math & COPT & Solution accuracy & 
\begin{itemize}
  \item Developed \ac{or}-Instruct, a semi-automated process for generating synthetic data tailored to optimization modeling.
  \item Generated an industrial benchmark designed to evaluate \acp{llm} on real-world \ac{or} problems.
\end{itemize} \\
\hline

\cite{zhang2024solving} & General & -- & -- & Qwen, \ac{gpt}-3.5, \ac{gpt}-4 & --, MAPL Code & -- & 
\begin{itemize}
  \item Proposed OptLLM framework that supports iterative dialogues for solving optimization problems.
  \item Provided tutorials on three typical optimization applications and conducted experiments using both prompt-based \ac{gpt} models and a fine-tuned Qwen model.
\end{itemize} \\
\hline

\cite{jiang2024llmopt} & General & NL4Opt, Mamo, IndustryOR, NLP4LP, ComplexOR & NL4Opt: $1101$, Mamo: $863$, IndustryOR: $100$, NLP4LP: $65$, ComplexOR: $19$ & Qwen1.5 & --, Pyomo & Execution rate, solution accuracy, average solving time & 
\begin{itemize}
  \item Proposed LLMOPT that boosts optimization generalization through multi-instruction fine-tuning and model alignment for improving accuracy in problem solving and expanding the range of problem types that the model can handle.
\end{itemize} \\
\hline

\cite{chen2025solver} & General & NL4Opt, Mamo, IndustryOR & -- & Qwen2.5-7B, Qwen2.5-14B, DeepSeek, \ac{gpt}-4 & Gurobi & Solution accuracy, execution rate & 
\begin{itemize}
  \item Solver-informed \ac{rl} significantly outperforms baseline \acp{llm} (e.g., CodeLlama, \ac{gpt}-4) in mathematical reasoning tasks by integrating external verification signals.
\end{itemize} \\
\hline

\end{tabular}
}
}
\end{table*}


\subsubsection{Linear Optimization}

Solving linear optimization problems is considered the most straightforward entry point for enabling \acp{llm} to address mathematical modeling tasks, since these problems have well-defined structures, established solution methods, and lower computational complexity compared to nonlinear or combinatorial optimization problems. Amarasinghe~\textit{et al.}~\cite{amarasinghe2023aicopilotbusinessoptimisationframework} proposed AI-Copilot, an approach for \ac{llm} fine-tuning to automate business optimization problems. They introduced modularization and prompt engineering techniques to reduce the token length limitation. Although the proposed AI-Copilot was tested on the \ac{jss} problem formulation, this framework can be used for other types of business optimization problems. Similarly, Kad{\i}o{\u{g}}lu~\textit{et al.}~\cite{kadiouglu2024ner4opt} proposed Ner4Opt, combining classical \ac{nlp} with modern \acp{llm} and fine-tuned transformers, and achieved $50\%$ improvement. Table~\ref{lineardatafine} provides an overview of AI-Copilot and Ner4Opt in the context of the predefined review questions.
\begin{table*}
\centering
\renewcommand{\arraystretch}{1.3}
\caption{Linear optimization modeling via fine-tuning using datasets.}
\label{lineardatafine}
\resizebox{0.9\linewidth}{!}{
\small{
\begin{tabular}{|>{\centering\arraybackslash}m{0.7cm}
                ||>{\centering\arraybackslash}m{1cm}
                |>{\centering\arraybackslash}m{1.5cm}
                |>{\centering\arraybackslash}m{1.1cm}
                |>{\centering\arraybackslash}m{1.1cm}
                |>{\centering\arraybackslash}m{1.1cm}
                |>{\centering\arraybackslash}m{1.2cm}
                |>{\centering\arraybackslash}m{6.4cm}|}
\hline
\textbf{Study} & \textbf{RQ2} & \textbf{RQ3} & \textbf{RQ4} & \textbf{RQ5} & \textbf{RQ6} & \textbf{RQ9} & \textbf{Main findings} \\
\hline \hline

\cite{amarasinghe2023aicopilotbusinessoptimisationframework} & \ac{jss} & -- & $182$ & CodeRL &--, CPMpy & Solution accuracy & 
\begin{itemize}
  \item Proposed a new fine-tuning approach based on AI-Copilot for business optimization problem formulation.
  \item Developed new modularization and prompting techniques for complex problems.
  \item Proposed a new evaluation metric.
  \item Constructed a new dataset.
\end{itemize} \\
\hline

\cite{kadiouglu2024ner4opt} & General & Ner4Opt & $1101$ & \ac{gpt}-4 & --, MiniZinc & Precision, recall, F1-score, compilation accuracy & 
\begin{itemize}
  \item Presented Ner4Opt, addressing optimization-specific entity extraction from natural language.
  \item Utilized a hybrid approach combining classical \acp{nlp} with modern \acp{llm} and fine-tuned transformers.
  \item Improved optimization model compilation by nearly $50\%$ when guided by Ner4Opt annotations.
\end{itemize} \\
\hline

\end{tabular}
}
}
\end{table*}

\section{Mathematical Modeling via In-context Learning}\label{incontext}

Although fine-tuning is a powerful learning approach for adapting \acp{llm} to specific domains, it requires more computational power as it involves updating the model's weights. In-context learning offers a more lightweight and flexible alternative, allowing \acp{llm} to adapt to mathematical programming tasks without modifying their underlying parameters. In-context learning enables models to generalize from examples provided directly within the input prompt, making it an efficient and dynamic approach.

Depending on how many examples are used by the  \ac{llm}, there are different types of in-context learning: Zero shots, one-shot, few shots, \ac{cot}, \ac{tot}, and \ac{got}. In zero-shot learning, only the task description is provided to the \ac{llm}. One or a few examples, along with the task description, are provided when using one-shot and few-shot learning. In \ac{cot}, the task and reasoning guidance are provided. \Ac{tot} is an extension of the \ac{cot}; rather than following a single linear reasoning path, \ac{tot} explores multiple reasoning paths branching out from a given state, evaluating them to find the most optimal solution. Similarly, \ac{got} represents the reasoning paths as a graph rather than a tree.

Similar to fine-tuning, we classified the studies that adopted in-context learning in the literature based on how the knowledge base is constructed and the type of optimization model considered, as shown in Fig.~\ref{fig:incontexttax}.

\subsection{RAG-based Approaches}\label{subsec:RAG incontext}

\Ac{rag} can be combined with in-context learning, creating a powerful method for overcoming the prompt length limitation, which is the case in mathematical modeling. With \ac{rag}, the retrieved information and task-specific examples are included as part of the input prompt.

In~\cite{ahmaditeshnizi2024optimus}, the authors utilized \ac{rag} to enhance the capabilities of their framework, OptiMUS-3, which is an \ac{llm}-based agent framework designed for solving general-purpose \ac{milp} problems. The primary reason for implementing \ac{rag} was to improve the generation of accurate model objectives and constraints in their previous framework, OptiMUS~\cite{ahmaditeshnizi20241}. By retrieving examples of similar constraints and their formulations, these examples were incorporated into the \ac{llm} prompt to refine the model's output.

OptiMUS-3 consists of four main steps: The first is to divide the optimization problem into smaller, manageable sub-tasks. The second step is to extract each constraint and objective independently while employing a graph to represent relationships between the components of the optimization problem. In the third step, the \ac{llm} will generate a code for solving the \ac{milp} problems and iteratively execute and debug to achieve the desired results. Finally, the accuracy of a given solution is calculated. The study further demonstrated that adapting \ac{rag} with \ac{gpt}-4 and Llama improved the accuracy of the generated solution. Using \acp{rag} drastically enhanced the \ac{llm} capabilities in solving \acp{vrp}, as experimentally shown in~\cite{anonymous2024droc}.
Zhang~\textit{et al.}~\cite{zhang2024generative} employed \ac{rag} to retrieve domain-specific satellite knowledge, thereby enhancing the capability of \acp{llm} to formulate and solve satellite optimization problems. Table~\ref{combinatorial_incontext_rag} summarizes the literature that tackles combinatorial optimization modeling via in-context learning using \acp{rag} in light of the predefined review questions.

\begin{table*}
\centering
\renewcommand{\arraystretch}{1.3}
\caption{Combinatorial optimization modeling via in-context learning using RAGs.}
\label{combinatorial_incontext_rag}
\resizebox{0.9\linewidth}{!}{
\small{
\begin{tabular}{|>{\centering\arraybackslash}m{0.7cm}
                ||>{\centering\arraybackslash}m{1cm}
                |>{\centering\arraybackslash}m{1.6cm}
                |>{\centering\arraybackslash}m{1.1cm}
                |>{\centering\arraybackslash}m{1.1cm}
                |>{\centering\arraybackslash}m{1.1cm}
                |>{\centering\arraybackslash}m{1.3cm}
                |>{\centering\arraybackslash}m{6.2cm}|}
\hline
\textbf{Study} & \textbf{RQ2} & \textbf{RQ3} & \textbf{RQ4} & \textbf{RQ5} & \textbf{RQ6} & \textbf{RQ9} & \textbf{Main findings} \\
\hline \hline

\cite{anonymous2024droc} & \Acp{vrp} & -- & 48 & Llama, \ac{gpt}-4, \ac{gpt}-3.5 Turbo & \Ac{or}-Tools, Gurobi & Solution accuracy and optimality gap &
\begin{itemize}
  \item Proposed a new framework to enhance \acp{llm} in exploiting solvers to solve \acp{vrp} using \ac{rag} to extract the knowledge.
\end{itemize} \\
\hline

\cite{ahmaditeshnizi2024optimus} & General & NL4Opt, ComplexOR, NLP4LP & NL4Opt: $1101$, ComplexOR: $37$, NLP4LP: $354$ & \ac{gpt}-4o, Llama & Not specified & Solution accuracy &
\begin{itemize}
  \item Employed an \ac{llm}-based agent, a modular architecture that processes each constraint and objective independently.
  \item Utilized a graph to represent relationships between different components of the optimization problem.
\end{itemize} \\
\hline

\cite{zhang2024generative} & Wireless resource allocation & Their own & $8$ & \ac{gpt}-3.5 & -- & Utility improvement, relative regret, convergence performance, running time &
\begin{itemize}
  \item Proposed an approach to design transmission strategies in satellite communication networks using generative AI and a \ac{moe} with the proximal policy optimization (PPO) method.
  \item Utilized \ac{rag} to retrieve satellite expert knowledge that supports mathematical modeling.
  \item \ac{moe}-PPO outperforms traditional methods, greedy, and random baselines in performance (e.g., sum-rate and energy efficiency).
\end{itemize} \\
\hline

\end{tabular}
}
}
\end{table*}

\subsection{Dataset-driven Approaches}\label{subsec:dataset incontext}

When utilizing datasets for in-context learning, \acp{llm} learn patterns and relationships by relying on examples given in the input prompt rather than making explicit parameter updates. 
In the context of mathematical modeling, this approach leverages \ac{llm} capabilities in learning how to approach \ac{or} problems without requiring extensive retraining. By carefully crafting datasets and optimizing the selection of in-context examples, researchers can improve model adaptability, reduce computational costs, and enhance decision-making. 

\subsubsection{Combinatorial Optimization}

Combinatorial problems and in-context learning via datasets are the most common directions among researchers in the field of mathematical modeling due to the existence of combinatorial problems in real-world scenarios and the simplicity of implementing in-context learning. 
Consequently, OptiMUS-3~\cite{ahmaditeshnizi2024optimus} used \ac{rag} to incorporate relevant past examples of mathematical formulations into its prompts, enabling better generalization without retraining. On the other hand, Wang~\textit{et al.}~\cite{anonymous2024optibench} presented a new dataset to enhance the ability of \acp{llm} in mathematical modeling. 
The \ac{coe} framework employs a multi-agent setup, where each agent specializes in an \ac{or} task under a central conductor, improving complex problem-solving through \ac{cot} reasoning~\cite{xiao2023chain, nammouchi2024towards}.
OptiChat~\cite{chen2024diagnosing} offers natural language-based explanations and diagnostics for infeasible optimization models by combining \ac{gpt}-4 with solver tools and enhanced prompting techniques, such as \ac{cot} and sentiment analysis. Similarly, \ac{gpt}-4 in~\cite{gemp2024steering} was used for negotiation tasks for a game-theoretic framework.

Several other frameworks explored different problem-solving strategies by integrating \acp{llm} with optimization tools. MEoH~\cite{yao2024multi}, proposed by Microsoft, used evolutionary algorithms and zero-shot in-context learning to solve multi-objective problems, such as \ac{tsp} and the bin packing problem. \ac{milp}-Evolve~\cite{li2024towards} generates diverse \ac{milp} instances using \ac{cot}-guided prompting. OPRO~\cite{yang2024large}, proposed by DeepMind, tackled combinatorial problems by iteratively refining prompts based on previous solution quality. Additionally, MLPrompt~\cite{wang2024large, astorga2024autoformulation} coupled \acp{llm} with Monte Carlo tree search (MCTS) to explore and refine optimization hypotheses, leveraging few-shot prompting for model generation.
Liu~\textit{et al.}~\cite{liu2024variable} integrated rule-based extraction with \acp{llm}. In~\cite{wang2025ormind}, Wang~\textit{et al.} used \acp{llm} to create a cognitive-inspired framework. 

Other efforts focus on domain-specific applications and generalization strategies. LLMFP~\cite{hao2024planning} provided a zero-shot planning framework that models and solves optimization problems through self-assessment and code refinement. City-LEO~\cite{jiao2024city} used \acp{llm} in agent-based city management, applying in-context learning for prediction and optimization. 
A similar problem that is common in cities is \ac{tsp}, which was tackled by Luzzi~\textit{et al.}~\cite{luzzi2025chatgpt} using \acp{llm}. 

Lastly, frameworks, such as the one proposed by Sun~\textit{et al.}~\cite{sun2024generative}, applied \acp{llm} as black-box search operators, decomposing \ac{uav}-related multi-objective problems into smaller sub-problems. Collectively, these approaches showcase the growing ability of \acp{llm} to autonomously understand, solve, and optimize complex real-world problems across domains. Table~\ref{incontextDatasetCombinatorial} summarizes the aforementioned studies in light of the predefined review questions. From Table~\ref{incontextDatasetCombinatorial}, it can be observed that the studies~\cite{ahmaditeshnizi2024optimus, sun2024generative,luzzi2025chatgpt} did not consider employing solvers to determine the objective value of the generated mathematical models.
\begin{table*}
\centering
\renewcommand{\arraystretch}{1.3}
\caption{Combinatorial optimization modeling via in-context learning using datasets.}
\label{incontextDatasetCombinatorial}
\resizebox{0.9\linewidth}{!}{
\small{
\begin{tabular}{|>{\centering\arraybackslash}m{0.7cm}
                ||>{\centering\arraybackslash}m{1cm}
                |>{\centering\arraybackslash}m{1.6cm}
                |>{\centering\arraybackslash}m{1.1cm}
                |>{\centering\arraybackslash}m{1.1cm}
                |>{\centering\arraybackslash}m{1cm}
                |>{\centering\arraybackslash}m{1.2cm}
                |>{\centering\arraybackslash}m{6.4cm}|}
\hline
\textbf{Study} & \textbf{RQ2} & \textbf{RQ3} & \textbf{RQ4} & \textbf{RQ5} & \textbf{RQ6} & \textbf{RQ9} & \textbf{Main findings} \\
\hline \hline

\cite{ahmaditeshnizi2024optimus} & General & NL4Opt, ComplexOR, NLP4LP & NL4Opt: $1101$, ComplexOR: $37$, NLP4LP: $354$ & \ac{gpt}-4o, Llama & -- & Solution accuracy &
\begin{itemize}
  \item Employed an \ac{llm}-based agent, a modular architecture that processes each constraint and objective independently.
  \item Utilized a graph to represent relationships between different components of the optimization problem.
\end{itemize} \\
\hline

\cite{xiao2023chain} & General & ComplexOR, \ac{lpwp} & $1286$ & \ac{gpt}-3.5 & Gurobi & Solution accuracy, compilation accuracy, running time &
\begin{itemize}
  \item Presented a new dataset, ComplexOR, which consists of complex \ac{or} problems.
  \item Proposed a new methodology for utilizing \acp{llm} to solve \ac{or} problems called \ac{coe}, involving a multi-agent solution where each agent is assigned a specific role and domain knowledge.
\end{itemize} \\
\hline

\cite{chen2024diagnosing} & General & Pyomo Cookbook
 & $63$ & \ac{gpt}-4 & Gurobi & Solution accuracy &
\begin{itemize}
  \item Proposed OptiChat that helps users understand and troubleshoot infeasible optimization models using natural language.
  \item Utilized \ac{gpt}-4 to interface with solvers and identify the minimal subset of constraints causing infeasibility.
\end{itemize} \\
\hline

\cite{yao2024multi} & General & -- & $5064$ & \ac{gpt}-3.5-turbo & Not specified & HV, IGD &
\begin{itemize}
  \item Developed MEoH, which automatically generates a diverse set of heuristics in a single run that offers more trade-off options than existing methods.
\end{itemize} \\
\hline

\cite{li2024towards} & General & MIPLIB, Their own & $1$ Million & \ac{gpt}-4o & Not specified & Integrity gap &
\begin{itemize}
  \item Proposed a new framework \ac{milp}-Evolve based on \ac{gnn} for generating \ac{milp} instances for \ac{llm} learning.
\end{itemize} \\
\hline

\cite{anonymous2024optibench} & General & OptiBench & $816$ & \ac{gpt}-3.5, \ac{gpt}-4, Llama, DeepSeek & SCIP & Solution accuracy &
\begin{itemize}
  \item Presented a new dataset, OptBinch, reflecting the complexity of real-world optimization challenges.
  \item Employed a model-data separation format to enhance \ac{llm}'s understanding during evaluation.
\end{itemize} \\
\hline

\cite{yang2024large} & General & GSM8K, MultiArith, AQuA, BBH & GSM8K: $7473$, BBH: $250$/task & PaLM 2-L, \ac{gpt}-3.5-turbo, \ac{gpt}-4 & Gurobi & Solution accuracy &
\begin{itemize}
  \item Introduced OPRO, a novel method leveraging \acp{llm} to solve math and \ac{tsp} problems and tune parameters in linear regression.
\end{itemize} \\
\hline

\cite{sun2024generative} & \Ac{uav} & Raw data & -- & -- & -- & Optimality gap &
\begin{itemize}
  \item Proposed a novel generative AI-based framework for \ac{uav} networking.
  \item Presented a case study on optimizing transmission rate and spectrum map estimation.
\end{itemize} \\
\hline

\cite{wang2024large} & General & ComplexOR & 60 & \ac{gpt}-3.5, \ac{gpt}-4, \ac{gpt}-4o, \ac{gpt}-4o-mini & Gurobi, OptVerse, CPLEX & Solution accuracy &
\begin{itemize}
  \item Introduced MLPrompt, a novel prompting strategy that translates error-prone rules into less dominant language representations, enhancing \ac{llm} reasoning in complex contexts.
\end{itemize} \\
\hline
\cite{astorga2024autoformulation} & General & NL4Opt, IndustryOR & $344$ & \ac{gpt}-4o-mini & Gurobi, CPLEX & Solution accuracy &
\begin{itemize}
  \item Introduced a novel method that leverages \acp{llm} within a Monte Carlo tree search framework by exploiting the hierarchical nature of optimization modeling.
\end{itemize} \\
\hline

\end{tabular}
}
}
\end{table*}

\begin{table*}
\centering
\renewcommand{\arraystretch}{1.3}
\resizebox{0.9\linewidth}{!}{
\small{
\begin{tabular}{|>{\centering\arraybackslash}m{0.8cm}
                ||>{\centering\arraybackslash}m{1.2cm}
                |>{\centering\arraybackslash}m{1.7cm}
                |>{\centering\arraybackslash}m{1cm}
                |>{\centering\arraybackslash}m{1.2cm}
                |>{\centering\arraybackslash}m{1.5cm}
                |>{\centering\arraybackslash}m{1.3cm}
                |>{\centering\arraybackslash}m{5.6cm}|}
\hline
\textbf{Study} & \textbf{RQ2} & \textbf{RQ3} & \textbf{RQ4} & \textbf{RQ5} & \textbf{RQ6} & \textbf{RQ9} & \textbf{Main findings} \\
\hline \hline


\cite{hao2024planning} & Planning & -- & $9$ & \ac{gpt}-4, CLAUDE-3.5 & \Ac{smt} solver & Optimality gap &
\begin{itemize}
  \item Demonstrated the capability of \acp{llm} to handle planning tasks without fine-tuning.
  \item Introduced a framework integrating \acp{llm} into formal programming pipelines with interpretable and verifiable outputs.
\end{itemize} \\
\hline

\cite{jiao2024city} & Cycle Share & -- & $283,143$ trip & -- & Gurobi & Solution accuracy &
\begin{itemize}
  \item Presented City-LEO, an \ac{llm}-based agent for efficient and transparent city management.
  \item Integrated prediction and optimization to handle environmental uncertainty and complex queries.
\end{itemize} \\
\hline

\cite{wang2025ormind} & General & NL4Opt, ComplexOR & NL4Opt: $1101$, ComplexOR: $36$ & \ac{gpt}-3.5-turbo, \ac{gpt}-4, \ac{gpt}-4o-mini &--, PuLP & Solution accuracy, execution rate &
\begin{itemize}
  \item Proposed ORMind, a cognitive-inspired framework achieving $68.8\%$ on NL4Opt and $40.5\%$ on ComplexOR.
  \item Outperformed \ac{cot}, reflection, and \ac{tot}; deployed successfully at Lenovo AI assistant.
\end{itemize} \\
\hline

\cite{huang2025llms} & General & -- & 863 & \ac{gpt}-4 series, Claude, Gemini, DeepSeek, Llama-3.1, Qwen-2.5, Mixtral & COPT, Gurobi & Solution accuracy &
\begin{itemize}
  \item Introduced Mamo dataset covering ODEs, \ac{lp}, \ac{milp}.
  \item Proposed a process-oriented framework for automatic mathematical modeling using \acp{llm}.
\end{itemize} \\
\hline

\cite{nammouchi2024towards} & Resource allocation problem & -- & $290$ & \ac{gpt}-4 and \ac{gpt}-3.5-turbo & Gurobi & Solution accuracy, running time, expert qualitative assessment &
\begin{itemize}
  \item Presented Chat-SGP framework for translating NL queries into Gurobi-based optimization formulations.
  \item Used a multi-agent setup (coder, optimizer, interpreter) for better control, debugging, and interpretability.
  \item Provided clear human-readable explanations for the results.
\end{itemize} \\
\hline

\cite{luzzi2025chatgpt} & Shortest path problem & -- & $120$ & \ac{gpt}-4 & -- & Average solving time &
\begin{itemize}
  \item Utilized ChatGPT to address several variants of shortest path problems.
\end{itemize} \\
\hline
\cite{liu2024variable} & Optimize variable extraction & Scientific variable extraction benchmark & $22$ & \ac{gpt}-3.5-turbo, \ac{gpt}-4o, Llama-3-8B-Instruct, Mistral-7B-Instruct & -- & Semantic similarity &
\begin{itemize}
  \item Demonstrated the best performance in extracting mathematical model variables from scientific literature using transfer learning and instruction tuning.
  \item Integrated rule-based extraction outputs with \acp{llm} to boost performance.
\end{itemize} \\
\hline

\cite{gemp2024steering} & General & -- & $3200$ & PaLM 2 & -- & Utility improvement &
\begin{itemize}
  \item Presented a new framework that integrates game-theoretic solvers with natural language dialogue using \acp{llm}.
  \item Produced less exploitable and more rewarding dialogue in negotiation tasks when \acp{llm} were guided by game-theoretic solvers.
\end{itemize} \\
\hline
\end{tabular}
}
}
\end{table*}





\subsubsection{Combinatorial and Linear Optimization}


Tang~\textit{et al.}~\cite{tang2024orlm} aimed to create a dataset that represents real-world scenarios, featuring a variety of combinatorial and linear optimization problems for in-context learning. To achieve this, they developed \ac{or}-Instruct for generating synthetic data. When addressing both combinatorial and linear optimization problems, exploratory questions and iterative discussions become essential. Therefore, Zhang~\textit{et al.}~\cite{zhang2024solving} and Li~\textit{et al.}~\cite{li2023large} tackled these issues by offering answers to what-if questions and providing iterative dialogues.


Multi-agent \acp{llm} help in decomposing the problem into smaller subproblems and solving each problem independently. For example, in OptiMUS~\cite{ahmaditeshnizi2024optimus}, the process is initiated by transforming a structured problem into three essential components: Parameters, clauses (which encompass objectives and constraints), and background information. These components are subsequently analyzed by four agents that collaborate to create a connection graph for the constraints. Similarly, Mostajabdaveh~\textit{et al.}~\cite{mostajabdaveh2024optimization} used a multi-agent \ac{llm} in their framework. 

Other frameworks, such as NL2OR~\cite{li2024nl2or}, consist of four key stages: Identifying query type, converting natural language to a domain-specific language, building and instantiating an abstract model, and finally storing and reporting solutions. Together, these studies highlight the growing sophistication and adaptability of \acp{llm} in optimization modeling via structured in-context learning and agent collaboration.
Table~\ref{incontextDatacombandlinear} provides a summary of the studies in light of the predefined review questions. 
\begin{table*}
\centering
\renewcommand{\arraystretch}{1.3}
\caption{Combinatorial and linear optimization modeling via in-context learning using datasets.}
\label{incontextDatacombandlinear}
\resizebox{0.9\linewidth}{!}{
\small{
\begin{tabular}{|>{\centering\arraybackslash}m{0.8cm}
                ||>{\centering\arraybackslash}m{1.2cm}
                |>{\centering\arraybackslash}m{1.7cm}
                |>{\centering\arraybackslash}m{1cm}
                |>{\centering\arraybackslash}m{1.2cm}
                |>{\centering\arraybackslash}m{1.5cm}
                |>{\centering\arraybackslash}m{1.3cm}
                |>{\centering\arraybackslash}m{5.6cm}|}
\hline
\textbf{Study} & \textbf{RQ2} & \textbf{RQ3} & \textbf{RQ4} & \textbf{RQ5} & \textbf{RQ6} & \textbf{RQ9} & \textbf{Main findings} \\
\hline \hline

\cite{li2023large} & General & Their own & 57 & \ac{gpt}-4, text-davinci-003 & Gurobi & Solution accuracy &
\begin{itemize}
  \item Proposed OptiGuide that uses \acp{llm} to translate human queries to optimization code.
  \item OptiGuide handles what-if queries by modifying inputs and rerunning the solver.
\end{itemize} \\
\hline

\cite{ahmaditeshnizi2024optimus} & General & NL4Opt & 67 & \ac{gpt}-4 & Gurobi & Solution accuracy &
\begin{itemize}
  \item Proposed a new dataset NL4Opt.
  \item Developed a framework using \ac{gpt}-4 with agents to solve \ac{milp} problems.
\end{itemize} \\
\hline

\cite{tang2024orlm} & General & IndustryOR & 100 & Mistral, Llama, DeepSeek, Math & COPT & Solution accuracy &
\begin{itemize}
  \item Developed \ac{or}-Instruct, a semi-automated pipeline for generating synthetic data for optimization.
  \item Created an industrial benchmark for evaluating \acp{llm} on real-world problems.
\end{itemize} \\
\hline

\cite{zhang2024solving} & General & -- & -- & Qwen, \ac{gpt}-3.5, \ac{gpt}-4 & --, MAPL Code & -- &
\begin{itemize}
  \item Proposed the OptLLM framework supporting iterative dialogue for solving optimization problems.
  \item Provided tutorials and experimental comparisons between prompt-based \ac{gpt} and fine-tuned Qwen models.
\end{itemize} \\
\hline

\cite{mostajabdaveh2024optimization} & General & NL4Opt, NLP4LP, custom dataset & -- & Llama, Zephyr & --, Pyomo & Solution accuracy &
\begin{itemize}
  \item Introduced a multi-agent, multi-stage approach where distinct \acp{llm} are used across stages to handle complex optimization modeling.
\end{itemize} \\
\hline

\cite{li2024nl2or} & General & -- & 30 & \ac{gpt}-3.5, \ac{gpt}-4 & Gurobi, CPLEX, \ac{or}-Tools, AMPL & Valid@k within time $t$&
\begin{itemize}
  \item Proposed NL2OR, an end-to-end pipeline to automate formulation and solve \ac{or} problems.
\end{itemize} \\
\hline

\end{tabular}
}
}
\end{table*}

\subsubsection{Linear Optimization}

In the context of solving only linear mathematical modeling using in-context learning, Zhang~\textit{et al.}~\cite{zhang2025or} demonstrated that their framework can achieve $85.54$ solution accuracy through \acp{cot} prompts. On the other hand, Deng~\textit{et al.}~\cite{deng24cafa} introduced CAFA, a method that simplifies \ac{lp} problem-solving by employing a single, compact prompt. This approach eliminates the complexity of multi-step pipelines by directly guiding \acp{llm} to generate executable optimization code, improving efficiency and usability in \ac{lp} problem formulation and execution. 
Li~\textit{et al.}~\cite{li2025abstract} developed an end-to-end framework that allows non-expert users to create and edit abstract \ac{or} models using queries expressed in natural language. Similarly, Zhang~\textit{et al.}~\cite{zhang2025decision} introduced the use of \acp{llm} as an explainable tool for mathematical models. Kad{\i}o{\u{g}}lu~\textit{et al.}~\cite{kadiouglu2024ner4opt} presented a method that combined feature engineering and data augmentation to exploit the language of optimization problems and solve annotated linear optimization problems.
Table~\ref{incontextdasetlinear} presents a summary of the literature in light of the review questions.

\begin{table*}
\centering
\renewcommand{\arraystretch}{1.3}
\caption{Linear optimization modeling via in-context learning using datasets.}
\label{incontextdasetlinear}
\resizebox{0.9\linewidth}{!}{
\small{
\begin{tabular}{|>{\centering\arraybackslash}m{0.8cm}
                ||>{\centering\arraybackslash}m{1.2cm}
                |>{\centering\arraybackslash}m{1.7cm}
                |>{\centering\arraybackslash}m{1cm}
                |>{\centering\arraybackslash}m{1.2cm}
                |>{\centering\arraybackslash}m{1.5cm}
                |>{\centering\arraybackslash}m{1.3cm}
                |>{\centering\arraybackslash}m{5.6cm}|}
\hline
\textbf{Study} & \textbf{RQ2} & \textbf{RQ3} & \textbf{RQ4} & \textbf{RQ5} & \textbf{RQ6} & \textbf{RQ9} & \textbf{Main findings} \\
\hline \hline

\cite{deng24cafa} & General & NL4Opt & -- & \ac{gpt}-4, \ac{gpt}-3.5-turbo, DeepSeek, Llama & CPLEX, Gurobi & Solution accuracy &
\begin{itemize}
  \item Proposed an \ac{llm}-based model to translate \ac{lp} problems into executable code.
  \item Introduced CAFA, which uses a compact prompt to instruct \acp{llm} to generate solver-ready code, simplifying pipelines.
\end{itemize} \\
\hline

\cite{zhang2025or} & General & -- & 83 & \ac{gpt}-3.5-o-mini, \ac{gpt}-4o, DeepSeek, Gemini Flash, Claude & Gurobi & Execution rate, solution accuracy &
\begin{itemize}
  \item \Ac{or}-\ac{llm}-Agent achieved 100\% execution success and 85.54\% solution accuracy.
  \item Outperformed SOTA \acp{llm} on real-world \ac{or} tasks.
  \item Demonstrated robust \ac{cot}-based reasoning and self-verification via sandboxed Gurobi execution.
\end{itemize} \\
\hline

\cite{li2025abstract} & General & \ac{lpwp} & 287 & \ac{gpt}-3.5, \ac{gpt}-4, \ac{gpt}-4o & Gurobi, CPLEX, \ac{or}-Tools, AMPL & Solution accuracy &
\begin{itemize}
  \item Presented NL2OR: an end-to-end framework with a multi-turn chat system for automating mathematical programming.
\end{itemize} \\
\hline

\cite{kadiouglu2024ner4opt} & General & Ner4Opt & 1101 & \ac{gpt}-4 & --, MiniZinc & Precision, recall, F1-score, compilation accuracy &
\begin{itemize}
  \item Presented Ner4Opt, addressing optimization-specific entity extraction from natural language.
  \item Improved model compilation performance by nearly $50\%$ via NER-guided annotations.
\end{itemize} \\
\hline

\cite{zhang2025decision} & General & -- & -- & \ac{gpt}-4, \ac{gpt}-4-turbo & Gurobi & Solution accuracy &
\begin{itemize}
  \item Proposed explainable operations research (EOR), enabling \acp{llm} to generate human-readable explanations of \ac{or} model behavior.
  \item Quantified ``decision information'' via bipartite graphs and supported what-if analysis.
  \item Released a new industrial benchmark for explainable \ac{or}.
\end{itemize} \\
\hline

\end{tabular}
}
}
\end{table*}

\section{Meta-Analysis}\label{meatanalysis}
In order to answer the main research question of this study, in this section, we conduct a meta-analysis based on the aforementioned review questions. These review questions reveal a profound grasp of the latest trends in leveraging \acp{llm} for mathematical modeling, highlighting their significance and potential impact.

\begin{figure}
    \centering
    \includegraphics[width=0.7\linewidth]{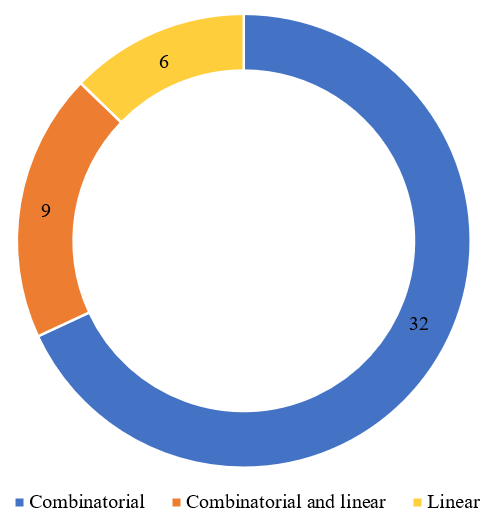}
    \caption{Classes of optimization problems studied in the literature using \acp{llm}, with combinatorial problems being the most frequent.}
    \label{fig:RQ1}
\end{figure}

Fifty-four papers were used to conduct the following meta-analysis. The first step in utilizing \acp{llm} for automating mathematical modeling is to determine the specific domain of interest. This helps define the types of problems that can be addressed using \acp{llm}. As shown in Fig.~\ref{fig:RQ1}, the results of \textbf{RQ1} indicate that combinatorial problems are the most prevalent in mathematical modeling. The main reason for this trend is that the current capabilities of \acp{llm} in reasoning make \ac{lp} problems not particularly challenging.
Additionally, combinatorial problems reflect the complexity of real-world problems.

Using \acp{llm} in mathematical modeling is a relatively new topic that has only recently begun to develop; thus, it is worth noting that most studies were not specialized in a specific domain, meaning that they tend to test the \ac{llm}'s capability in solving general problems. In other words, the answers to \textbf{RQ2} showed that $34$ papers focused on general domains.  

\subsection{Dataset Construction}\label{subsec:dataset construction}

Constructing a knowledge base for training \acp{llm} is a crucial step, as shown in Fig.~\ref{fig:steps}. There are two approaches: We can either use a predefined dataset to train the \ac{llm} or employ \ac{rag}. Out of the $54$ papers, only four employ \ac{rag}; this is because most papers tend to address general problems rather than domain-specific problems. \Ac{rag} is a powerful tool to address many issues, and one of them is to make \acp{llm} capable of solving domain-specific tasks, as discussed in~\cite{Raft}. 

\begin{figure}
\centering\includegraphics[width=1\linewidth]{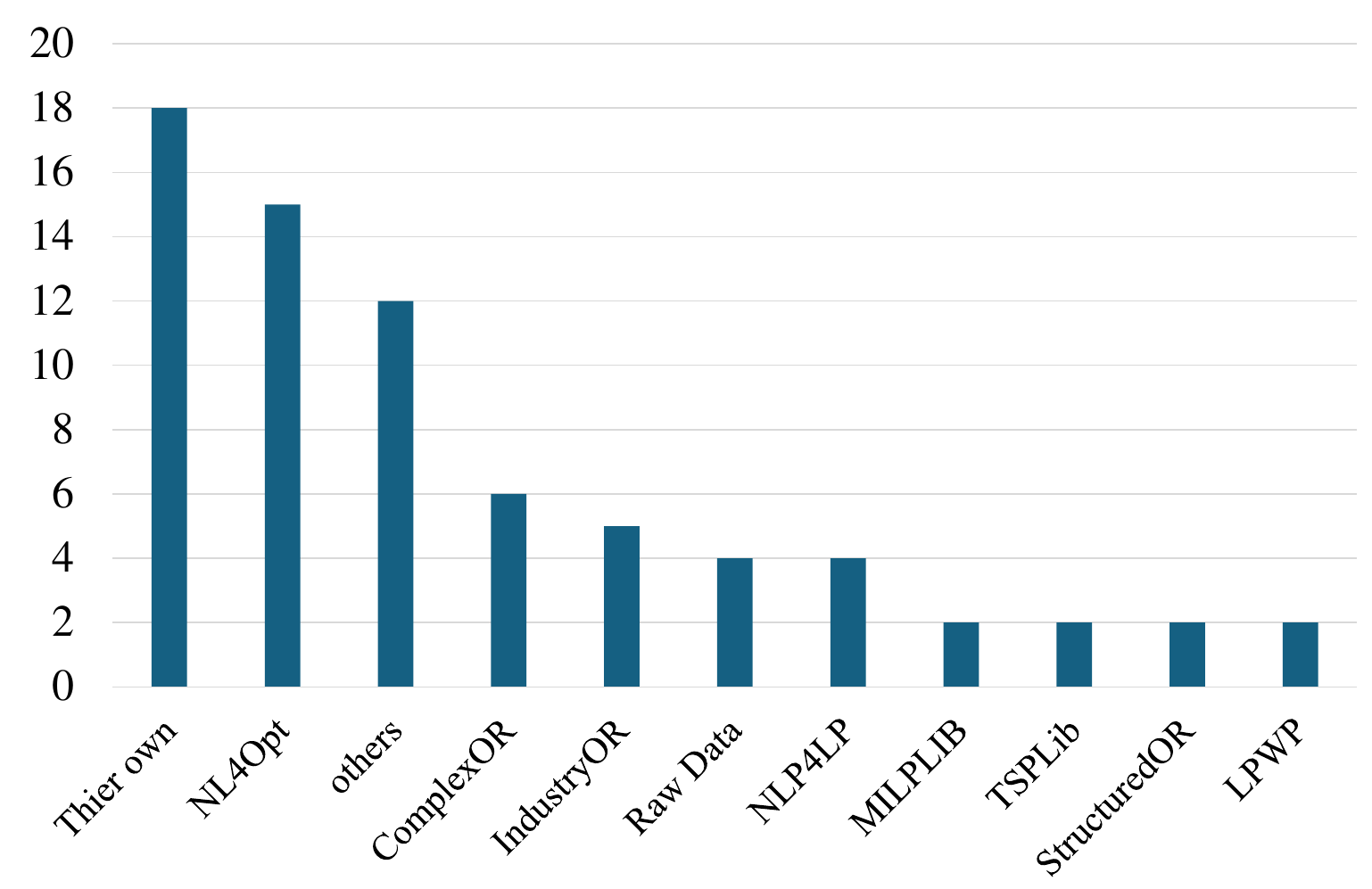}
\caption{Frequency of the datasets used in the literature, with NL4Opt being the most frequently used.}
\label{fig:RQ3}
\end{figure}

When addressing mathematical modeling, researchers tend to construct and use different benchmarks to automate the process of mathematical modeling. Studying the effect of \textbf{RQ3} demonstrates that the most common dataset is NL4Opt, as depicted in Fig.~\ref{fig:RQ3}. NL4Opt is a dataset proposed in a competition at NeurIPS 2022 to bridge \ac{or} with \ac{nlp}. Nevertheless, this dataset has its limitations regarding the complexity of the considered problems. Consequently, other datasets have been proposed, such as ComplexOR, which is the second most used dataset. Other researchers construct their own datasets for teaching \acp{llm} to formulate \ac{or} problems. The ``other'' column includes the following datasets, where each one contributes one unit in the ``other'' column: DiDi operational, StarJob, Cycle share, StructuredOR, OptBinch, ReSocratic, the Pyomo Cookbook, GSM8K, MultiArith, AQuA, BBH, Mamo.

\subsection{LLM Learning}\label{subsec:LLM training}

\begin{figure}
\centering
\includegraphics[width=0.7\linewidth]{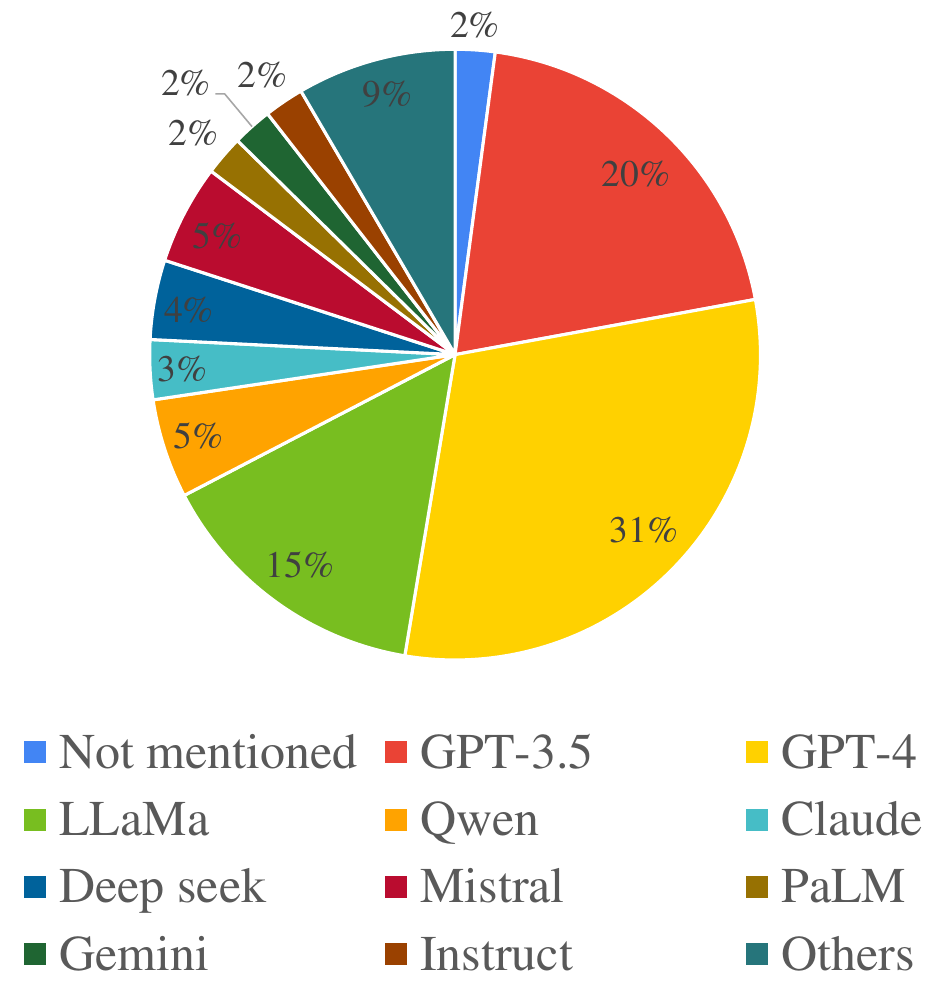}
\caption{Distribution of the \acp{llm} used in the literature, with \ac{gpt}-4 being the most frequent \ac{llm}.}
\label{fig:RQ5}
\end{figure}

Recently, there has been significant progress in the development of \acp{llm} that can automate mathematical modeling processes. Table~\ref{tab:llm_comparison} provides a brief comparison of the most commonly available \acp{llm}. After analyzing the effects of \textbf{RQ5}, as illustrated in Fig.~\ref{fig:RQ5}, 
\ac{gpt}-4 stands out as the most widely used \ac{llm} in mathematical modeling, largely due to its impressive reasoning capabilities.
\begin{table*}
\centering
\renewcommand{\arraystretch}{1.3}
\caption{Comparison of various \acp{llm}.}
\label{tab:llm_comparison}
\resizebox{0.9\linewidth}{!}{
\small{
\begin{tabular}{|>{\centering\arraybackslash}m{2.5cm}
                ||>{\centering\arraybackslash}m{2.5cm}
                |>{\centering\arraybackslash}m{4.2cm}
                |>{\centering\arraybackslash}m{3.2cm}
                |>{\centering\arraybackslash}m{2.8cm}|}
\hline
\textbf{Model} & \textbf{Creator} & \textbf{Available sizes (Parameters)} & \textbf{Architecture} & \textbf{Main domain} \\
\hline \hline
\ac{gpt}-4/o & OpenAI & Not publicly disclosed & Transformer-based & General-purpose \\
\hline
\ac{gpt}-3 & OpenAI & $175$B & Transformer-based & General-purpose \\
\hline
\ac{gpt}-3.5 & OpenAI & Not publicly disclosed & Transformer-based & General-purpose \\
\hline
DeepSeek-R1 & DeepSeek & $671$B total ($37$B active per forward pass) & \ac{moe} & Reasoning-focused \\
\hline
Llama 3.1 & Meta & $405$B & Transformer-based & General-purpose \\
\hline
Mistral Large 2 & Mistral AI & $123$B & Transformer-based & General-purpose \\
\hline
Claude 3.5 Sonnet & Anthropic & Not publicly disclosed & Transformer-based & General-purpose \\
\hline
PaLM & Google & $540$B & Transformer-based & General-purpose \\
\hline
Phi-3.5-\ac{moe} & Microsoft & $60.8$B total ($6.6$B active per forward pass) & \ac{moe} & General-purpose \\
\hline
CodeRL & Salesforce & 770M & Transformer-based & Code generation \\
\hline
\end{tabular}
}
}
\end{table*}

\begin{figure}
\centering
\includegraphics[width=1\linewidth]{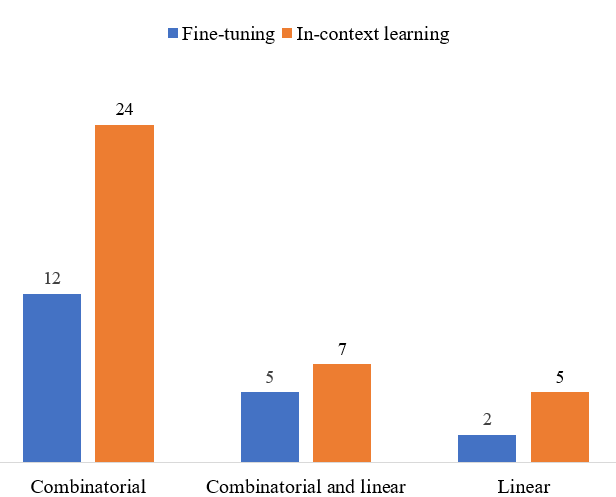}
\caption{Frequency of in-context learning and fine-tuning across different classes of optimization problems.}
\label{fig:InContextvsFine}
\end{figure}

As previously discussed, there are two effective methods for enabling \acp{llm} to adapt to new fields: In-context learning and fine-tuning. This leads us to an important investigation of the effects of~\textbf{RQ8} in the realm of mathematical modeling. As illustrated in Fig.~\ref{fig:InContextvsFine}, in-context learning is more common than fine-tuning across all problem types: Combinatorial, combinatorial and linear, and linear, with a total of $36$ research papers for in-context learning and $19$ for fine-tuning. This preference reflects the efficiency and scalability of in-context learning, particularly given the substantial computational demands of fine-tuning. Nevertheless, nearly $35\%$ of use cases still rely on fine-tuning, which remains a valid approach in light of the intricate reasoning often required in mathematical modeling. In such cases, updating model parameters through fine-tuning may be necessary to tackle complex, structured problems effectively.

\begin{figure}
\centering
\includegraphics[width=0.9\linewidth]{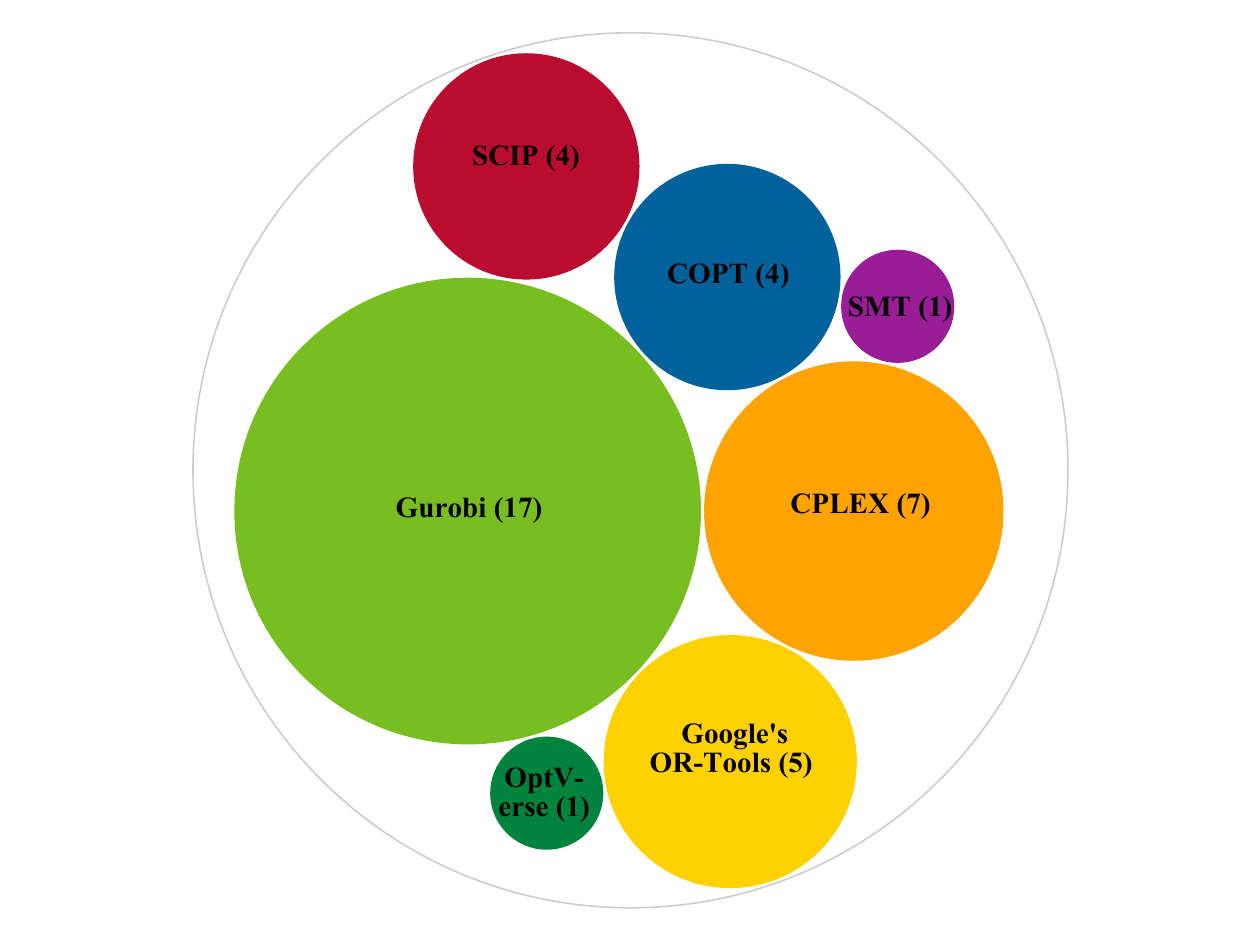}
\caption{Frequency of the solvers used in the literature, with Gurobi being the most frequently used.}
\label{fig:solver}
\end{figure}

Once a mathematical model is created with \acp{llm}, the next step is to find the objective function value considering any dataset. To facilitate this, some researchers train \acp{llm} to generate code that utilizes different solvers to solve the model. Upon studying the effect of \textbf{RQ6}, it can be noticed that not all of the proposed frameworks consider the process of solving the mathematical model and finding the objective value, despite its importance for decision-makers. As depicted in Fig.~\ref{fig:solver}, Gurobi, CPLEX, and Google's \ac{or}-Tools were the most commonly adopted solvers in the literature.
Gurobi’s cutting-edge parallel optimization algorithms and advanced heuristics make it highly responsive to dynamic and real-time changes in complex optimization problems.
CPLEX, on the other hand, uses advanced algorithms like branch-and-cut and dual simplex to efficiently solve large-scale optimization problems. Similar to CPLEX, SCIP combines cutting-edge algorithms, such as branch-and-bound, cutting planes, and branch-and-cut.
In the context of large-scale optimization problems, Google's OR-Tools can integrate seamlessly with Google Cloud. COPT and OptVerse integrate ML and AI-based techniques into their optimization processes, allowing for faster convergence and handling dynamic constraints.
Table~\ref{tab:solver_comparison} provides a comparison between the solvers used in the literature. These solvers can solve different types of problems, such as \ac{lp}, \ac{milp}, \ac{qp}, and \ac{cp} problems.
\begin{table}
\centering
\caption{Comparison of solvers used in mathematical modeling.}
\small{
\begin{tabular}{|>{\centering\arraybackslash}m{0.12\linewidth}||>{\centering\arraybackslash}m{0.17\linewidth}|>{\centering\arraybackslash}m{0.13\linewidth}|>{\centering\arraybackslash}m{0.04\linewidth}|>{\centering\arraybackslash}m{0.09\linewidth}|>{\centering\arraybackslash}m{0.04\linewidth}|>{\centering\arraybackslash}m{0.04\linewidth}|}
\hline
\textbf{Solver} & \textbf{License} & \textbf{Best for} & \textbf{\ac{lp}} & \textbf{\ac{milp}} & \textbf{\ac{qp}} & \textbf{\ac{cp}} \\
\hline \hline
Gurobi \cite{gurobi_manual12}& Commercial & \ac{milp} (fast solver) & \cmark & \cmark & \cmark & \xmark \\ \hline
CPLEX \cite{cplex_manual126}& Commercial & Business applications & \cmark & \cmark & \cmark & \xmark \\ \hline
\Ac{or}-Tools \cite{or-tools_manual}& Open-source & General optimization & \cmark & \cmark & \xmark & \cmark \\ \hline 
COPT \cite{copt_manual}& Commercial &
Large-scale optimization
& \cmark & \cmark& \cmark & \xmark \\ \hline
OptVerse \cite{optverse_manual}& Commercial & 
Dynamic optimization
& \cmark & \cmark & \cmark & \xmark \\ \hline
SCIP \cite{scip_manual}& Open-source & \Ac{milp} and \ac{cp} & \cmark & \cmark & \xmark & \cmark \\ \hline
\Ac{smt} \cite{z3_manual}& Open-source & Formulas satisfiability check  & - & - & - & \cmark \\ \hline
\end{tabular} 
}
\label{tab:solver_comparison}
\end{table}

\begin{figure}
    \centering
    \includegraphics[width=1\linewidth]{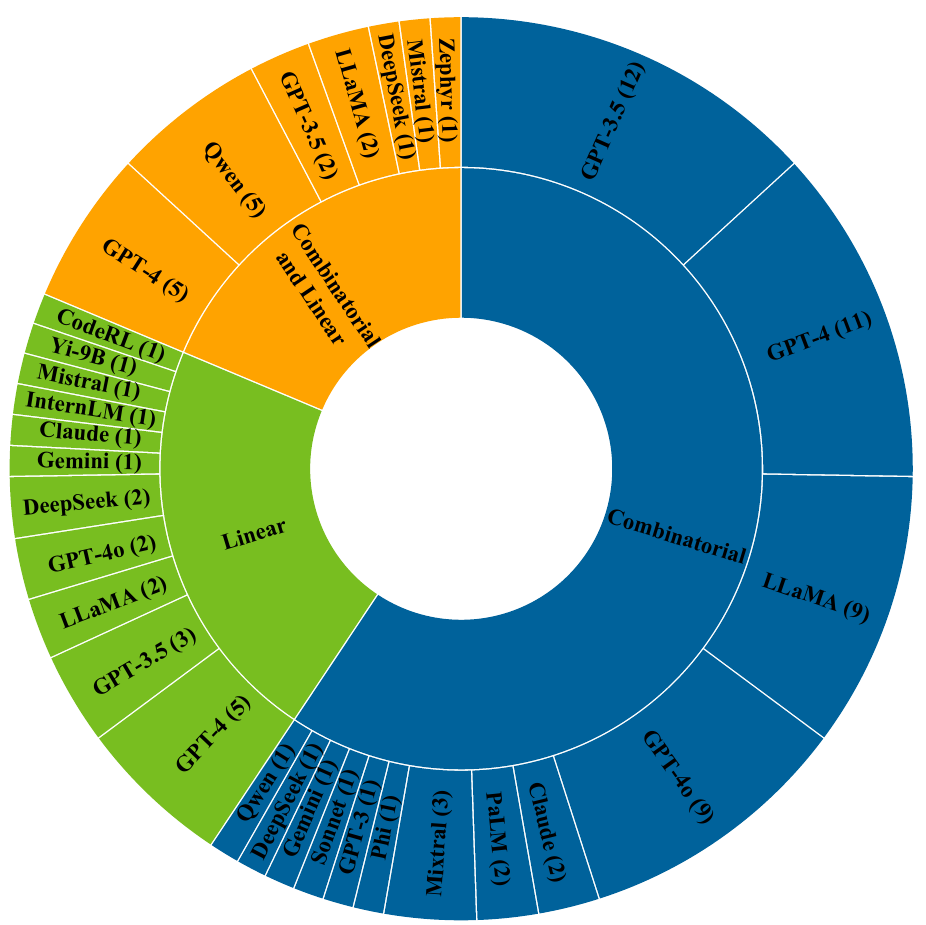}
    \caption{Frequency of the optimization classes studied and adopted \acp{llm}.}
    \label{fig:Mathimatical_LLM}
\end{figure}

The complexity of various types of mathematical modeling can differ significantly. Therefore, we aimed to examine the most common \ac{llm} for each type of mathematical models by analyzing the impact of \textbf{RQ1} and \textbf{RQ5}, as shown in Fig.~\ref{fig:Mathimatical_LLM}. It can be noticed that the capability of \ac{gpt}-4 in solving different types of mathematical modeling was extensively studied in the literature, and this is due to its multimodal capability allows users to interact with the model in more diverse and flexible ways, enhancing its applicability across different domains. Other \acp{llm} are rapidly evolving and require further investigation regarding their capabilities for various levels of complexity in mathematical modeling.

\begin{figure}
    \centering
    \includegraphics[width=0.8\linewidth]{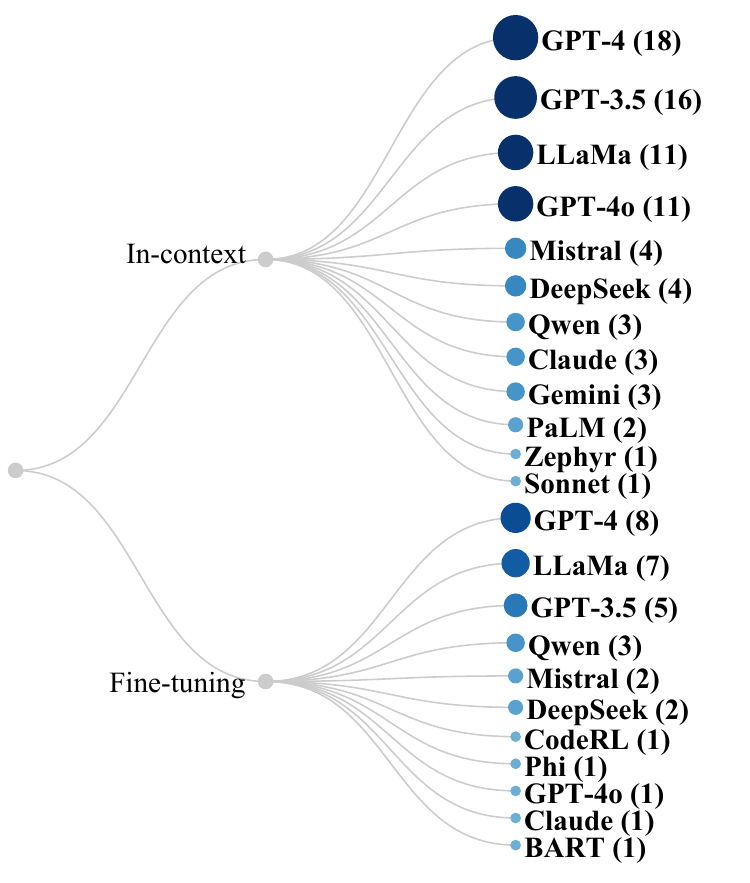}
    \caption{Frequency of in-context learning and fine-tuning across the \acp{llm} used for mathematical modeling.}
    \label{fig:LLM_learning}
\end{figure}

The difference in the complexities of mathematical models also requires different ways of dealing with these complexities by the \acp{llm}, and this can be handled by either fine-tuning or in-context learning. Thus, we have examined the frequency of in-context learning and fine-tuning across the \acp{llm} used in mathematical modeling. This investigation indicates the capabilities of \acp{llm}, determining whether they require simple in-context learning or if their reasoning abilities necessitate fine-tuning instead. From Fig.~\ref{fig:LLM_learning} it can be seen that \ac{gpt}-4 is the most frequently used model for in-context learning ($18$ occurrences), followed by \ac{gpt}-3.5 ($16$) and Llama ($11$), indicating a strong preference for these models in scenarios where adaptation is needed without modifying model parameters. Regarding fine-tuning, Llama and \ac{gpt}-4 are the most commonly \acp{llm} ($8$ occurrences). While both are widely employed in both approaches, fine-tuning appears to be applied more selectively, likely due to computational demands. Overall, the figure highlights that \ac{gpt}-4 and Llama dominate automatic mathematical modeling, with in-context learning being the more widely adopted approach across different models.

\begin{figure}
\centering
\includegraphics[width=1\linewidth]{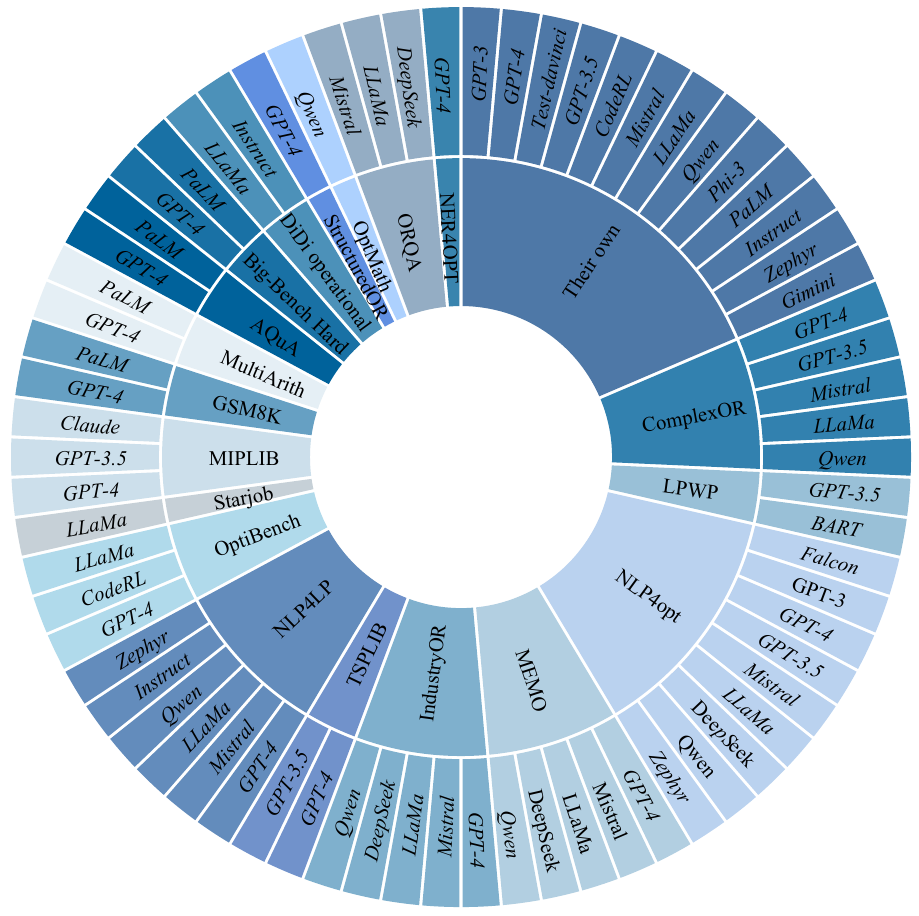}
\caption{Frequency of the datasets used and adopted \acp{llm}.}
\label{fig:Dataset_LLM}
\end{figure}

The dataset always has a direct impact on the capability of any model when approaching any given problem, regardless of the learning technique. Thus, the effect of the \textbf{RQ3} and \textbf{RQ5} was inspected. Fig.~\ref{fig:Dataset_LLM} illustrates the frequency of datasets used for training various \acp{llm}. \Ac{gpt}-4, Llama, and \ac{gpt}-3.5 are the most frequently used \acp{llm} for mathematical modeling, as they appear extensively across the literature and are trained on a diverse range of datasets. Among these datasets, NL4Opt stands out as the most commonly utilized across various \acp{llm}, suggesting its significance in optimizing learning for mathematical modeling tasks. Notably, almost all \acp{llm} have been trained, at least in part, on proprietary datasets that are not publicly available (``their own''). Additionally, certain datasets, such as IndustryOR, DiDi Operational, AQUA, and MultiArith, were tested on only one or two \acp{llm}, indicating limited adoption. This suggests the need for further investigation into their applicability, effectiveness, and potential for broader use in training mathematical models. Understanding how these datasets influence model performance could provide valuable insights into their suitability for different learning tasks.  

\begin{figure}
    \centering
    \includegraphics[width=0.9\linewidth]{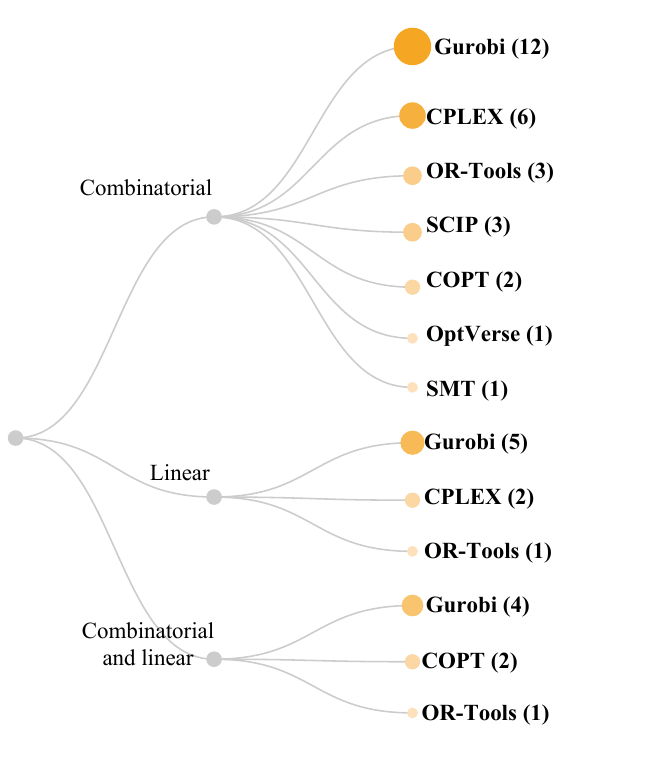}
    \caption{Frequency of the optimization classes studied and adopted solvers.}
    \label{fig:LLM_Solver}
\end{figure}

Given the variety of well-known solvers available for addressing different \ac{or} problems, we investigate the distribution of solvers utilized in LLM-generated code and analyze the categories of problems for which they are applied.
As a result, the effect of \textbf{RQ1} and \textbf{RQ6} is illustrated in Fig.~\ref{fig:LLM_Solver}. It can be noticed that Gurobi was the most used solver regardless of the mathematical problem type. This is due to its reliability and efficiency in solving optimization problems. Other solvers, such as SCIP, COPT, OptVerse, and \ac{smt} solvers, were used but less frequently. Studies that addressed both combinatorial problems adopt a wider variety of solvers, including Gurobi, CPLEX, COPT, and Google's \ac{or}-Tools, SCIP, OptVerse, and SMT. Research on both combinatorial and linear appears to be the least explored in the context of generating solver codes. It can be seen that such studies used Gurobi, COPT, and \ac{or}-Tools, but with fewer occurrences per solver.


\subsection{Model Evaluation}\label{subsec:model evaluation}

This section summarizes the evaluation metrics used to assess \ac{llm}-generated optimization formulations introduced in \textbf{RQ9}. It provides precise definitions with explicit equations for the principal measures and brief explanations for less common ones, and it reports their prevalence in the literature based on our meta-analysis. The metrics are grouped into the following categories: Solution quality, surface-form accuracy, buildability, runtime robustness, feasibility/model soundness, efficiency, search effectiveness, domain utility outcomes, multi-objective quality, mathematical fidelity, and human-centered evaluation.


\subsubsection{Solution Quality Metrics}
Solution quality metrics evaluate how good a feasible, successfully executed solution is relative to the original.

\begin{itemize}
\item \textbf{Optimality gap}: This metric measures the relative difference between the objective value produced by the \ac{llm} formulation and the known optimal solution. Let `optimal value' represent the best-known objective value for the problem, and `model's objective value' be the value obtained by solving the generated model. The optimality gap is defined as:
     \begin{equation}
        \textstyle
        \text{Optimality gap} = \frac{|\text{optimal value} - \text{model's objective value}|}{|\text{optimal value}|}.
    \end{equation}

\item \textbf{Average improvement ratio (AIR) (vs. human-designed heuristic)}:
We evaluate on $M$ benchmark problems (test instances). For each problem $m \in \{1, \ldots, M\}$, measure how far a method $s$ is from the best-known optimum using its \emph{relative gap} 
 $g_m(s)=\frac{|\text{Sol}_m(s)-\text{Opt}_m|}{|\text{Opt}_m|+\varepsilon}$,
 where $\text{Sol}_m(s)$ is the objective achieved by $s$, $\text{Opt}_m$ is the best-known optimum, and $\varepsilon>0$ is a tiny constant to avoid dividing by zero. 
 Let $h$ be the best human-designed heuristic with gap $g_m(h)$. 
 AIR is the average, across all problems, of your method’s gap \emph{relative} to the heuristic’s gap, which can be expressed as:
 \begin{equation}
 \mathrm{AIR}(s) \;=\; \frac{1}{M}\sum_{m=1}^{M} \frac{g_m(s)}{g_m(h)}.
 \end{equation}
 $\mathrm{AIR}(s)<1$ means your method is better than the heuristic on average; $=1$ means about the same; $>1$ means worse.
\end{itemize}


 \subsubsection{Surface-form Accuracy Metrics}
 Surface-form accuracy measures literal overlap with the original at the token/component level.

 \begin{itemize}
     \item \textbf{Precision}: Fraction of produced items that are correct. Let TP be the true positives and FP be the false positives. Then, 
     \begin{equation}
        \text{Precision} = \frac{\text{TP}}{\text{TP} + \text{FP}}.
    \end{equation}

    \item \textbf{Recall}: Fraction of required items that were produced. Let FN be the false negatives. Then, 
    \begin{equation}
        \text{Recall} = \frac{\text{TP}}{\text{TP} + \text{FN}}.
    \end{equation}

    \item \textbf{F1-score}: Harmonic mean of precision and recall.
    \begin{equation}
       \text{F1-score} = \frac{2 \times \text{Precision} \times \text{Recall}}{\text{Precision} + \text{Recall}}.
    \end{equation}
\end{itemize}


\subsubsection{Buildability Metrics}
 Buildability checks whether the generated formulation can be parsed and accepted before any execution.

\begin{itemize}
     \item \textbf{Compilation accuracy}: The proportion of generated formulations that pass (parsing, typing, and schema) checks. Let $\#\text{Compiled}$ be the number that compile and $\#\text{Generated}$ the total. Then,
     \begin{equation}
         \text{Compilation accuracy} = \frac{\#\text{Compiled}}{\#\text{Generated}}.
     \end{equation}
 \end{itemize}


\subsubsection{Runtime Robustness Metrics}
Runtime robustness evaluates whether compiled programs finish execution cleanly.

 \begin{itemize}
    \item \textbf{Execution rate}: Among compilable runs, the share that finish without runtime errors or timeouts. Let $\#\text{Ran}$ be runs that finished and $\#\text{Compiled}$ those that compiled. Then,
    \begin{equation}
        \text{Execution rate} = \frac{\#\text{Ran}}{\#\text{Compiled}}.
     \end{equation}

 \end{itemize}


 \subsubsection{Feasibility / Model Soundness Metrics}
 Feasibility records whether a compiled run yields a solver-feasible solution.

 \begin{itemize}
     \item \textbf{Feasibility pass rate}: The share of compiled runs that return a feasible (or optimal) status. Let $\#\text{Feasible}$ be feasible outcomes. Then, 
     \begin{equation}
         \text{Feasibility pass rate} = \frac{\#\text{Feasible}}{\#\text{Compiled}}.
     \end{equation}
 \end{itemize}


\subsubsection{Efficiency Metrics}
 Efficiency metrics measure (i) how long it takes to build and solve the \ac{llm}-generated formulation on each instance, and (ii) how fast the solution quality improves during training/search.

 \begin{itemize}
     \item \textbf{Average solving time}: Wall-clock time per instance to build and solve the generated formulation (shorter is better). 
     Let $T_j$ be the elapsed time to finish instance $j$ (including model build and solve), $t_{\max}$ a time limit (if used), and $N$ the number of instances. Then, the average solving time, $\overline{T}$, is given by:
     \begin{equation}
         \overline{T} \;=\; \frac{1}{N}\sum_{j=1}^{N} \min\!\big(T_j,\, t_{\max}\big).
     \end{equation}

     \item \textbf{Convergence performance curve (episodes vs.\ performance)}: Average performance of the solution \textit{obtained by solving the generated formulation} after each training/search episode (higher or lower is better depending on the metric). 
     Let $S_j(e)$ denote the chosen performance measure for instance $j$ after episode $e$ (e.g., $-\!$optimality gap, achievable sum-rate).
     The aggregated curve is:
    \begin{equation}
        \overline{S}(e) \;=\; \frac{1}{N}\sum_{j=1}^{N} S_j(e).
     \end{equation}
 \end{itemize}


\subsubsection{Search Effectiveness Metrics}
 Search effectiveness evaluates whether sampling multiple generations yields at least one usable formulation within \(k\) attempts.

 \begin{itemize}
 \item \textbf{Valid@k}: For each instance \(j=1,\dots,N\), generate \(k\) attempts. 
 Define \(S_{j,i}=1\) if attempt \(i\) for instance \(j\) compiles and executes to completion (i.e., yields a runnable solver run), and \(S_{j,i}=0\) otherwise. 
 Valid@k, denoted by $V_k$, is the fraction of instances for which at least one attempt succeeds:
 \begin{equation}
 V_k \;=\; \frac{1}{N}\sum_{j=1}^{N} \max_{\,i \le k}\, S_{j,i}.
 \end{equation}
 \end{itemize}


 \subsubsection{Domain Utility Outcomes}
 Domain utility reports performance in the problem’s own units (e.g., Mbps, \$, minutes).

 \begin{itemize}
     \item \textbf{Utility improvement (vs. baseline)}: How much better the \emph{solution obtained from the \ac{llm}-generated formulation} performs compared with a baseline method on the same instance. 
    Let $U_j$ be the payoff computed from the solution returned by the \ac{llm}-generated formulation for instance $j$ (in native units), 
    $U_j^{\text{base}}$ be the payoff from the baseline method on that instance, and $\varepsilon>0$ be a tiny constant to avoid division by zero. Then, 
    \begin{equation}
        \text{Utility improvement} \;=\; \frac{U_j - U_j^{\text{base}}}{|U_j^{\text{base}}| + \varepsilon}.
    \end{equation}
    For ``higher-is-better'' payoffs (e.g., throughput, revenue), positive values mean the \ac{llm}-generated formulation improved over the baseline; for ``lower-is-better'' payoffs (e.g., delay, cost), report the percent decrease or define the payoff so that larger is better.
\end{itemize}


 \subsubsection{Multi-objective Quality Metrics}
 When the generated formulation must balance several goals (e.g., cost and delay), solving it produces a set of trade-off solutions (a Pareto set). We use two metrics to judge this set: One for coverage and one for closeness to a reference front.

\begin{itemize}
    \item \textbf{Hypervolume (HV)}: How much of the objective space is covered by the Pareto set produced by the \emph{\ac{llm}-generated formulation}. A larger HV means your set spans a bigger, better region of trade-offs.
    Let $P$ be the set of solutions you obtained, $\mathrm{ND}(P)$ its nondominated subset, and $r$ a fixed reference point that is worse than all solutions (per objective). Then, 
     \begin{equation}
         \mathrm{HV}(P, r) \;=\; \lambda\!\Big( \bigcup_{p \in \mathrm{ND}(P)} [p,\, r] \Big).
     \end{equation}

    \item \textbf{Inverted generational distance (IGD)}: Shows how close your produced set is to a \textit{reference Pareto front}. A smaller IGD means your set better approximates the desired front.
     Let $Z^*$ be the reference set (true front if known, or the nondominated union across methods), and $P$ be the set from the \ac{llm}-generated formulation. Then,
     \begin{equation}
         \mathrm{IGD}(P, Z^*) \;=\; \frac{1}{|Z^*|}\sum_{z \in Z^*} \min_{p \in P} \| z - p \|_2.
     \end{equation}
 \end{itemize}


 \subsubsection{Mathematical Fidelity Metrics}
 Mathematical fidelity checks whether the LLM-generated formulation encodes the same mathematics as the original formulation.

 \begin{itemize}
     \item \textbf{Integrity gap (structure)}: Measures how different the variable–constraint incidence pattern is from the original. 
     Let $E^{\mathrm{gen}}$ and $E^{\mathrm{org}}$ be the sets of incidence edges (a variable appears in a constraint) after simple normalization (consistent naming, ordering, and sign conventions). 
     The integrity gap is the complement of the Jaccard overlap, and is given by:
     \begin{equation}
         \mathrm{IntegrityGap} \;=\; 1 \;-\; \frac{|E^{\mathrm{gen}}\cap E^{\mathrm{org}}|}{|E^{\mathrm{gen}}\cup E^{\mathrm{org}}|}.
     \end{equation}
     $0$ means identical structure; larger values indicate a worse structural match.

     \item \textbf{Semantic similarity (meaning)}: 
     Check whether the two formulations express the same relationships overall, even if written differently. 
     After normalizing both formulations (e.g., consistent variable naming, ordering, and scaling), map each to an embedding with $\phi(\cdot)$ and compute cosine similarity:
     \begin{equation}
     \small
     \mathrm{Semantic Similarity} \;=\; 
     \frac{\langle \phi(\text{form}^{\mathrm{gen}}),\,\phi(\text{form}^{\mathrm{org}})\rangle}
          {\|\phi(\text{form}^{\mathrm{gen}})\|_2\,\|\phi(\text{form}^{\mathrm{org}})\|_2}.
     \end{equation}
     As an alternative, an ``\ac{llm}-judge'' can label whether predicted elements semantically match the reference and the results can be summarized with precision, recall, and F1~\cite{liu2024variable}.

 \end{itemize}

\begin{figure}
     \centering
     \includegraphics[width=1\linewidth]{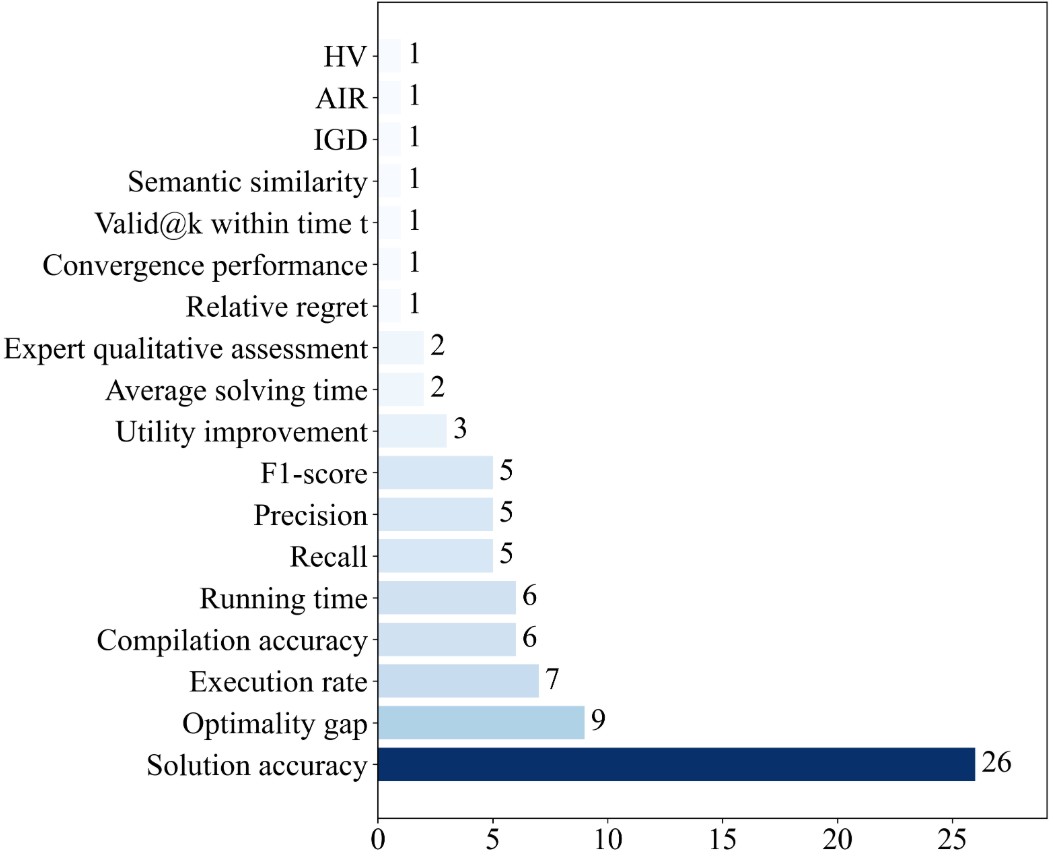}
     \caption{Frequency of the metrics used to measure the \acp{llm}' ability in formulating and solving mathematical optimization problems. Accuracy is the most used metric. }
     \label{fig:RQ7}
 \end{figure}
\begin{figure}
     \centering
     \includegraphics[width=1\linewidth]{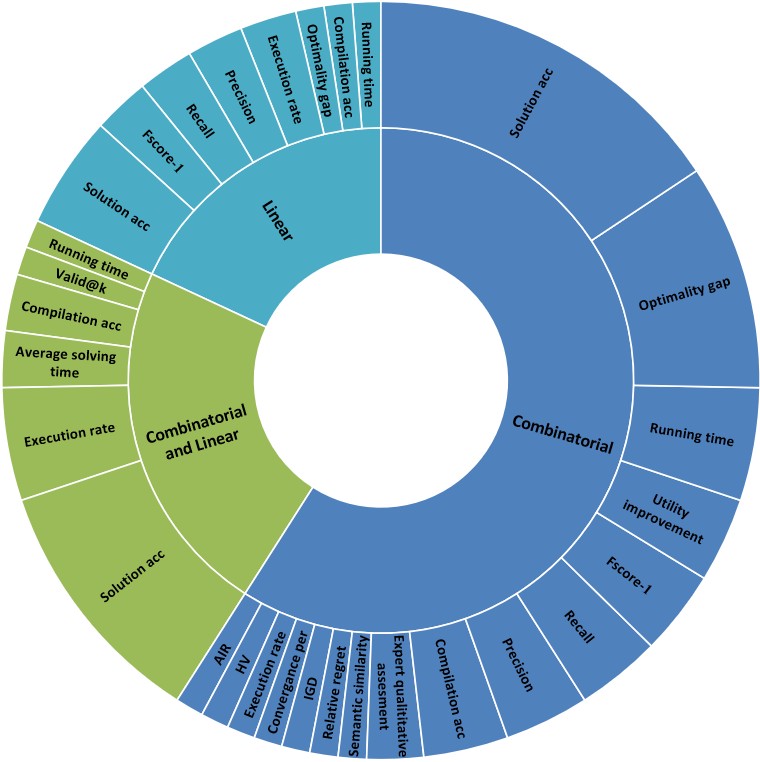}
     \caption{Distribution of the optimization classes studied and adopted evaluation metrics.}
     \label{fig:Mathematical_Metrics}
 \end{figure}

 Fig.~\ref{fig:RQ7} presents a meta-analysis of metric usage frequency, offering insights into which metrics are most frequently employed across existing studies. In addition, Fig.~\ref{fig:Mathematical_Metrics} shows how different types of mathematical modeling tasks are associated with specific evaluation metrics.

\section{Limitations and Opportunities for LLMs in Mathematical Modeling}\label{sec:discussion}

Building on the results of meta-analysis, this section highlights the key limitations and potential opportunities of using \acp{llm} for mathematical modeling. While the analysis confirms that \acp{llm} can effectively handle simple and structured tasks, it also uncovers several critical challenges, such as limited numerical reasoning, sensitivity to input complexity, and overreliance on narrow datasets, that constrain their performance in realistic scenarios. We summarize these limitations based on both empirical evidence and findings from prior studies, and we outline emerging research directions aimed at enhancing the reasoning ability, generalizability, and robustness of \acp{llm} for complex, domain-specific optimization tasks.

\subsection{Empirical Insights from Meta-Analysis}\label{subsec:empirical insights}

Our meta-analysis reveals several critical limitations, along with promising opportunities, in the current use of \acp{llm} for mathematical modeling.

A prominent limitation across the literature is the heavy reliance on a narrow set of simple datasets, typically comprising small-scale linear or combinatorial optimization problems with only a few decision variables and constraints. While these datasets offer a convenient and controlled environment for benchmarking, they fail to reflect the complexity, variability, and scale of real-world applications. This over-reliance leads to inflated performance assessments and significantly limits the generalizability of results to more realistic, domain-specific scenarios. However, this limitation presents a clear opportunity: The development and adoption of richer, more representative datasets, including non-linear, high-dimensional, and stochastic problems, would enable a more rigorous and meaningful evaluation of \acp{llm} capabilities in practical contexts.

Another key challenge is the predominant use of closed-source models, particularly those developed by commercial providers. Although these models often deliver strong performance, their widespread adoption raises concerns regarding transparency, reproducibility, and long-term sustainability within the research community. In contrast, open-source \acp{llm} remain considerably underexplored, despite offering significant advantages such as flexibility, cost-efficiency, and greater control over data privacy and deployment. This underutilization signals an important opportunity: Advancing and systematically evaluating open-source models tailored to the needs of mathematical modeling could promote more transparent, accessible, and sustainable research practices.

While in-context learning has emerged as the most commonly adopted learning strategy, primarily due to its flexibility, scalability, and low resource demands, it is not always sufficient, particularly in domain-specific applications. In such cases, fine-tuning remains crucial, as it enables models to internalize structured patterns and knowledge derived from datasets tailored to specific fields or tasks. This distinction highlights a key insight from the meta-analysis: Although in-context learning is often favored for its ease of use and broad applicability, fine-tuning plays a vital role in achieving the depth and precision required for specialized modeling tasks. This reveals a broader opportunity to more effectively align training strategies with the complexity and contextual demands of the problem, ensuring that \acp{llm} are robustly optimized for both general-purpose and domain-specific mathematical modeling.

Finally, a major shortcoming in the current literature is the absence of a unified evaluation framework for assessing \ac{llm}-generated formulations. Most studies emphasize accuracy while neglecting equally important factors such as solution feasibility, optimality, execution time, and robustness. This fragmented evaluation approach reduces the reliability, comparability, and interpretability of reported results. There is a clear need for a comprehensive framework that captures multiple performance dimensions simultaneously, enabling a more accurate and holistic assessment of model outputs in mathematical modeling tasks.

In summary, the meta-analysis reveals several key limitations in the current literature: A lack of diversity in problem types, the frequent use of overly simplistic datasets, a strong reliance on closed-source models, and the absence of standardized, multi-dimensional evaluation practices. At the same time, it points to clear directions for future progress. These include the development of more realistic and representative datasets, greater adoption of open-source \acp{llm}, targeted fine-tuning on domain-specific datasets to improve specialization, and the design of comprehensive evaluation frameworks that assess model performance across multiple critical dimensions. Together, these opportunities provide a foundation for advancing the reliability and applicability of \acp{llm} in real-world mathematical modeling.


\subsection{Limitations and Opportunities from Literature}\label{subsec:limitations}

Recent literature highlights several key limitations that constrain the effectiveness of \acp{llm} in mathematical modeling, alongside emerging research directions aimed at overcoming these challenges.

One major limitation is the difficulty \acp{llm} face in numerical reasoning. Although these models excel at language generation, they often struggle with arithmetic operations and fail to accurately interpret numerical relationships~\cite{applepaper}. This shortcoming stems from their statistical architecture and the absence of symbolic understanding. To address this, researchers have explored integrating symbolic reasoning components into language models, as well as experimenting with alternative number encoding techniques to improve numerical comprehension. In the long term, a promising direction involves designing architectures that generate structured, executable formulations, allowing external solvers to ensure correctness rather than relying solely on the model’s internal reasoning.

Another persistent challenge is the sensitivity of \acp{llm} to input length. When handling long or complex prompts, these models often lose critical context, an issue that is especially problematic in mathematical modeling tasks, which frequently require detailed and extended inputs~\cite{tokenlimitation,SameToken}. To mitigate this, prompt compression techniques and sub-word regularization strategies such as byte pair encoding (BPE) dropout have been proposed. However, these solutions are not always sufficient. A promising opportunity lies in developing models or workflows capable of decomposing large problems into smaller, manageable subproblems that preserve coherence and semantic integrity across the full task.

A more fundamental limitation is the tendency of \acp{llm} to rely on surface-level pattern matching rather than true reasoning. Rather than deducing solutions through logical inference, models often generate responses based on statistical patterns learned during training~\cite{applepaper}. This limits their reliability in tasks that require structured reasoning or complex decision-making. To address this, several prompting strategies, such as \ac{cot}, \ac{tot}, and \ac{got}, have been proposed to guide models through step-by-step reasoning. Additionally, frameworks such as \ac{rag} and \ac{rl} are being explored to enhance logical depth and factual consistency. Ongoing research is needed to strike the right balance between generative fluency and explicit reasoning, particularly for high-stakes applications.

Another important limitation is the difficulty \acp{llm} faces in filtering out irrelevant information. When exposed to prompts containing unnecessary or distracting content, models often fail to distinguish between relevant and extraneous details, resulting in incoherent or erroneous outputs~\cite{irrelevantlimitation}. While structured prompting and preprocessing can partially mitigate this issue, more robust solutions are needed. Future work could focus on developing instruction-aware architectures and relevance-guided processing mechanisms that help models prioritize essential information during inference.

In summary, the literature identifies a range of limitations affecting the use of \acp{llm} in mathematical modeling, from weaknesses in numerical reasoning and input length sensitivity to shallow logic and poor relevance filtering. However, it also points to a growing body of work focused on integrating symbolic reasoning, improving prompting strategies, incorporating retrieval mechanisms, and coordinating specialized models. These directions offer a valuable path forward for enhancing the robustness, adaptability, and domain-specific effectiveness of \acp{llm} in mathematical and decision-oriented applications.


\subsection{Answering Research Questions}\label{subsec:RQ answers}

Based on the preceding meta-analysis and literature review, we now address the four research questions that guided this study.

\textbf{What are the most commonly used \acp{llm} for generating mathematical models?}  
Our findings indicate that \ac{gpt}-based models, particularly those developed by OpenAI, are the most widely adopted in the current literature. These models dominate both in frequency of use and reported effectiveness across various mathematical modeling tasks. Their popularity is largely attributed to their strong general performance, accessibility via APIs, and integration into existing optimization workflows. In contrast, open-source models such as Llama and DeepSeek are significantly underutilized, despite offering advantages in customization, transparency, and data governance.

\textbf{To what extent are \acp{llm} capable of generating mathematical models?}  
\Acp{llm} demonstrate promising capabilities in generating mathematical models, especially for well-structured, small-scale, and general-purpose problems. They are often successful in producing syntactically correct formulations for linear and combinatorial optimization tasks. However, their performance is notably limited when applied to domain-specific or complex problems that involve non-linearity, uncertainty, or large-scale constraints. These limitations highlight the need for improvements in reasoning capabilities, domain adaptation, and structured problem representation to ensure accurate and context-aware model generation.

\textbf{What is the best approach to utilize \acp{llm} for generating and formulating mathematical models?}  
Current evidence suggests that the most effective approach involves combining well-structured, representative datasets with in-context learning techniques. These strategies help guide the model toward producing coherent and relevant outputs. Multi-agent architectures, where different models collaborate or specialize in distinct subtasks, also show potential in enhancing the formulation process. However, techniques such as fine-tuning and \ac{rag}, though conceptually promising, have not been sufficiently explored in this context. Further empirical evaluation is needed to determine their comparative effectiveness.

\textbf{What are the key challenges and future directions for improving \ac{llm}-based mathematical modeling?}  
The key challenges identified include limited numerical reasoning, sensitivity to input complexity, reliance on pattern matching rather than logical inference, and the inability to filter irrelevant information. In addition, the lack of diverse datasets, open-source model utilization, and comprehensive evaluation frameworks limits progress in the field. 

\section{Experimental Design and Results}\label{sec:experiments_section}

This section presents the experimental design and empirical findings of our study, which investigates the ability of \acp{llm} to generate mathematical optimization formulations from natural language descriptions.
We implemented an experiment designed to evaluate the performance of language models under both zero-shot and few-shot prompting conditions. We analyze performance trends, identify observed limitations, and highlight potential opportunities. These findings lay the foundation for the subsequent sections, which highlight key challenges and propose directions for future research.

 
\subsection{Experimental Design}\label{sec:experimental_setup}

This section outlines the core components of the experiment, including the design of optimization problems, prompting strategies, the selected \acp{llm} and their configurations, and the evaluation metrics.


\subsubsection{Problems Design}\label{sec:problems_design}

Despite the recent surge in applying \acp{llm} to a wide range of reasoning and problem-solving tasks in \ac{or}, the availability of datasets specifically designed to test the ability of \acp{llm} in mathematical modeling in the computer networks domain remains a critical gap.

Existing datasets primarily address general mathematical modeling problems or simple scheduling tasks, thus neglecting the essential domain-specific datasets. As each domain has its own characteristics, terms, and constraints that might not necessarily exist in other domains, there is a need for a domain-specific dataset.

To address this gap, we present ten optimization problems specifically designed to evaluate \ac{llm}'s capabilities in mathematical modeling in the computer network domain, following the structure of the most recent challenging dataset, ComplexOR~\cite{complexORdataset}. Appendix~\ref{app} provides a description of each problem. 




\subsubsection{Prompting Strategies}\label{sec:prompt_strategies}

To guide the \acp{llm} in translating natural language problem descriptions into formal mathematical formulations, we developed three distinct prompt templates. Each prompt reflects a unique reasoning strategy and was carefully designed to elicit structured and consistent outputs from the models. Prompts were presented alongside the problem description and sample data as part of the input. The designed prompts are shown below.

    \begin{tcolorbox}[colback=blue!9!white, colframe=blue!60!black, title = Prompt 1: Act-As-Expert]
    You are an expert in mathematical optimization. Solve the problem by clearly defining:
    \begin{enumerate}
        \item Decision variables.
        \item Objective function.
        \item Constraints.  
    \end{enumerate}

    \textbf{Finally:} Output the complete mathematical formulation. Do not include any explanations.
    \end{tcolorbox}
    
    \begin{tcolorbox}[colback=green!5!white, colframe=green!50!black, title = Prompt 2: \ac{cot}]
    Solve the problem using a logical step-by-step approach:
    \begin{enumerate}
    \item Define the decision variables.
    \item Formulate the objective function.
    \item Specify the constraints.  
    \end{enumerate}
    
    \textbf{Finally:} Provide the complete mathematical formulation. Do not include any explanations.
    \end{tcolorbox}
    
    \begin{tcolorbox}[colback=orange!5!white, colframe=orange!85!black, title = Prompt 3: Self-Consistency]
    Generate three independent mathematical formulations of the same optimization problem:
    \begin{itemize}
    \item Each should include decision variables, an objective function, and constraints.
    \item Internally evaluate and compare them, then choose the best formulation.  
    \end{itemize}
    
    \textbf{Finally:} Output only the final selected formulation. Do not include any explanations.
    \end{tcolorbox}


\subsubsection{\acp{llm} and Configurations}

We evaluated two state-of-the-art \acp{llm}: DeepSeek Math (open-source) and \ac{gpt}-4o (closed-source). These models were selected based on their demonstrated strengths in mathematical reasoning and optimization tasks. To ensure a fair comparison, both models were configured with the same temperature and maximum token limit. Table~\ref{table: model_details} summarizes the key settings.
\begin{table}
\centering
    \caption{Summary of model details and configurations.}\label{table: model_details}
    \centering
    \begin{tabular}{|>{\centering\arraybackslash}m{0.4in}||>{\centering\arraybackslash}m{0.9in}|>{\centering\arraybackslash}m{0.4in}|>{\centering\arraybackslash}m{0.4in}|>{\centering\arraybackslash}m{0.4in}|}
    \hline
    \centering \textbf{Model}  & \centering\textbf{Model Version} & \centering\textbf{Temp} & \centering\textbf{Max Tokens} & \centering\textbf{Source} \tabularnewline
    \hline
    \hline
    \centering DeepSeek Math  & \centering DeepSeek-Math-7b-Instruct & \centering $0.1$ & \centering $900$ & \centering Open \tabularnewline
    \hline 
    \centering \ac{gpt}-4o  & \centering \ac{gpt}-4o & \centering $0.1$ & \centering $900$ & \centering Closed\tabularnewline
    \hline
    \end{tabular}
\end{table}


\subsubsection{Evaluation Metrics}

In the experiment, we evaluated the quality of the \ac{llm}-generated optimization formulations using three key metrics: Optimality gap, token-level F1 score, and compilation accuracy. These metrics, previously defined in Section~\ref{meatanalysis}, are selected to capture different aspects of correctness. The optimality gap quantifies the deviation between the objective value obtained from the generated formulation and the known optimal value, providing a measure of numerical accuracy. The token-level F1 score assesses symbolic similarity by comparing tokenized expressions in the generated and reference formulations, capturing structural correctness. The compilation accuracy metric verifies whether the generated code can be parsed and executed without errors, serving as a prerequisite for downstream evaluations. Together, these metrics provide a multidimensional assessment of model performance across functional, structural, and syntactic levels.


\subsection{Experimental Results}
\label{sec:experiment_combined}

This experiment investigates the ability of \acp{llm} to translate natural language descriptions into formal mathematical optimization formulations under both zero-shot and in-context learning settings. The goal is to evaluate how prompting strategies and contextual demonstrations influence formulation quality across a diverse set of optimization problems.

As outlined in the experimental setup, we use ten networking optimization problems of varying complexity, covering tasks such as routing, resource allocation, and scheduling. Each problem is paired with three prompting strategies: Act-as-expert, \ac{cot}, and self-consistency, designed to guide model reasoning in different ways. Two state-of-the-art \acp{llm}, DeepSeek Math and \ac{gpt}-4o, are evaluated using identical configurations to ensure a fair comparison. We consider three prompting configurations: Zero-shot (problem and prompt only), one-shot (a single solved example), and two-shot (two solved examples), with examples drawn from the ComplexOR dataset~\cite{complexORdataset}. This unified setup allows us to assess the models' performance across multiple levels of contextual support using a consistent experimental protocol.

\begin{figure}
\centering
\includegraphics[width=1\linewidth]{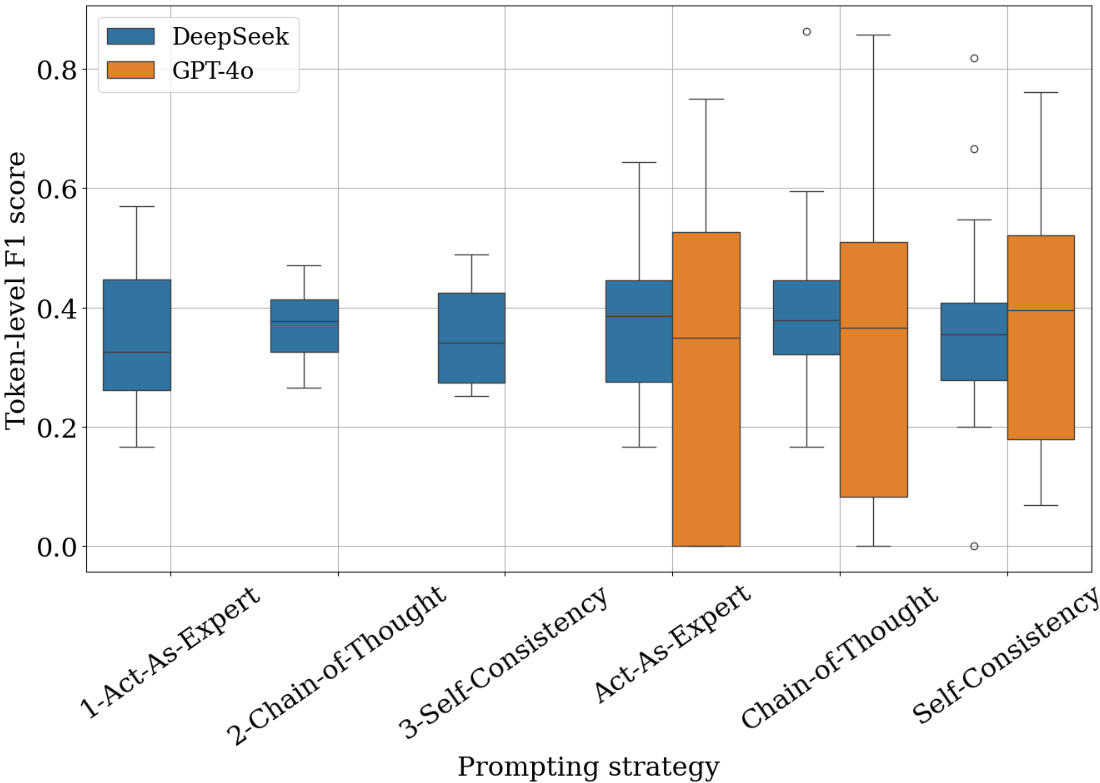}
\caption{Token-level F1 score distribution across different prompt shot settings for \ac{gpt}-4o and DeepSeek-Math-7B-Instruct. \ac{gpt}-4o exhibits consistently higher median scores and lower variance.}
\label{fig:tokenf1_boxplot}
\end{figure}

To evaluate the generated formulations, we use three complementary metrics: Optimality gap, token-level F1 score, and compilation accuracy, each capturing a distinct aspect of performance; numerical accuracy, symbolic structure, and syntactic soundness, respectively.  Tables~\ref{table:IncontextLearning-DeepSeek} and~\ref{table:IncontextLearning-GPT4} report the optimality gaps for DeepSeek Math and \ac{gpt}-4o, respectively, across all prompting strategies and in-context configurations. Tables~\ref{table:tokenF1_DeepSeek} and~\ref{table:tokenF1_GPT4} present the token-level F1 scores for both models, Fig.~\ref{fig:tokenf1_boxplot} visualizes the distribution of token-level F1 scores, and Table~\ref{tab:syntactic_validity} summarizes the compilation accuracy comparisons.

\begin{table*}
\centering
\caption{Optimality gaps for the DeepSeek model across different in-context learning settings and prompting strategies.}
\label{table:IncontextLearning-DeepSeek}
\small{
\begin{tabular}{|c||c|c|c||c|c|c||c|c|c|}
\hline
\multirow{2}{*}{\textbf{Problem}} 
& \multicolumn{3}{c||}{\textbf{Act-As-Expert}} 
& \multicolumn{3}{c||}{\textbf{\Ac{cot}}} 
& \multicolumn{3}{c|}{\textbf{Self-Consistency}} \\
\cline{2-10}
& \textbf{Zero-shot} & \textbf{One-shot} & \textbf{Two-shot} 
& \textbf{Zero-shot} & \textbf{One-shot} & \textbf{Two-shot} 
& \textbf{Zero-shot} & \textbf{One-shot} & \textbf{Two-shot} \\
\hline \hline
\textbf{P1}  & $0$    & $0$    & $0$    & $0$    &$ 0.57$ & $0$     & $0.57 $& $0$    & $0$ \\\hline
\textbf{P2}  & $0$    & $0$    & $0$    & $0$    & $0$    & $0$     & $0$    & $0$    & $0$ \\\hline
\textbf{P3}  & $1$    & $0.25$ & $0.5 $ & $0.25$ & $0.25$ & $0.75 $ & $0.97$ & $1$    & $1$ \\\hline
\textbf{P4}  & $1$    & 0.14 & $1$    & $1$    & $1$    & $0$     & $1$    & $1$    & $1$ \\\hline
\textbf{P5}  & $0.13$ & $0$    & $0.13$ &$ 0.1$  &$ 0.98$ &$ 0.13 $ & $0.23$ & $0.98$ & $0.13$ \\\hline
\textbf{P6}  & $0$    & $0$    & $0$    & $0$    &$ 0$    &$ 0$     & $0$    & $0$    & $0 $\\\hline
\textbf{P7}  & 1.82 & $0.64$ & $1.82$ &$ 4.64$ &$ 1.66$ & $6.36 $ & $1.82 $& $1 $   & $0.12$ \\\hline
\textbf{P8}  & $0.45$ & $0$    & $0$    & $0$    & $0.1$  & $0$     & $0.63$ & $0.1$  & $0$ \\\hline
\textbf{P9}  & $1 $   & $1$    & $1$    & $1$    & $1$    & $0.06$  & $1  $  &$ 1$    & $1$ \\\hline
\textbf{P10} &$ 0$    & $0 $   & $0 $   &$ 0$    & $0 $   & $0 $    & $0  $  & $0  $  &$ 0$ \\
\hline
\end{tabular}
}
\end{table*}

\vspace{1em}

\begin{table*}
\centering
\caption{Optimality gaps for the \ac{gpt}-4o model across different in-context learning settings and prompting strategies.}
\label{table:IncontextLearning-GPT4}
\small{
\begin{tabular}{|c||c|c|c||c|c|c||c|c|c|}
\hline
\multirow{2}{*}{\textbf{Problem}} 
& \multicolumn{3}{c||}{\textbf{Act-As-Expert}} 
& \multicolumn{3}{c||}{\textbf{\Ac{cot}}} 
& \multicolumn{3}{c|}{\textbf{Self-Consistency}} \\
\cline{2-10}
& \textbf{Zero-shot} & \textbf{One-shot} & \textbf{Two-shot} 
& \textbf{Zero-shot} & \textbf{One-shot} & \textbf{Two-shot} 
& \textbf{Zero-shot} & \textbf{One-shot} & \textbf{Two-shot} \\
\hline \hline
\textbf{P1}  & $0$ & $0.04$ & $0.04$ & $0$ & $0.04$ & $0.04$ & $0$ & $0.04$ &$ 0.04$ \\\hline
\textbf{P2}  & $0$ & $0$ & $0$ & $0$ & $0$ & $0$ & $0$ & $0$ & $0$ \\\hline
\textbf{P3}  &$ 0.25$ & $0.25$ & $0.25$ & $0.75$ & $0.25 $&$ 0.25$ &$ 0.25$ & $0.25 $& $0.25$ \\\hline
\textbf{P4}  & $0$ & $0$ & $0$ & $0$ & $0$ & $0$ & $0$ & $0$ & $0$ \\\hline
\textbf{P5}  & $0.13 $& $0$ & $0$ & $0.13$ & $0$ & $0$ & $0.13$ & $0$ & $0$ \\\hline
\textbf{P6}  & $0$ & $0$ & $0$ & $0$ & $0$ & $0$ & $0$ & $0$ & $0$ \\\hline
\textbf{P7}  & $0.18$ & $2.5$ & $2.5$ & $0.18$ & $2.5$ & $2.5 $&$ 0.18 $& $2.5 $& $2.5$ \\\hline
\textbf{P8}  & $0$ & $0$ & $0$ & $0$ & $0$ & $0$ & $0$ & $0$ & $0$ \\\hline
\textbf{P9}  & $1$ & $1$ & $1$ & $1$ & $1$ & $1$ & $1$ & $1$ & $1$ \\\hline
\textbf{P10} & $0$ & $0$ & $0$ & $0$ & $0$ & $0$ & $0$ & $0$ & $0$ \\
\hline
\end{tabular}
}
\end{table*}

\begin{table*}
\centering
\caption{Token-level F1 scores for DeepSeek-Math-7B-Instruct across prompting strategies and in-context settings.}
\label{table:tokenF1_DeepSeek}
\small{
\begin{tabular}{|c||c|c|c||c|c|c||c|c|c|}
\hline
\multirow{2}{*}{\textbf{Problem}} 
& \multicolumn{3}{c||}{\textbf{Act-as-Expert}} 
& \multicolumn{3}{c||}{\textbf{\Ac{cot}}} 
& \multicolumn{3}{c|}{\textbf{Self-Consistency}} \\
\cline{2-10}
& \textbf{Zero-shot} & \textbf{One-shot} & \textbf{Two-shot} 
& \textbf{Zero-shot} & \textbf{One-shot} & \textbf{Two-shot} 
& \textbf{Zero-shot} & \textbf{One-shot} & \textbf{Two-shot} \\
\hline \hline
\textbf{P1} &$ 0.41$ & $0.49$ & $0.53$ &$ 0.36$ & $0.43$ & $0.42$ & $0.29$ & $0.13$ & $0.33$ \\\hline
\textbf{P2} & $0.42$ & $0.52$ & $0.44 $& $0.40$ & $0.42$ & $0.40$ & $0.38$ & $0.37$ & $0.41$ \\\hline
\textbf{P3} & $0.22$ & $0.27$ & $0.22$ & $0.18$ & $0.23 $& $0.20$ & $0.17$ & $0.14$ & $0.20$ \\\hline
\textbf{P4} & $0.20$ & $0.28$ & $0.20$ & $0.15$ & $0.25$ & $0.17$ & $0.19$ & $0.00$ & $0.16$ \\\hline
\textbf{P5} & $0.37 $& $0.41$ & $0.46$ & $0.39$ & $0.41 $& $0.42$ & $0.39$ & $0.41$ & $0.45 $\\\hline
\textbf{P6} & $0.49$ & $0.51$ &$ 0.53$ & $0.48$ & $0.51$ & $0.52 $& $0.47$ & $0.50$ & $0.51 $\\\hline
\textbf{P7} & $0.53$ & $0.59$ & $0.58$ & $0.53$ & $0.58 $& $0.60$ & $0.51$ & $0.54$ & $0.59$ \\\hline
\textbf{P8} & $0.38$ & $0.42$ & $0.41$ & $0.40 $& $0.41 $& $0.40$ & $0.41$ & $0.39$ & $0.40 $\\\hline
\textbf{P9} & $0.76$ & $0.81$ & $0.83$ & $0.73 $& $0.86$ & $0.81$ & $0.68$ & $0.78$ & $0.82$ \\\hline
\textbf{P10} & $0.59$ & $0.62$ & $0.60$ & $0.60 $& $0.62$ & $0.61$ & $0.59$ & $0.82$ &$ 0.62$ \\
\hline
\end{tabular}
}
\end{table*}

\begin{table*}
\centering
\caption{Token-level F1 scores for \ac{gpt}-4o across prompting strategies and in-context settings.}
\label{table:tokenF1_GPT4}
\small{
\begin{tabular}{|c||c|c|c||c|c|c||c|c|c|}
\hline
\multirow{2}{*}{\textbf{Problem}} 
& \multicolumn{3}{c||}{\textbf{Act-As-Expert}} 
& \multicolumn{3}{c||}{\textbf{\Ac{cot}}} 
& \multicolumn{3}{c|}{\textbf{Self-Consistency}} \\
\cline{2-10}
& \textbf{Zero-shot} & \textbf{One-shot} & \textbf{Two-shot} 
& \textbf{Zero-shot} & \textbf{One-shot} & \textbf{Two-shot} 
& \textbf{Zero-shot} & \textbf{One-shot} & \textbf{Two-shot} \\
\hline \hline
\textbf{P1} & $0.50 $& $0.59$ & $0.56$ & $0.53$ & $0.64$ & $0.60$ & $0.51$ & $0.55$ & $0.58 $\\\hline
\textbf{P2} & $0.53$ & $0.58$ & $0.57$ & $0.55$ & $0.63$ & $0.60$ & $0.54$ & $0.62$ & $0.61$ \\\hline
\textbf{P3} & $0.25 $& $0.32$ & $0.30 $& $0.23$ & $0.29$ & $0.27$ & $0.21$ & $0.25$ & $0.28$ \\\hline
\textbf{P4} & $0.28$ &$ 0.35$ &$ 0.30$ & $0.25$ & $0.35 $& 0.29 & $0.26$ &$ 0.32 $& $0.31$ \\\hline
\textbf{P5} & $0.44$ & $0.48$ & $0.46$ & $0.45$ & $0.49$ & $0.47$ & $0.43$ & $0.47$ & $0.47 $\\\hline
\textbf{P6} & $0.54$ & $0.56$ &$ 0.55$ &$ 0.52$ & $0.55$ & $0.54$ & $0.52$ &$ 0.54 $& $0.53 $\\\hline
\textbf{P7} & $0.62$ & $0.65$ &$ 0.63$ & $0.60$ &$ 0.64$ & $0.62$ & $0.60$ & $0.62$ &$ 0.61$ \\\hline
\textbf{P8} & $0.44$ & $0.49$ & $0.46$ & $0.45$ &$ 0.49$ & $0.47$ & $0.44$ &$ 0.48$ &$ 0.47$ \\\hline
\textbf{P9} & $0.73$ &$ 0.76$ & $0.75 $& $0.71$ &$ 0.76$ & 0.75 & $0.69$ & $0.75$ & $0.74$ \\\hline
\textbf{P10} & $0.59$ &$ 0.61$ &$ 0.60$ & $0.58$ &$ 0.61$ & $0.60$ & 0.57 & $0.60 $& $0.59$ \\
\hline
\end{tabular}
}
\end{table*}

\begin{table*}
\centering
\caption{Compilation accuracy comparison between \ac{gpt}-4o and DeepSeek across prompting strategies.}
\label{tab:syntactic_validity}
\renewcommand{\arraystretch}{1.3}

\small{
\begin{tabular}{|>{\centering\arraybackslash}m{0.15\linewidth}
                ||>{\centering\arraybackslash}m{0.15\linewidth}
                |>{\centering\arraybackslash}m{0.15\linewidth}
                |>{\centering\arraybackslash}m{0.15\linewidth}|}
\hline
\textbf{Prompt Style} & \textbf{\ac{gpt}-4o (Mean)} & \textbf{DeepSeek (Mean)} & \textbf{Difference (\%)} \\
\hline \hline
Act-As-Expert      & $1.00$ & $0.95 $& +$5.0\%$ \\\hline
\Ac{cot}   & $0.95$ &$ 0.75$ & $+26.7\%$ \\\hline
Self-Consistency   & $0.85$ & $0.70$ & $+21.4\%$ \\\hline
Overall Average    & $0.93$ &$ 0.80 $& $+16.7\%$ \\
\hline
\end{tabular}
}
\end{table*}


\subsubsection{Results Analysis}

The results reveal mixed and often inconsistent patterns regarding the effect of in-context learning on model performance. However, a consistent trend emerges in the numerical evaluation: \ac{gpt}-4o demonstrates superior performance in terms of optimality gap, achieving significantly lower values than DeepSeek Math across nearly all problem instances and prompting configurations, as shown in Tables~\ref{table:IncontextLearning-DeepSeek} and~\ref{table:IncontextLearning-GPT4}. \ac{gpt}-4o attains near-zero gaps for simpler problems such as P1, P2, P6, P8, and P10, and maintains stable accuracy even on more complex tasks. In contrast, DeepSeek Math exhibits greater variability and more frequent degradation in performance as problem complexity increases. Although some improvements are observed under one-shot prompting, particularly with the act-as-expert strategy, these gains are inconsistent. These findings suggest that \ac{gpt}-4o benefits from stronger internal reasoning capabilities and is less dependent on contextual examples.

In terms of symbolic fidelity, token-level F1 scores, as shown in Tables~\ref{table:tokenF1_DeepSeek} and~\ref{table:tokenF1_GPT4}, highlight \ac{gpt}-4o’s clear advantage in generating formulations that are structurally and semantically aligned with the reference solutions. \ac{gpt}-4o outperforms DeepSeek Math across all prompting strategies and context settings, with particularly notable gains under \ac{cot} and self-consistency prompts. This suggests that \ac{gpt}-4o is more effective at capturing the symbolic structure of optimization problems, even when the surface form or phrasing varies. Fig.~\ref{fig:tokenf1_boxplot} further illustrates this advantage, showing \ac{gpt}-4o’s consistently higher median F1 scores and lower variance, indicating more stable and reliable symbolic output across varied scenarios.

Finally, compilation accuracy results, as shown in Table~\ref{tab:syntactic_validity}, demonstrate that \ac{gpt}-4o is also more reliable in generating executable Python code. It achieves a $16.7\%$ higher average compilation accuracy compared to DeepSeek Math and performs especially well under reasoning-intensive prompts. \ac{gpt}-4o reaches perfect execution under the act-as-expert prompt and maintains strong results across other configurations. In contrast, DeepSeek Math struggles under \ac{cot} and self-consistency settings, frequently producing code that fails to compile or execute. These findings reinforce \ac{gpt}-4o’s robustness not only in numerical and symbolic correctness but also in producing syntactically sound and functionally usable optimization models.

\subsubsection{Limitations}

Despite the strengths observed in \ac{gpt}-4o and the structured experimental setup, several limitations emerge from this study. First, prompting strategies are not universally effective. While the act-as-expert prompt often performs well, especially for \ac{gpt}-4o, no strategy consistently delivers superior results across all problems and models. DeepSeek Math, in particular, is highly sensitive to prompt phrasing and structure, resulting in unstable outputs and inconsistent gains from in-context learning.

Second, model behavior remains unpredictable in complex problem scenarios. Both models struggle with tasks that require implicit constraint modeling or multiple interdependent variables, most notably P3, P4, P7, and P9. Even when examples are provided through one-shot or two-shot prompting, performance on these complex problems often fails to improve and, in some cases, degrades. This indicates that current models still lack the deep abstraction and generalization capabilities needed for complex optimization formulations.

Third, the evaluation framework itself poses limitations. Although the optimality gap, token-level F1 score, and compilation accuracy offer complementary perspectives, they do not explain the precise source of errors. For instance, a poor optimality gap could result from incorrect decision variable definitions, misaligned constraints, or an erroneous objective function, but these components are not individually assessed. This lack of granularity limits our ability to diagnose model failures or guide fine-grained improvements in prompt design and model architecture.

\subsubsection{Opportunities}

Despite the challenges identified, the experimental findings point to several opportunities for advancing \ac{llm} performance in mathematical optimization formulation. First, the relative success of structured prompts such as act-as-expert suggests that better prompting strategies can improve model reasoning. However, the inconsistency across problems indicates that a one-size-fits-all approach is insufficient. This opens the door for adaptive prompting methods that can adjust prompt structure based on the characteristics of the problem, such as complexity, domain, or variable types.

Second, the performance gap observed between \ac{gpt}-4o and DeepSeek Math may stem less from differences in model scale and more from disparities in training objectives, instruction tuning, or domain alignment. Rather than assuming that larger models inherently perform better, recent work has shown that small language models (SLMs) can be highly effective in structured, agent-based systems due to their modularity, controllability, and efficiency~\cite{SMLs}. This highlights the importance of training strategies that prioritize task decomposition, clarity of representation, and domain-specific supervision. Developing datasets that explicitly annotate decision variables, constraints, and objective functions, with varied representations and complexity levels, may significantly improve both large and small models’ ability to learn the structure of optimization problems and generalize across domains.

Third, improving the structure of problem inputs presents another promising direction. Presenting tasks in a modular or semi-formal format, explicitly separating components such as objectives, variables, and constraints, could reduce ambiguity and help models process complex formulations more effectively. Such representations also make it easier to apply multi-agent or modular model strategies, where different \acp{llm} specialize in generating different parts of the formulation.

Finally, our results underscore the need for more granular evaluation methods. While current metrics provide useful high-level signals, they do not explain where models fail or how to improve them. Moving toward component-level evaluation that separately scores variable definitions, constraint correctness, and objective alignment could provide more actionable feedback for both prompt design and model development. These future directions, taken together, represent a promising path toward more accurate, interpretable, and robust use of \acp{llm} in mathematical optimization modeling.

Building on the combined insights from our meta-analysis of the existing literature and the empirical findings from both zero-shot and in-context learning experiments, the following two sections outline future research directions centered on two key areas: (i) Advancing \ac{llm} learning and mathematical reasoning capabilities, and (ii) improving the understanding and diagnosis of \ac{llm} limitations in the context of mathematical optimization modeling.


\section{Enhancing LLM Learning and Mathematical Reasoning Capabilities}\label{sec:FutureDirection1}

In light of the limitations uncovered through our meta-analysis and experiments, it is clear that current \acp{llm} face significant challenges in reliably formulating mathematical optimization problems. These challenges include sensitivity to prompt phrasing, limited generalization to structurally complex tasks, and inconsistent performance when confronted with unfamiliar or varied problem types. To address these limitations, this section explores several promising directions for improving how \acp{llm} learn and reason in mathematical contexts. Specifically, we focus on the development of structured training datasets, the adoption of modular or collaborative model architectures, the integration of \ac{rag} pipelines, and the refinement of prompting strategies. Collectively, these approaches aim to enhance the robustness, adaptability, and reasoning depth of \acp{llm} in formulating optimization problems. 


\subsection{Training Specialized LLMs with Structured Datasets}
\label{sec:FutureDirection1.1}

A promising direction for improving the reasoning capabilities of \acp{llm} in mathematical optimization is the development of structured, domain-specific training datasets. Instead of relying solely on general-purpose corpora, models can benefit substantially from datasets curated for specific application domains, such as scheduling, logistics, or network design, where the problem structure, terminology, and modeling conventions are well defined. Domain specialization of this kind enhances the model’s ability to generalize within its context and generate accurate, interpretable formulations.

The following subsections outline key design principles for constructing such datasets, focusing on structured representation, complexity variation, and data augmentation, each critical for training \acp{llm} to reason effectively in mathematical optimization contexts.

\subsubsection{Structuring Datasets to Capture Reasoning}

Many existing datasets present optimization problems using free-form text, code snippets, or solution outputs without explicitly identifying the core components of the formulation, including the decision variables, objective function, and constraints. This limitation is evident in resources such as~\cite{IndustryORdataset}, which lack the modular structure necessary for teaching models how to reason through the formulation process itself. To support learning that goes beyond simple input-output mapping, datasets should be designed to expose the logical structure of the modeling task transparently and consistently.

Among current resources, the most effective structured datasets typically provide three key components for each record. First, a \textbf{problem statement} offers a natural language description of the optimization task. Second, a file containing the \textbf{mathematical elements} defines the sets, indices, parameters, and decision variables used in the formulation. Third, the \textbf{mathematical formulation} presents the complete formal model, including the objective function and constraints~\cite{complexORdataset}. While this three-part structure establishes a solid foundation, it can be enhanced to better support step-by-step reasoning and interpretability.

We propose extending this structure by introducing a fourth component: A detailed breakdown of the problem statement into its semantic and logical elements. This leads to a standardized four-part structure for each dataset entry:
\begin{enumerate}
\item \textbf{Problem statement}: A clear, natural language description of the optimization scenario.
\item \textbf{Problem breakdown (new addition)}: A structured decomposition of the problem written in English. This includes:
    \begin{itemize}
        \item Identification of the objective to be optimized.
        \item Description of the decision variables (distinguishing between true and auxiliary variables).
        \item Categorization of constraints based on their role: (i) those directly stated in the problem statement, (ii) those added during reformulation (e.g., for linearization), and (iii) auxiliary constraints used to link different model components or support structural decomposition.
    \end{itemize}

    \item \textbf{Mathematical elements}: Definitions of sets, indices, parameters, and associated data. This file should include semantic annotations that map textual concepts to symbolic representations.

    \item \textbf{Mathematical formulation}: The complete optimization model, presented in a modular format (objective function, decision variables, and constraints), with metadata specifying the origin and role of each element.
\end{enumerate}

This extended structure enables models to internalize not only the mapping from text to equations but also the intermediate reasoning steps required to construct a correct and meaningful formulation. It improves transparency, supports error analysis, and facilitates reverse engineering of the modeling process.


A notable step toward structured modeling has been introduced in the OptiMUS system~\cite{ahmaditeshnizi2024optimus}, which demonstrates the power of modular, multi-agent \ac{llm} pipelines that translate natural language descriptions into executable optimization code. While OptiMUS leverages a connection graph and structured prompts to guide agent-based reasoning at inference time, our proposal complements this by focusing on building datasets that embed semantic structure and reasoning pathways directly into the training data. This distinction makes our approach highly synergistic; the proposed dataset format could help train future agent-based systems to reason more effectively, even with limited context or smaller models.


\subsubsection{Incorporating Problem Complexity}

A second key direction is to ensure that datasets span a wide range of problem complexities. Our experiments show that \acp{llm} such as \ac{gpt}-4o and DeepSeek Math struggle with more complex tasks, especially those involving multiple constraints, auxiliary variables, or interdependent decision components. These limitations underscore the need to include both simple and advanced formulations in training data.

By exposing models to problems with varying structural difficulty, we can encourage more robust reasoning behavior and better equip models to generalize to real-world scenarios with nuanced requirements. A diverse complexity range allows \acp{llm} to learn how to deconstruct and solve optimization tasks of differing scales and sophistication.


\subsubsection{Applying Data Augmentation Techniques}
A third important strategy is the use of data augmentation to improve dataset diversity and robustness. Augmentation exposes models to multiple representations of the same problem, strengthening both linguistic and structural generalization, especially in zero-shot and few-shot settings. Effective augmentation techniques include:
\begin{itemize}
    \item \textbf{Paraphrasing}: Rewriting the problem statement using varied natural language.
    \item \textbf{Mathematical rewriting}: Reordering constraints or renaming variables while preserving semantic equivalence to encourage structural flexibility.
    \item \textbf{Constraint variation}: Adding, modifying, or removing constraints to simulate real-world variability.
\end{itemize}

These strategies help models recognize that different surface forms can represent the same underlying logic, which is essential for reasoning-aware generalization.


In summary, improving \ac{llm} reasoning in mathematical optimization requires more than domain coverage. It demands: (i) well-structured datasets that capture intermediate reasoning steps, (ii) inclusion of problems with varying levels of complexity, and (iii) systematic augmentation to increase linguistic and structural diversity. Together, these directions provide a strong foundation for building robust, generalizable, and interpretable optimization modeling systems.

\begin{figure}
\centering
\includegraphics[width=1\linewidth]{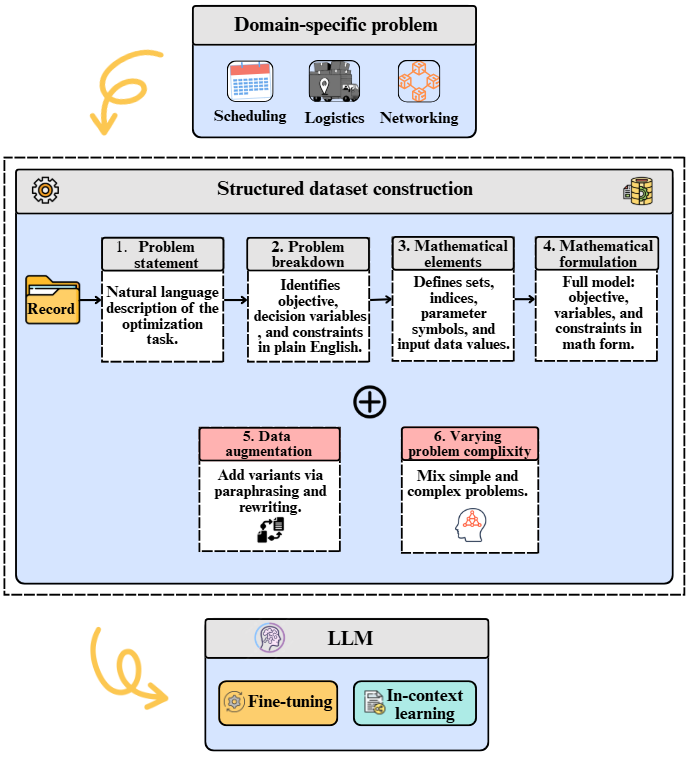}
\caption{Visual representation of the proposed structured dataset construction framework for optimization modeling. Each record consists of four components: (i) A problem statement describing the task in natural language, (ii) a plain-English breakdown of objectives, variables, and constraints, (iii) a list of sets, indices, parameter symbols, and corresponding input data values, and (iv) the complete mathematical formulation. Dataset-level enhancements, such as data augmentation and problem complexity variation, are applied to increase diversity and robustness. The resulting structured dataset is then used for both fine-tuning and in-context learning of \acp{llm}.}
\label{fig:structured_dataset}
\end{figure}

Fig.~\ref{fig:structured_dataset} illustrates the proposed four-part dataset structure, along with enhancements for complexity and augmentation.


\subsection{Modular Use of Multiple LLMs}
\label{sec:FutureDirection1.2}

A promising direction for enhancing the performance of \acp{llm} in mathematical optimization formulation is the transition from single-model pipelines to a modular, multi-agent architecture. Instead of relying on a single \ac{llm} to generate the entire formulation end-to-end, the task can be decomposed into smaller, manageable subtasks, each delegated to a specialized model. This divide-and-conquer strategy enables each \ac{llm} to focus on a specific aspect of the modeling process, potentially improving the overall accuracy, interpretability, and reliability of the formulation pipeline.

The motivation for such modularity is strongly supported by recent advances in multi-agent and cross-team \ac{llm} coordination. Du~\textit{et al.}~\cite{crossteam} presented a comprehensive survey detailing how \acp{llm} can manage multimodal and multi-component tasks effectively. They outline two major strategies: (i) Tool-augmented design, where a central \ac{llm} delegates subtasks to multiple external models tailored for specific operations, and (ii) monolithic \acp{llm} trained to handle multiple modalities within a single model.

Building on this foundation, Zhou~\textit{et al.}~\cite{crossdomain} proposed a central orchestration framework in which an \ac{llm} coordinates domain-specific generative models (DGMs) to synthesize multi-modal sensor data. Likewise, He~\textit{et al.}~\cite{multimodal} introduced a dynamic `cross-team orchestration' approach that resolves the inefficiencies and coordination bottlenecks often encountered in traditional static multi-agent systems, particularly as the complexity of the task scales. These works collectively demonstrate that task decomposition and model specialization lead to substantial improvements in reasoning quality, task success rate, and system flexibility.

Inspired by these results, we propose decomposing the mathematical modeling task across multiple \acp{llm}, each specializing in a particular modeling component or formulation paradigm. We outline four key strategies for task decomposition: (i) Component-wise specialization, (ii) cross-domain specialization, (iii) formulation-type specialization, and (iv) hierarchical decomposition. The following subsections explore each in detail.


\subsubsection{Component-wise Specialization}
\label{sec:FutureDirection1.2.1}

One promising strategy for improving \ac{llm} performance in mathematical optimization is to divide the formulation task into distinct components and assign each to a dedicated model. In this component-wise setup, different \acp{llm} are responsible for specific subtasks, such as identifying decision variables, constructing constraints, or extracting sets and parameters. Each model is trained or fine-tuned to specialize in its role, allowing for more focused reasoning and reducing the overall complexity of the task.

For example, one \ac{llm} may focus on understanding what the problem is asking, interpreting the objective, and identifying the appropriate decision variables. Another may handle the formal definition of constraints, while a third is responsible for extracting relevant sets, indices, and input data from the problem description. This modular structure mirrors how human experts build optimization models: by breaking down the problem into logical parts that can be reasoned about and refined independently. By narrowing the focus of each \ac{llm}, this approach improves accuracy, interpretability, and robustness in the overall formulation process.

\begin{figure*}
\centering
\includegraphics[width=0.75\linewidth]{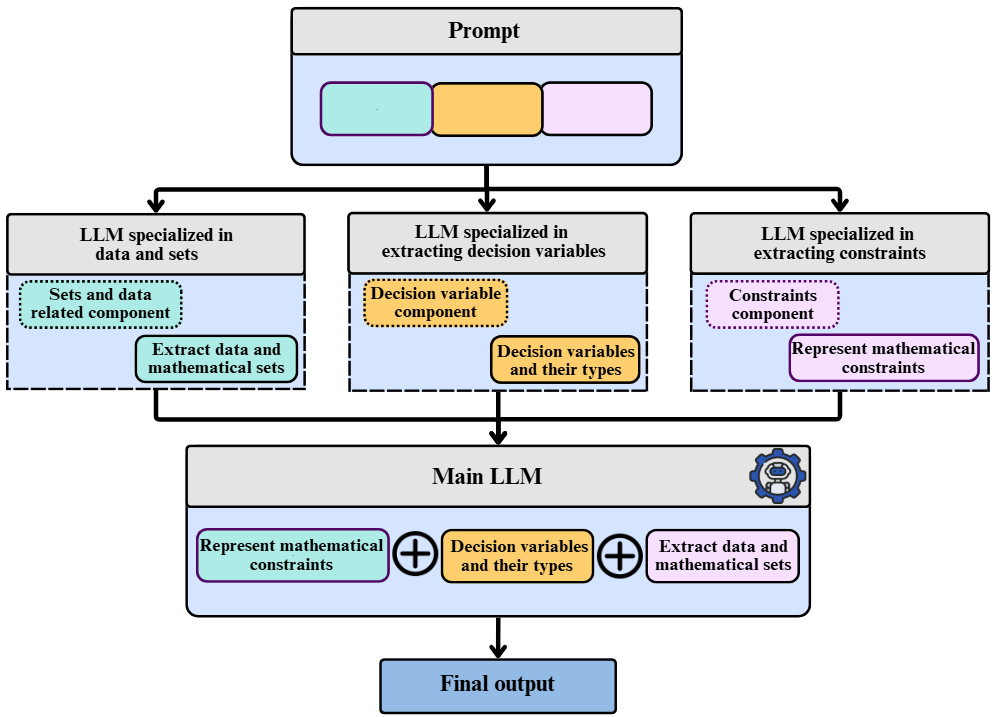}
\caption{Component-wise specialization of the formulation task across dedicated \acp{llm}. One model extracts sets and data, another identifies decision variables and their types, and a third represents mathematical constraints. A main \ac{llm} then integrates these outputs into a unified optimization model.}
\label{fig:fdir}
\end{figure*}

To further illustrate this approach, Fig.~\ref{fig:fdir} presents a more granular component-wise architecture. In this setup, three specialized \acp{llm} are responsible for: (i) Extracting sets and data, (ii) identifying decision variables and their types, and (iii) representing mathematical constraints. A central \ac{llm} then aggregates their outputs into a unified, executable optimization model.

\subsubsection{Cross-domain Specialization}
\label{sec:FutureDirection1.2.2}

Another effective modular strategy involves assigning \acp{llm} based on domain expertise. In this setup, each \ac{llm} is specialized in a specific knowledge domain, such as healthcare, logistics, or scheduling, and is responsible for interpreting problems within that context. For instance, a healthcare-specific \ac{llm} may understand medical terminology, policy constraints, or clinical workflows, while a logistics \ac{llm} focuses on supply chain structures, transportation rules, or delivery windows. These domain-specific \acp{llm} extract relevant information from natural language descriptions, including entities, objectives, and constraints.

Once this domain-level understanding is established, the extracted information is passed to a separate \ac{llm} trained specifically in mathematical modeling. This model is responsible for transforming the contextual knowledge into a formal optimization formulation. The final step is handled by a central \ac{llm} that integrates the mathematical components and domain-specific insights into a complete and coherent model.

\begin{figure*}
\centering
\includegraphics[width=0.75\linewidth]{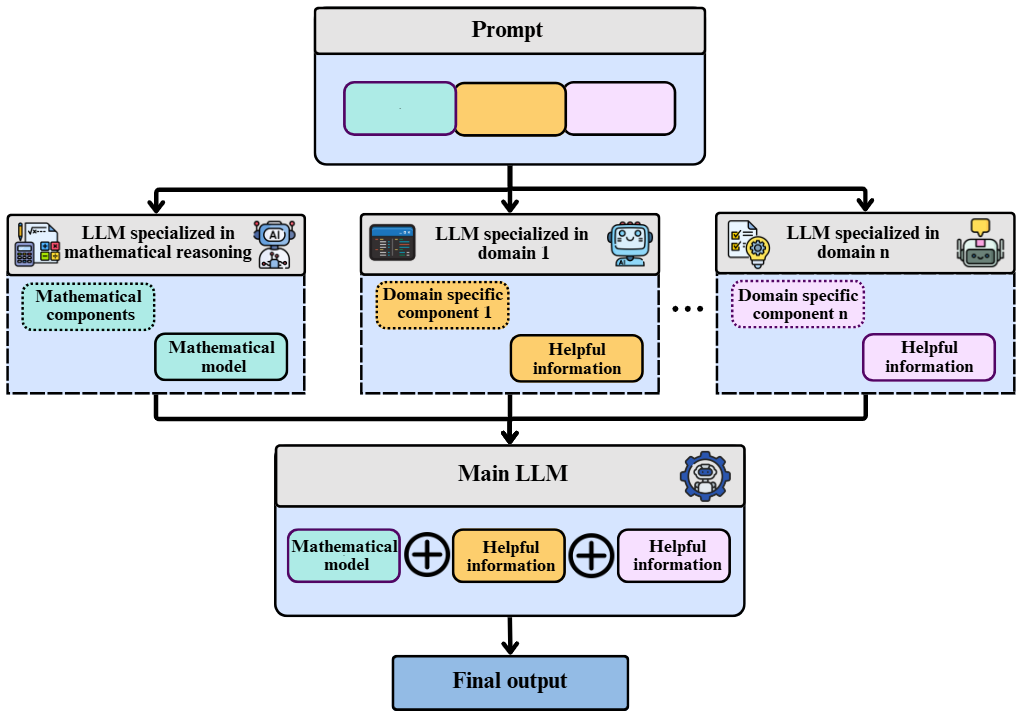}
\caption{Cross-domain specialization for mathematical modeling. Each \ac{llm} is trained on a specific application domain (e.g., healthcare, logistics) and extracts relevant contextual information from the prompt. A separate \ac{llm}, specialized in mathematical reasoning, transforms this information into a formal optimization model. The main \ac{llm} integrates these outputs into a unified formulation.}
\label{fig:fdir1}
\end{figure*}

Fig.~\ref{fig:fdir1} illustrates this cross-domain specialization strategy. Each domain-specific \ac{llm} contributes contextual understanding tailored to its area of expertise, while the mathematical reasoning \ac{llm} formalizes this input into symbolic structures. The outputs are then aggregated by the main \ac{llm} to generate the final optimization formulation.

\begin{figure*}
    \centering
    \includegraphics[width=0.8\linewidth]{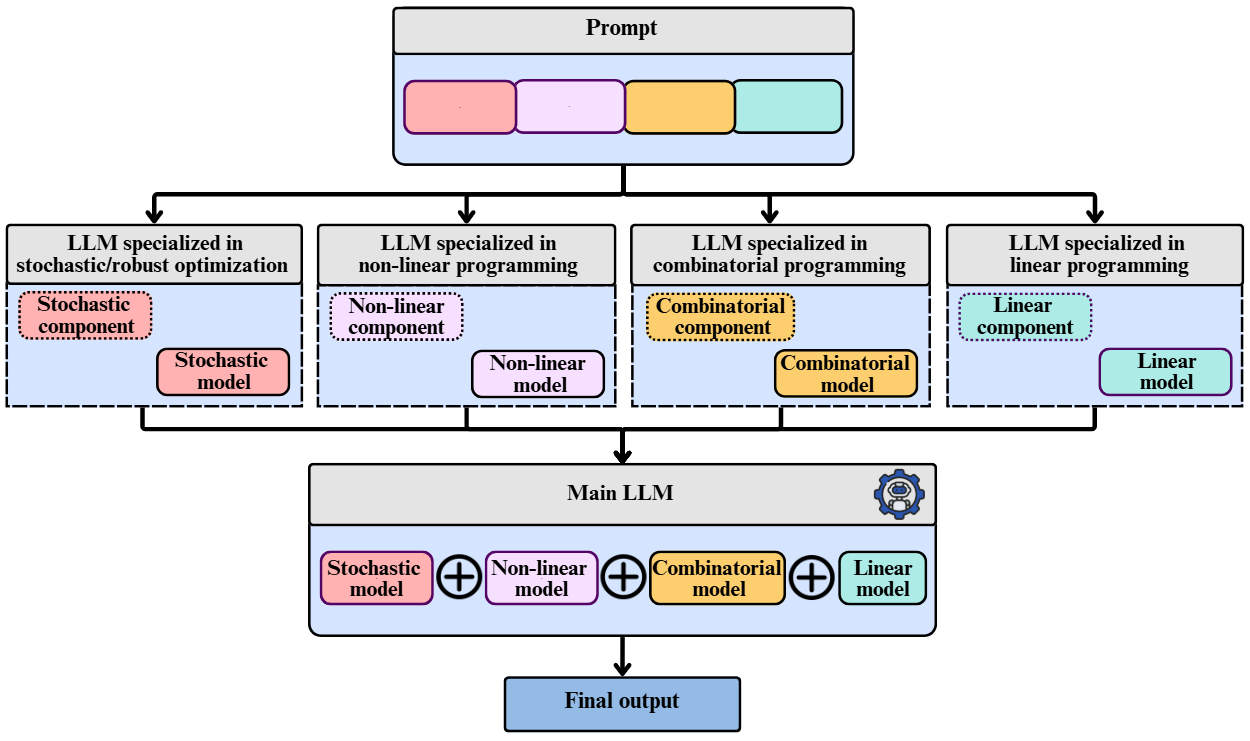}
    \caption{A modular architecture based on formulation-type specialization. Each \ac{llm} is responsible for handling a specific class of optimization problems, such as linear, combinatorial, non-linear, or stochastic programming. These specialized models extract and construct sub-models aligned with their formulation type, which a central \ac{llm} then aggregates into a unified optimization model.}
    \label{fig:fdir3}
\end{figure*}

\subsubsection{Formulation-type Specialization}
\label{sec:FutureDirection1.2.3}

Another modular strategy involves dividing the formulation process based on the type of optimization problem being addressed. In this approach, each \ac{llm} is specialized in a particular mathematical formulation type, such as linear, integer, non-linear, stochastic, or combinatorial optimization. This setup reduces the need for a single model to handle multiple mathematical paradigms, enabling each specialized \ac{llm} to focus on the unique structure and reasoning patterns of its assigned formulation type.

As illustrated in Fig.~\ref{fig:fdir3}, the initial prompt is parsed and routed to the appropriate \acp{llm} based on the characteristics of the problem. For instance, one \ac{llm} may handle formulations involving continuous variables and linear constraints, while another is responsible for integer programming elements. Additional models may address non-linear structures or uncertainty-based (stochastic) components. Each specialized \ac{llm} generates a portion of the overall model, tailored to its formulation type, and these outputs are then aggregated by a central \ac{llm} into a complete, coherent optimization formulation.


\subsubsection{Hierarchical Decomposition}
\label{sec:FutureDirection1.2.3}

Another promising direction is the use of a hierarchical architecture, in which a high-level controller \ac{llm} is responsible for decomposing a complex problem into smaller sub-problems based on structural, domain-specific, or formulation-related complexity. Each sub-problem can then be assigned to a specialized \ac{llm} for independent processing. Once the subtasks are completed, the controller reassembles the outputs into a coherent and complete optimization formulation. This hierarchical setup enables dynamic problem structuring, supports domain-specific fine-tuning at the sub-problem level, and offers scalability for handling multi-objective or highly complex formulations.

\subsection{Chain of RAGs}
\label{sec:FutureDirection1.3}

RAGs have emerged as a powerful technique for enhancing the reasoning capabilities of \acp{llm} by incorporating external knowledge into the generation process. Traditional \ac{rag} frameworks typically perform a single retrieval step before generation, using the retrieved content as static context. While effective in many knowledge-intensive tasks, this approach is fundamentally limited by the quality and relevance of the retrieved information. In complex scenarios, a single retrieval step often fails to capture all necessary aspects of the problem, leading to incomplete, irrelevant, or even hallucinated outputs. To overcome this limitation, the chain-of-RAGs paradigm was proposed by Wang~\textit{et al.}~\cite{wang2025chain}. In this framework, retrieval is performed iteratively: Each step refines the context based on the evolving state of reasoning. Rather than retrieving all relevant documents at once, the model incrementally forms sub-queries, gathers targeted evidence at each stage, and progressively builds a more informed and accurate representation. This iterative retrieval mechanism improves contextual alignment, reduces noise, and supports deeper, multi-hop reasoning, especially in domains where stepwise information acquisition is essential.

In the context of mathematical optimization, chain-of-RAGs presents a promising opportunity to enhance \ac{llm} reasoning by mirroring the natural workflow of human modelers. Rather than attempting to formulate the entire problem in a single step, the \ac{llm} can engage in a structured, staged process, retrieving and reasoning over key components sequentially. For example, the model might begin by identifying and retrieving domain-specific patterns for decision variables (e.g., assignment variables in routing or scheduling tasks), followed by a focused retrieval stage for constraints (such as capacity limits, time windows, or precedence relations), and finally, formulate the objective function based on the previously gathered structure. Each retrieval stage would be guided by the partial formulation constructed thus far, allowing the \ac{llm} to refine its understanding dynamically. Moreover, integrating structured external knowledge, such as optimization templates, technical documentation, or domain-specific knowledge graphs, into the retrieval pipeline can further guide the model toward more accurate and complete formulations. This approach has the potential to reduce hallucinations, increase interpretability, and improve performance on complex, multi-component problems where current end-to-end prompting strategies often fall short. As such, chain-of-RAGs offers a compelling pathway for enhancing the reasoning depth and formulation accuracy of \acp{llm} in mathematical optimization tasks.

\subsection{Improving Prompting Strategies}\label{sec:FutureDirection1.4}

Prompting plays a critical role in guiding \acp{llm} to generate mathematical optimization formulations. However, as observed in our experiment and the literature, there is no universally optimal prompt. A prompt that performs well for one type of problem may fail on another. For example, while some strategies may be effective for simple assignment problems, they often fall short in capturing the complexity of scheduling or multi-stage tasks. This inconsistency highlights the need for more flexible and intelligent prompting strategies.

One promising direction is the development of dynamic prompting frameworks that can adapt prompts based on the characteristics of the problem. For instance, if a task involves numerous binary variables or nested constraints, the system could automatically select a prompt format better suited to that structure. A classifier or meta-model could be trained to predict the most appropriate prompt style for different problem types.

Another valuable direction involves designing new, general-purpose prompts specifically tailored for optimization tasks. These could combine structured expert-style guidance with step-by-step reasoning, for example, prompting the model to define decision variables first, followed by constraints, and finally the objective function. Such prompts could embed a standardized formulation scaffold, leading the model through a logical and interpretable workflow rather than relying on open-ended instructions. This structured approach may help reduce ambiguity, improve consistency across problem types, and minimize common formulation errors such as missing components or misaligned constraints.

In summary, improving prompting strategies, whether through adaptive mechanisms or robust, domain-specific prompt templates, remains a key avenue for enhancing the accuracy, reliability, and generalizability of \acp{llm} in mathematical optimization formulation.

\subsection{Neuro-Symbolic Formulation in Mathematical Optimization}\label{sec:FutureDirection1.5}

Despite recent advances in applying \acp{llm} to optimization modeling, recent investigations have identified persistent limitations in \acp{llm}’ ability to generate correct and verifiable mathematical formulations~\cite{luzzi2025chatgpt,liu2024variable,kadiouglu2024ner4opt}. These limitations are particularly critical in mathematical optimization tasks, where structural feasibility, constraint satisfaction, and solution optimality must be strictly preserved. Addressing these challenges calls for hybrid neuro-symbolic systems that combine neural language understanding with symbolic solvers, verification procedures, and programmatic constraint reasoning.

One promising direction is the integration of program-aided reasoning frameworks, such as program-aided language (PAL) models and tool-integrated reasoning agents (ToRA), where \acp{llm} are fine-tuned or prompted to generate executable optimization code using libraries, such as Gurobi, \ac{or}-Tools, or PuLP~\cite{gao2023pal,gou2024tora}. These frameworks translate natural language descriptions into formal mathematical models and offload the solving process to dedicated optimization solvers. Additionally, reasoning agents based on architectures like ReAct and \ac{or}-\ac{llm}-Agent enhance this interaction by allowing the \ac{llm} to decide when to call an external solver, how to handle infeasibility, and how to iteratively refine model components~\cite{yao2023react,zhang2025orllm}.

Symbolic feedback mechanisms, such as reporting constraint violations, infeasibility, or numerical instability, can guide \acp{llm} to revise and regenerate improved formulations. These feedback loops, forming a core part of hybrid neuro-symbolic systems, support the iterative validation of solution structures and enhance robustness in deployment~\cite{zhang2025or,kadiouglu2024ner4opt}. Furthermore, synthetic data generation using optimization solvers provides a scalable path to fine-tune \acp{llm} with structurally sound, domain-specific training instances~\cite{dui2025generative,liu2024variable}. This process supports learning optimization patterns across application domains like logistics, energy, and scheduling, and serves as a valuable complement to manually curated datasets. It also enables domain adaptation while ensuring solver-grounded correctness and feasibility.

A key enabler of correctness in such systems is formal constraint verification, often supported by \ac{smt} solvers or feasibility checkers~\cite{gemp2024steering}. \Ac{smt} solvers verify whether a symbolic model meets logical and numerical constraints across domains such as scheduling, planning, or energy balancing. Similarly, abstract planning modules serve as high-level reasoning layers that generate constraint sketches and structural model outlines before detailed formulation~\cite{li2025abstract}. In summary, neuro-symbolic architectures offer a compelling pathway to enable verifiable and scalable optimization modeling. These systems blend the generative strengths of \acp{llm} with the rigor of symbolic solvers and structured reasoning. 

\begin{figure}
\centering
\includegraphics[width=0.95\linewidth]{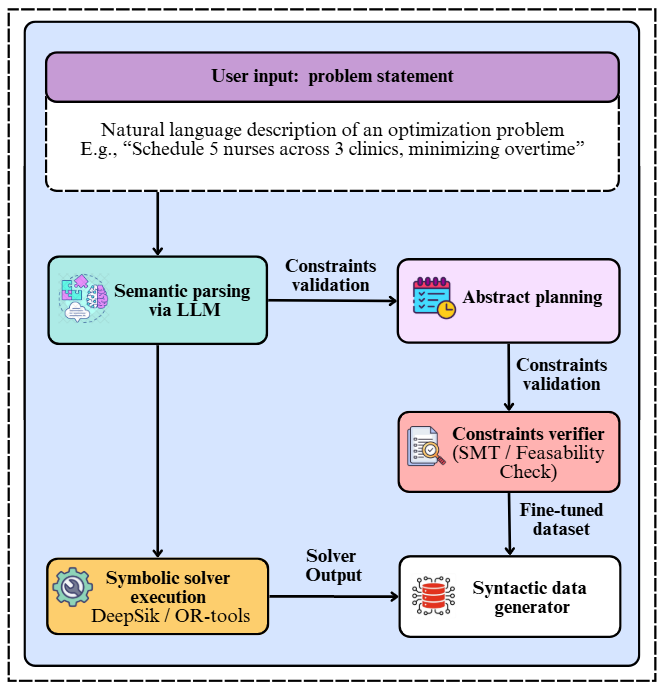}
\caption{Neuro-symbolic architecture combines language models with symbolic solvers for accurate optimization.}
\label{fig:neuro-symbolic-architecture}
\end{figure}

Fig.~\ref{fig:neuro-symbolic-architecture} presents a conceptual neuro-symbolic architecture for optimization modeling, integrating semantic interpretation, symbolic validation, and solver feedback. The process begins with a natural language description of a decision problem, which is parsed by a language model (e.g., DeepSeek) to generate an initial formulation. The \ac{llm} output is then handled in two parallel paths. One path sends it to a symbolic solver (e.g., DeepSeek, \ac{or}-Tools) for computation, while the other runs a validation workflow. This includes an abstract planning step, an intermediate stage that extracts structural elements such as objectives, variables, and constraints from the user's intent. Validation continues via constraint checkers, including \ac{smt}-based feasibility checks to ensure logical and mathematical consistency. If errors are detected, the system may regenerate or revise the formulation. A synthetic data generator can also create structurally similar validated problems to fine-tune the \ac{llm} and reinforce solver-aligned learning. The architecture distills emerging design patterns across recent neuro-symbolic reasoning work~\cite{gao2023pal,gou2024tora,zhang2025orllm,yao2023react}, supporting iterative verification, trust, and correctness in \ac{llm}-driven optimization.



\section{Understanding and Diagnosing LLM Limitations}\label{sec:FutureDirection2}

Despite ongoing efforts to improve the learning and reasoning capabilities of \acp{llm}, a fundamental challenge remains: We still lack a clear understanding of why \acp{llm} succeed in some formulation tasks and fail in others. As demonstrated in our experiment, a model may generate accurate constraints but incorrect decision variables, solve one problem but struggle with a structurally similar one, or respond inconsistently to different prompts. These unpredictable behaviors underscore the need for a deeper understanding of how \acp{llm} reason and where their limitations lie.

To move beyond surface-level performance, it is essential to develop tools and methodologies that allow for more precise interpretation and analysis of model behavior. In this section, we propose two future directions: (i) designing improved evaluation metrics that enable fine-grained assessment of model outputs, and (ii) applying feature-based mapping techniques to identify patterns between problem characteristics and model performance. Together, these approaches aim to provide a clearer picture of what \acp{llm} can and can not do and to support the development of more robust, interpretable, and dependable models for mathematical optimization formulation.


\subsection{Designing Better Evaluation Metrics}\label{sec:FutureDirection2.1}



A fundamental challenge in evaluating \acp{llm} for mathematical optimization tasks is the lack of interpretability behind their successes and failures. In our experiment, we observed cases where an \ac{llm} produced a correct formulation for one problem but failed on another with a similar structure. In some instances, the model accurately captured the constraints but failed to define the decision variables or the objective function. These inconsistencies highlight a critical gap: While current evaluation metrics can assess overall formulation correctness, they often do so without sufficient precision to identify which components are correct or flawed.

To address this issue, it is important to examine the limitations of the evaluation metrics used in our experiments. Specifically, we employed three dimensions: Optimality gap, token-level F1 score, and compilation accuracy, to obtain a broad overview of model performance. While these metrics can assess formulation correctness to some extent, each carries inherent weaknesses that limit their ability to precisely diagnose where and why a formulation fails.


In our experiments, we evaluated \ac{llm} outputs using three primary dimensions: Optimality gap, token-level F1 score, and compilation accuracy, to obtain a broad overview of model performance and to reflect the metrics applied in our study. However, each metric carries inherent weaknesses that limit its ability to precisely diagnose where and why a formulation fails.

The token-level F1 score, designed to assess symbolic similarity between the generated and reference formulations, is highly sensitive to surface-level variations that do not impact semantic meaning. Differences in index notation (e.g., $i$ and $j$ vs. $d$ and $l$), variable naming, or formatting can result in low F1 scores even when the model's output is functionally equivalent to the reference. Moreover, \ac{llm}-generated outputs often omit formal elements, such as quantifiers or domain constraints, further complicating automated parsing and symbolic comparison.

The optimality gap, although widely used as a solver-level metric, also presents challenges when applied to \ac{llm}-generated formulations. A low gap may suggest that the model produced a correct or near-optimal solution, yet it could mask significant issues in variable definitions, constraint structures, or objective logic. Conversely, a high gap might stem from minor syntax errors or alternative encodings that are semantically valid but solver-incompatible. Critically, the optimality gap provides no information about which part of the formulation is flawed, limiting its usefulness for model diagnosis or improvement.

Compilation accuracy, which checks whether the generated code compiles and runs, serves as a basic quality filter. While valuable for detecting incomplete or malformed outputs, this metric offers only a shallow indication of correctness. Code that executes successfully may still represent an incorrect or incomplete optimization model. Additionally, syntactic success is often influenced by prompt phrasing, making it difficult to disentangle prompt engineering effects from actual model capability.

Together, these limitations reveal the need for more comprehensive and diagnostic evaluation strategies. As \acp{llm} are increasingly used in critical reasoning tasks, particularly in optimization and mathematical modeling, understanding why a model works or fails is essential.

Future directions in evaluation should include component-level metrics that assess the correctness of individual elements within a formulation, specifically, decision variables, constraints, and objective functions. Relying solely on aggregate scores can obscure which part of the formulation caused the failure. Additionally, graph-based equivalence testing can provide a more robust assessment of structural similarity by focusing on the underlying logic rather than superficial differences in syntax. Incorporating human-in-the-loop evaluation, especially for domain-specific or ambiguous problems, can further help calibrate and validate automated scoring mechanisms.

To enable both broad and detailed insights, we propose the development of a comprehensive evaluation framework that integrates multiple complementary metrics. Such a framework could combine solver-level, symbolic, and syntactic evaluations to provide a well-rounded assessment of model performance. Moreover, it could include a modular architecture comprising multiple \acp{llm}, each trained to evaluate or generate a specific component of the formulation. This would allow each element, such as constraints or decision variables, to be assessed in isolation, offering a clearer diagnosis of errors. By combining these metrics and component-wise evaluations, the framework would support both high-level benchmarking and fine-grained analysis, offering a more interpretable and actionable understanding of \ac{llm} behavior in mathematical optimization tasks.

In summary, advancing evaluation methods beyond single-score metrics is not merely a technical refinement; it is a necessary step toward building transparent, reliable, and interpretable \ac{llm} systems for complex reasoning tasks.


\subsection{Feature Mapping for Failure Analysis}\label{sec:FutureDirection2.2}

Traditional evaluation metrics, such as optimality gap, token-level F1 score, and compilation accuracy, offer a quantitative assessment of model performance but provide limited insight into why an \ac{llm} succeeds on certain optimization problems and fails on others. In our experiment, we observed inconsistent behavior across problem instances. For example, a model may fully solve P1, generate a partially correct formulation for P2, and fail on Problem~P3 even when the problems appear structurally similar. These unpredictable outcomes highlight the limitations of surface-level metrics and underscore the need for more diagnostic tools. To address this, we propose feature mapping as a promising approach for better understanding and interpreting \ac{llm} performance in optimization problem formulation.

Feature mapping involves characterizing each problem using a set of descriptive attributes that capture its structural and semantic properties. These attributes may include the type and number of decision variables (e.g., binary, integer, or continuous), the number and nature of constraints (e.g., equality, inequality, or logical conditions), the complexity of the objective function (e.g., linear or nonlinear), the overall problem size (i.e., number of variables and constraints), and the application domain (e.g., routing, scheduling, or resource allocation). By aligning these characteristics with observed model outcomes, success, partial success, or failure, we can begin to identify patterns and uncover structural factors that influence \ac{llm} behavior.

For instance, an \ac{llm} may consistently succeed on problems involving only continuous variables and a small number of linear constraints but frequently fail on problems with binary variables and nested logical dependencies. When such patterns are systematically observed, they can serve as ``failure signatures'' or heuristic rules. These rules might indicate, for example, that the model is more reliable when the formulation includes fewer than five constraints and a linear objective, but less so when binary decision variables and conditional logic are present.

Feature mapping can be applied at multiple levels of granularity. At the problem level, it enables analysis of the overall formulation and corresponding outcomes. At a finer level, constraint-level mapping can help identify specific structures, such as disjunctive constraints or conditional dependencies, that consistently trigger formulation errors. This multilevel analysis supports a more precise understanding of model limitations.

Ultimately, feature mapping can inform a range of future developments. It enables rule-based prediction of model performance, supports adaptive prompt selection based on problem features, and provides actionable insights for targeted model refinement or fine-tuning. Most importantly, it offers a pathway toward more interpretable, predictable, and reliable use of \acp{llm} in complex reasoning tasks such as mathematical optimization.

\section{Conclusion}\label{sec:conclusion}

This study explored the capabilities and limitations of \acp{llm} in mathematical optimization modeling through a systematic literature review and empirical evaluation. Our meta-analysis revealed persistent empirical gaps in symbolic reasoning, formulation accuracy, and interpretability, underscoring key limitations in current \ac{llm}-based approaches. To investigate these challenges, we developed a domain-specific dataset and evaluated the performance of state-of-the-art \acp{llm} across multiple prompting strategies. While the models demonstrated progress in interpreting problem statements and generating partial formulations, the results revealed ongoing difficulties in producing complete, accurate, and interpretable mathematical formulations, particularly for complex, domain-specific problems.

In response, we propose a set of targeted research directions to address these limitations. These include designing structured, domain-adapted datasets; adopting modular, multi-agent architectures; advancing prompting strategies; and integrating neuro-symbolic reasoning with retrieval-augmented generation techniques. We highlight two key research priorities emerging from the identified empirical gaps: (i) advancing \ac{llm} learning and mathematical reasoning capabilities and (ii) improving the understanding and diagnosis of \ac{llm} limitations in the context of optimization modeling. These recommendations have the potential not only to enhance the effectiveness of \acp{llm} in solving complex real-world optimization problems but also to support better decision-making, ultimately leading to substantial cost savings for organizations.

\bibliography{arXiv.bib}

\bibliographystyle{IEEEtran}

 \section*{Appendix A:}\label{app}
 In our dataset, we have ten problems, each contains three main files: The problem description, the input data, and the ground truth model. In the following, we provide a detailed description of each problem:

  \begin{figure*}
 \begin{tcolorbox}[colback=green!5!white, colframe=green!50!black, title= Problem 1]
\textbf{Problem description}:\\
The network resource allocation problem aims to assign network devices (e.g., routers, switches, or servers) to communication links or data routes in order to minimize the total transmission cost while satisfying traffic demand constraints using limited available devices. The problem involves a set of network devices and a set of communication routes. Given the cost of assigning a particular device to a specific route, the objective is to minimize the total assignment cost. Each device type is available in limited quantity, and the allocation of devices can not exceed their availability. Each communication route has a traffic demand, and each device has a capacity representing the maximum traffic it can handle on that route. The demand constraint ensures that the total allocated capacity on each route meets or exceeds its required traffic load. The goal is to determine the most cost-efficient assignment of devices to routes while respecting both resource limitations and service requirements. \\
\textbf{Input data:} \\{
            ``Availability'': [2, 3, 1],
            ``Demand'': [100, 150],
            ``Capability'': [
                [50, 70],
                [60, 80],
                [70, 90]
            ],
            ``Cost'': [
                [100, 200],
                [150, 250],
                [200, 300]
            ] \\
\textbf{Output:} 700
    } \\
\textbf{Ground truth model}: \\
\textbf{Sets:}
\begin{itemize}
    \item $\text{Devices}$: Set of devices.
    \item $\text{Links}$: Set of links.
\end{itemize}

\textbf{Parameters:}
\begin{itemize}
    \item $\text{Availability}_d$: Availability of device $d$, $\forall d \in \text{Devices}$.
    \item $\text{Demand}_l$: Traffic demand for link $l$, $\forall l \in \text{Links}$.
    \item $\text{Capability}_{d,l}$: Capability of device $d$ for link $l$, $\forall d \in \text{Devices},\ l \in \text{Links}$.
    \item $\text{Cost}_{d,l}$: Cost of assigning device $d$ to route $l$, $\forall d \in \text{Devices},\ l \in \text{Links}$.
\end{itemize}

\textbf{Decision Variables:}
\begin{itemize}
    \item $\text{Allocation}_{d,l}$: Allocation of device $d$ to route $l$, $\forall d \in \text{Devices},\ l \in \text{Links}$.
\end{itemize}

\textbf{Objective:} Minimize the total cost of the assignment:
\[
\text{minimize} \sum_{d \in \text{Devices}} \sum_{l \in \text{Links}} \text{Cost}_{d,l} \cdot \text{Allocation}_{d,l}
\]

\textbf{Constraints:}
\begin{enumerate}
    \item Device availability constraint:
    \[
    \sum_{l \in \text{Links}} \text{Allocation}_{d,l} \leq \text{Availability}_d, \quad \forall d \in \text{Devices}
    \]
    
    \item Demand satisfaction constraint:
    \[
    \sum_{d \in \text{Devices}} \text{Allocation}_{d,l} \cdot \text{Capability}_{d,l} = \text{Demand}_l, \\ \quad \forall l \in \text{Links}
    \]
\end{enumerate}
\end{tcolorbox}
 \end{figure*}

\begin{figure*}
\begin{tcolorbox}[colback=green!5!white, colframe=green!50!black, title= Problem 2]
\textbf{Problem description:} \\
The network node assignment problem aims to assign client devices to network nodes (e.g., access points, servers, or routers) in a way that maximizes the total number of successful connections. The problem involves a set of client devices and a set of network nodes, where each client device is capable of connecting to a subset of nodes based on compatibility, location, or signal strength. The objective is to determine the optimal assignment of clients to nodes that satisfies certain network constraints, such as capacity limits or connection rules. \\
\textbf{Input data:} \\
``Clients'': [``C1'', ``C2'', ``C3''],
``Nodes'': [``N1'', ``N2'', ``N3''],
``PossibleAssignments'': [
                [1, 0, 1],
                [0, 1, 0],
                [1, 1, 1]
]\\
\textbf{Output:} \\
3 \\
\textbf{Ground truth model}: \\
\textbf{Sets:}
\begin{itemize}
    \item $\text{Clients}$: Set of all clients.
    \item $\text{Nodes}$: Set of all network nodes.
\end{itemize}

\textbf{Parameter:}
\begin{itemize}
    \item $\text{PossibleAssignments}_{c,n}$: Indicates if client $c$ is interested in node $n$, $\forall c \in \text{Clients},\ n \in \text{Nodes}$.
\end{itemize}

\textbf{Decision Variables:}
\begin{itemize}
    \item $\text{Assignment}_{c,n}$: 1 if client $c$ is assigned to node $n$, 0 otherwise, $\forall c \in \text{Clients},\ n \in \text{Nodes}$.
\end{itemize}

\textbf{Objective:} Maximize the total number of assignments:
\[
\text{maximize} \sum_{c \in \text{Clients}} \sum_{n \in \text{Nodes}} \text{Assignment}_{c,n}
\]

\textbf{Constraints:}
\begin{enumerate}
    \item Available assignments:
    \[
    \text{Assignment}_{c,n} \leq \text{PossibleAssignments}_{c,n}, \quad \forall c \in \text{Clients},\ \forall n \in \text{Nodes}
    \]
    
    \item Each client is assigned to at most one node:
    \[
    \sum_{n \in \text{Nodes}} \text{Assignment}_{c,n} \leq 1, \quad \forall c \in \text{Clients}
    \]
    
    \item Each node is assigned to at most one client:
    \[
    \sum_{c \in \text{Clients}} \text{Assignment}_{c,n} \leq 1, \quad \forall n \in \text{Nodes}
    \]
\end{enumerate}
\end{tcolorbox}
\end{figure*}

\begin{figure*}
    
\begin{tcolorbox}[colback=green!5!white, colframe=green!50!black, title= Problem 3]
\textbf{Problem description:} 

A telecom company needs to build a set of cell towers to provide signal coverage for the inhabitants of a given city. A number of potential locations where the towers could be built have been identified. The towers have a fixed range, and due to budget constraints, only a limited number of them can be built. Given these restrictions, the company wishes to provide coverage to the largest percentage of the population possible. To simplify the problem, the company has split the area it wishes to cover into a set of regions, each of which has a known population. The goal is then to choose which of the potential locations the company should build cell towers on in order to provide coverage to as many people as possible.\\
\textbf{Input data: }\\
``$\Delta$'': [[1, 0, 1],
                [0, 1, 0]
            ],
            ``Cost'': [3, 4],
            ``Population'': [100, 200, 150],
            ``Budget'': 4\\
\textbf{Output:} \\
200\\
\textbf{Ground truth model:} \\
\textbf{Sets:}
\begin{itemize}
    \item $\text{Towers}$: Set of potential sites to build a tower.
    \item $\text{Regions}$: Set of regions.
\end{itemize}

\textbf{Parameters:}
\begin{itemize}
    \item $\Delta_{i,j}$: Binary parameter, equals 1 if site $i$ covers region $j$, 0 otherwise; $\forall i \in \text{Towers},\ j \in \text{Regions}$.
    \item $\text{Cost}_i$: Cost of setting up a tower at site $i$, $\forall i \in \text{Towers}$.
    \item $\text{Population}_j$: Population of region $j$, $\forall j \in \text{Regions}$.
    \item $\text{Budget}$: Total budget allocated to build the towers.
\end{itemize}

\textbf{Decision Variables:}
\begin{itemize}
    \item $\text{Covered}_j \in \{0,1\}$: 1 if region $j$ is covered, 0 otherwise; $\forall j \in \text{Regions}$.
    \item $\text{Build}_i \in \{0,1\}$: 1 if tower at site $i$ is built, 0 otherwise; $\forall i \in \text{Towers}$.
\end{itemize}

\textbf{Objective:} Maximize the total population covered:
\[
\text{maximize} \sum_{j \in \text{Regions}} \text{Population}_j \cdot \text{Covered}_j
\]
\textbf{Constraints:}
\begin{enumerate}
    \item Region coverage constraint:
    \[
    \sum_{i \in \text{Tower}} \Delta_{i,j} \cdot \text{Build}_i \geq \text{Covered}_j, \quad \forall j \in \text{Regions}
    \]
    
    \item Budget constraint:
    \[
    \sum_{i \in \text{Tower}} \text{Cost}_i \cdot \text{Build}_i \leq \text{Budget}
    \]
\end{enumerate}
\end{tcolorbox}
\end{figure*}

\begin{figure*}
\begin{tcolorbox}[colback=green!5!white, colframe=green!50!black, title = Problem 4]
\textbf{Problem description:} \\
A set of data flows, Flows, needs to be transmitted through a series of network functions or nodes, Nodes (e.g., firewalls, NATs, intrusion detection systems), in a fixed sequential order, from Node $1$ to Node $M$. Each node is capable of processing multiple flows in parallel. The transmission workflow is as follows: The first flow in the sequence is forwarded to the first node for processing, while the remaining flows wait. Once the first node completes processing the first flow, that flow moves on to the second node, and the second flow begins processing at the first node. This continues in a pipelined fashion. The time required to process flow $f$ at node $n$ is ProcessTime$_{f, n}$. The objective is to determine the optimal scheduling of flows through the nodes that minimizes the total makespan (i.e., the time at which all flows have been fully processed through the entire sequence of network nodes).\\
\textbf{Input data: }\\
``Flows'': [1, 2, 3],
            ``Schedules'': [1, 2, 3],
            ``Nodes'': [1, 2],
            ``ProcessTime'': [
                [1, 3],
                [2, 2],
                [3, 1]
            ]\\
\textbf{Output:} \\
7\\
\textbf{Ground truth model:} \\
\textbf{Sets:}
\begin{itemize}
    \item $\text{Flows}$: Set of all flows.
    \item $\text{Schedules}$: Set of all schedule positions.
    \item $\text{Nodes}$: Set of all processing nodes.
\end{itemize}

\textbf{Parameters:}
\begin{itemize}
    \item $\text{ProcessTime}_{f,n}$: Time required to process flow $f$ on node $n$, $\forall f \in \text{Flows},\ n \in \text{Nodes}$.
\end{itemize}

\textbf{Decision Variables:}
\begin{itemize}
    \item $\text{Flowschedule}_{f,s} \in \{0,1\}$: $1$ if flow $f$ is assigned to schedule position $s$, $0$ otherwise, $\forall f \in \text{Flows},\ s \in \text{Schedules}$.
    \item $\text{StartTime}_{s,n} \geq 0$: Start time of schedule position $s$ on node $n$, $\forall s \in \text{Schedules},\ n \in \text{Nodes}$.
\end{itemize}

\textbf{Objective:} Minimize total processing time:
\[
\text{minimize} \left( \text{StartTime}_{s_{\max}, n_{\max}} + \sum_{f \in \text{Flows}} \text{ProcesTime}_{f,n_{\max}} \cdot \text{Flowschedule}_{f,s_{\max}} \right)
\]

\textbf{Constraints:}
\begin{enumerate}
    \item Only one flow is assigned to each schedule position:
    \[
    \sum_{f \in \text{Flows}} \text{Flowschedule}_{f,s} = 1, \quad \forall s \in \text{Schedules}
    \]
    
    \item Each flow is assigned exactly one schedule position:
    \[
    \sum_{s \in \text{Schedules}} \text{Flowschedule}_{f,s} = 1, \quad \forall f \in \text{Flows}
    \]
    
    \item Schedule position $s$ on node $n+1$ must start no earlier than the finish time of the same schedule position on node $n$:
    \[
    \text{StartTime}_{s,n+1} \geq \text{StartTime}_{s,n} + \sum_{f \in \text{Flows}} \text{ProcesTime}_{f,n} \cdot \text{Flowschedule}_{f,s}, \quad \forall s \in \text{Schedules},\ n \in \text{Nodes} \setminus \{n_{\max}\}
    \]
    
    \item Schedule position $s+1$ on node $n$ must start after the previous schedule finishes on the same node:
    \[
    \text{StartTime}_{s+1,n} \geq \text{StartTime}_{s,n} + \sum_{f \in \text{Flows}} \text{ProcesTime}_{f,n} \cdot \text{Flowschedule}_{f,s}, \quad \forall s \in \text{Schedules} \setminus \{s_{\max}\},\ n \in \text{Nodes}
    \]
\end{enumerate}
\end{tcolorbox}
\end{figure*}

\begin{figure*}
\begin{tcolorbox}[colback=green!5!white, colframe=green!50!black, title = Problem 5]
\textbf{Problem description:} \\
We have a set of network links (point-to-point data transmission paths), each with limited bandwidth capacity. Based on service design and market analysis, we define a set of end-to-end communication services (e.g., VPNs, video streaming paths, or application-level sessions) to offer as packages, each associated with a specific price. Each package consists of a predefined path through one or more network links. For each package, we have an estimated user demand. The question is: How many units of each service package should be provisioned or sold in order to maximize total revenue? Bandwidth is reserved on the underlying network links according to the number of service packages we commit to deliver.

\textbf{Input data:} \\
``AvailableBandwidth'': [50, 60, 70],
            ``Demand'': [30, 40],
            ``Revenue'': [100, 150],
            ``$\Delta$'': [[1, 1, 0], [0, 1, 1]]\\
\textbf{Output:}\\
8000\\
\textbf{Ground truth model: }\\
\textbf{Sets:}
\begin{itemize}
    \item $\text{NetworkLinks}$: Set of network links (point-to-point transmission paths).
    \item $\text{Packages}$: Set of packages.
\end{itemize}

\textbf{Parameters:}
\begin{itemize}
    \item $\text{AvailableBandwidth}_r$: Bandwidth capacity for link $r$, $\forall r \in \text{NetworkLinks}$.
    \item $\text{Demand}_p$: Estimated demand for package $p$, $\forall p \in \text{Packages}$.
    \item $\text{Revenue}_p$: Revenue per unit of package $p$, $\forall p \in \text{Packages}$.
    \item $\Delta_{p,r}$: Binary parameter; equals $1$ if package $p$ uses resource $r$, $0$ otherwise, $\forall p \in \text{Packages},\ r \in \text{NetworkLinks}$.
\end{itemize}

\textbf{Decision Variables:}
\begin{itemize}
    \item $\text{Sell}_p \geq 0$: Number of units of package $p$ to sell, $\forall p \in \text{Packages}$.
\end{itemize}

\textbf{Objective:} Maximize total revenue:
\[
\text{maximize} \sum_{p \in \text{Packages}} \text{Revenue}_p \cdot \text{Sell}_p
\]

\textbf{Constraints:}
\begin{enumerate}
    \item Do not sell more than the estimated demand:
    \[
    \text{Sell}_p \leq \text{Demand}_p, \quad \forall p \in \text{Packages}
    \]
    
    \item Do not exceed available bandwidth on any network link:
    \[
    \sum_{p \in \text{Packages}} \Delta_{p,r} \cdot \text{Sell}_p \leq \text{AvailableBandwidth}_r, \quad \forall r \in \text{NetworkLinks}
    \]
\end{enumerate}
\end{tcolorbox}
\end{figure*}

\begin{figure*}
\begin{tcolorbox}[colback=green!5!white, colframe=green!50!black, title = Problem 6]
\textbf{Problem description:} \\
Consider a data routing problem involving multiple types of network traffic. Given a set of nodes, Nodes, and a set of communication links, Links, connecting them, each node $i$ has a certain amount of outgoing traffic of type $t$, denoted by OutgoingTraffic$_{i, t}$, and a certain amount of incoming traffic requirement for type $t$, denoted by IncomingTraffic$_{i, t}$. The cost of transmitting one data unit of traffic type $t$ from node $i$ to node $j$ is TransmissionCost$_{i, j, t}$. Each link $(i, j)$ has a bandwidth capacity for each traffic type $t$, represented as BandwidthCapacity$_{i, j, t}$, and a total combined bandwidth capacity for all traffic types, denoted as TotalLinkCapacity$_{i, j}$. The objective is to determine how much traffic of each type $t$ should be routed from each node $i$ to each node $j$ such that the total cost of routing the data is minimized, and the total amount of traffic routed from $i$ to $j$ does not exceed TotalLinkCapacity$_{i, j}$. For each traffic type $t$, the total traffic entering a node should meet its incoming traffic requirement, and the total outgoing traffic should not exceed its outgoing traffic availability. No link can be overloaded for any individual traffic type or in total. This network flow optimization problem determines the optimal routing strategy for multiple traffic types while respecting link capacity constraints and minimizing overall transmission cost.\\
\textbf{Input data:} \\
``Nodes'': [``A'', ``B''], ``Links'': [ [A, B] ], ``TrafficTypes'': [``Product1''], ``OutgoingTraffic'': [ [10], [0] ], ``IncomingTraffic'': [ [0],[10] ], ``TransmissionCost'': [ [ [1] ] ],
``BandwidthCapacity'': [ [ [10] ] ], ``TotalLinkCapacity'': [ [10] ]
\\
\textbf{Output:}\\
10\\
\textbf{Ground truth model: }\\
\textbf{Sets:}
\begin{itemize}
    \item $\text{Nodes}$: Set of all nodes (cities).
    \item $\text{Links} \subseteq \text{Nodes} \times \text{Nodes}$: Set of directed links between cities.
    \item $\text{TrafficTypes}$: Set of traffic types.
\end{itemize}

\textbf{Parameters:}
\begin{itemize}
    \item $\text{OutgoingTraffic}_{i,t}$: Amount of traffic type $t$ available at node $i$, $\forall i \in \text{Nodes},\ t \in \text{TrafficTypes}$.
    \item $\text{IncomingTraffic}_{i,t}$: Amount of traffic type $t$ required at node $i$, $\forall i \in \text{Nodes},\ t \in \text{TrafficTypes}$.
    \item $\text{TransmissionCost}_{i,j,t}$: Cost of transmitting one unit of traffic type $t$ from node $i$ to node $j$, $\forall (i,j) \in \text{Links},\ t \in \text{TrafficTypes}$.
    \item $\text{BandwidthCapacity}_{i,j,t}$: Maximum number of units of traffic type $t$ that can be transmitted over link $(i,j)$, $\forall (i,j) \in \text{Links},\ t \in \text{TrafficTypes}$.
    \item $\text{TotalLinkCapacity}_{i,j}$: Total capacity (across all traffic types) on link $(i,j)$, $\forall (i,j) \in \text{Links}$.
\end{itemize}

\textbf{Decision Variables:}
\begin{itemize}
    \item $\text{Route}_{i,j,t} \geq 0$: Number of data units of traffic type $t$ routed from node $i$ to node $j$, $\forall (i,j) \in \text{Links},\ t \in \text{TrafficTypes}$.
\end{itemize}

\textbf{Objective:} Minimize total transmission cost:
\[
\text{minimize} \sum_{(i,j) \in \text{Links}} \sum_{t \in \text{TrafficTypes}} \text{TransmissionCost}_{i,j,t} \cdot \text{Route}_{i,j,t}
\]

\textbf{Constraints:}
\begin{enumerate}
    \item \textbf{Flow conservation for each node and traffic type:}
    \[
    \hspace{-0.3in} \sum_{i : (i,k) \in \text{Links}} \text{Route}_{i,k,t} + \text{OutgoingTraffic}_{k,t} = \text{IncomingTraffic}_{k,t} + \sum_{j : (k,j) \in \text{Links}} \text{Route}_{k,j,t}, \forall k \in \text{Nodes},\ t \in \text{TrafficTypes}
    \]

    \item \textbf{Total link capacity constraint:}
    \[
    \sum_{t \in \text{TrafficTypes}} \text{Route}_{i,j,t} \leq \text{TotalLinkCapacity}_{i,j}, \quad \forall (i,j) \in \text{Links}
    \]

    \item \textbf{Bandwidth capacity per traffic type:}
    \[
    \text{Route}_{i,j,t} \leq \text{BandwidthCapacity}_{i,j,t}, \quad \forall (i,j) \in \text{Links},\ t \in \text{TrafficTypes}
    \]
\end{enumerate}
\end{tcolorbox}
\end{figure*}

\begin{figure*}
\begin{tcolorbox}[colback=green!5!white, colframe=green!50!black, title = Problem 7]
\textbf{Problem description:} \\
Consider a wireless resource allocation problem. Given a set of data services, Services (e.g., video streaming, VoIP, file transfer), and a set of transmission bands, Bands (e.g., frequency channels or time slots). Each service $s$ has a certain transmission efficiency Efficiency$_{s, b}$ in each band $b$ (representing data throughput per unit resource) and a profit per unit of data Profit$_{s}$ (e.g., revenue per MB delivered). Each band $b$ has a limited amount of available transmission time or bandwidth Available$_{b}$ per scheduling window (e.g., per minute or scheduling period). Additionally, there are lower and upper bounds on the amount of data that can be served per service, represented by MinimumDemand$_{s}$ and MaximumCapacity$_{s}$, respectively. The goal is to maximize total network profit while ensuring that the total transmission time used in each band does not exceed its available resources. The decision involves determining how many units of data to transmit for each service $s$ (e.g., how much bandwidth or time to allocate to each service). How to decide the data volume to be allocated to each service $s$ to maximize total profit, while respecting per-band capacity and service-level constraints. \\
\textbf{Input data:} \\
 ``Services'': [``P1'', ``P2''],
            ``Bands'': [``S1'', ``S2''],
            ``Efficiency'': [
                [2, 3],
                [3, 2]
            ],
            ``Profit'': [10, 20],
            ``MinimumDemand'': [1, 2],
            ``MaximumCapacity'': [5, 4],
            ``Available'': [10, 8]\\
\textbf{Output:}\\
110\\
\textbf{Ground truth model: }\\
\textbf{Sets:}
\begin{itemize}
    \item $\text{Services}$: A set of data services.
    \item $\text{Bands}$: A set of wireless transmission bands (e.g., frequency channels or time slots).
\end{itemize}

\textbf{Parameters:}
\begin{itemize}
    \item $\text{Efficiency}_{s,b}$: Data units (e.g., MB) per time unit in band $b$ for service $s$, $\forall s \in \text{Services},\ b \in \text{Bands}$.
    \item $\text{Available}_b$: Time units available per scheduling period in band $b$, $\forall b \in \text{Bands}$.
    \item $\text{Profit}_s$: Profit per ton for product $s$, $\forall s \in \text{Services}$.
    \item $\text{MinimumDemand}_s$: Minimum required data transmission for service $s$ during a scheduling period, $\forall s \in \text{Services}$.
    \item $\text{MaximumCapacity}_s$: Maximum allowable data transmission for service $s$ during a scheduling period, $\forall s \in \text{Services}$.
\end{itemize}

\textbf{Decision Variables:}
\begin{itemize}
    \item $\text{Allocation}_s$: Amount of data (e.g., MB) to be transmitted for service $s$, $\forall s \in \text{Services}$.
\end{itemize}

\textbf{Objective:}
\[
\text{maximize} \sum_{s \in \text{Services}} \text{Profit}_s \cdot \text{Allocation}_s
\]

\textbf{Constraints:}
\begin{enumerate}
    \item In each band $b$, the total transmission time used by all services may not exceed the available time:
    \[
    \sum_{s \in \text{Products}} \left(\frac{1}{\text{Efficiency}_{s,b}}\right) \cdot \text{Profit}_s \leq \text{Available}_b, \quad \forall b \in \text{Bands}
    \]

    \item Minimum guaranteed data delivery per service $s$ must be met:
    \[
    \text{MinimumDemand}_s \leq \text{Allocation}_s, \quad \forall s \in \text{Services}
    \]

    \item Maximum production may not be larger than the upper limit on tons of products $s$ sold in a week:
    \[
    \text{Allocation}_s \leq \text{MaximumCapacity}_s, \quad \forall s \in \text{Services}
    \]
\end{enumerate}
\end{tcolorbox}
\end{figure*}

\begin{figure*}
\begin{tcolorbox}[colback=green!5!white, colframe=green!50!black, title = Problem 8]
\textbf{Problem description:} \\
In wireless computer networks, we want to decide which devices or applications to schedule for transmission within a limited resource window, such as energy. Each device or application has an associated utility value (e.g., QoS satisfaction score) and consumes a specific amount of a limited resource (e.g., transmission energy).
The goal is to select the most valuable subset of devices or applications to serve in a scheduling cycle such that the total utility is maximized without exceeding the available network resource capacity (e.g., total energy budget or number of slots per frame). This formulation captures common scheduling and access control problems in wireless energy-constrained IoT environments and delay-sensitive applications.

\textbf{Input data:} \\
 ``DeviceUtility'': [60, 100, 120],
``DeviceCost'': [10, 20, 30],
``NetworkCapacity'': 50\\
\textbf{Output:}\\
220\\
\textbf{Ground truth model: }\\
\textbf{Set:}
\begin{itemize}
    \item $\text{Devices}$: Set of items.
\end{itemize}

\textbf{Parameters:}
\begin{itemize}
    \item $\text{DeviceUtility}_d$: Value of item $d$, $\forall d \in \text{Devices}$.
    \item $\text{DeviceCost}_d$: The energy consumption of item $d$, $\forall d \in \text{Devices}$.
    \item $\text{NetworkCapacity}$: Maximum capacity of the network.
\end{itemize}

\textbf{Decision Variables:}
\begin{itemize}
    \item $\text{ScheduleDevice}_d$: Item $d$ is placed in network, $\forall d \in \text{Devices}$.
\end{itemize}

\textbf{Objective:}
\[
\text{maximize} \sum_{d \in \text{Devices}} \text{ScheduleDevice}_d \cdot \text{DeviceUtility}_d
\]

\textbf{Constraint:}
\begin{enumerate}
    \item Constraint on total network resource energy usage:
    \[
    \sum_{d \in \text{Devices}} \text{ScheduleDevice}_d \cdot \text{DeviceCost}_d \leq \text{NetworkCapacity}
    \]
\end{enumerate}
\end{tcolorbox}
\end{figure*}

\begin{figure*}
\begin{tcolorbox}[colback=green!5!white, colframe=green!50!black, title = Problem 9]
\textbf{Problem description:} \\
This is a multi-commodity secure data routing problem in computer network security. Given a set of source servers or data centers, Sources, a set of target nodes or client devices, Targets, and a set of data types or services, DataTypes. Each source $i$ has a certain availability of each data type $p$, denoted as Availability$_{i, p}$, and each target node $j$ has a certain requirement for each data type $p$, denoted as Requirement$_{j, p}$. The cost of securely transmitting one unit of data type $p$ from source $i$ to target $j$ (e.g., due to encryption overhead, latency, or risk exposure) is TransmissionCost$_{i, j, p}$. The problem aims to minimize the total secure transmission cost of routing all data types from sources to targets across the network. It is constrained such that the total amount of each data type $p$ transmitted from each source $i$ must equal its availability. The total amount of each data type $p$ received at each target $j$ must equal its requirement. The total amount of all data types transmitted from source $i$ to target $j$ must not exceed a network bandwidth or security policy constraint, denoted as Capacity$_{i, j}$. How to determine the number of units of each data type $p$ to be securely transmitted from each source $i$ to each target $j$?

\textbf{Input data:} \\
 ``Availability'': [
                [20, 30],
                [40, 10]
            ],
``Requirement'': [
                [30, 30],
                [30, 10]
            ],
``Capacity'': [
                [35, 25],
                [20, 30]
            ],
``TransmissionCost'': [
                [
                    [2, 3],
                    [4, 1]
                ],
                [
                    [3, 2],
                    [2, 4]
                ]
            ]\\
\textbf{Output:}\\
235\\
\textbf{Ground truth model: }\\
\textbf{Sets:}
\begin{itemize}
    \item $\text{Sources}$: A set of source servers or data centers.
    \item $\text{Targets}$: A set of target nodes or client devices.
    \item $\text{DataTypes}$: A set of data types or services.
\end{itemize}

\textbf{Parameters:}
\begin{itemize}
    \item $\text{Availability}_{i,p}$: Amount of each data type $p$ available at source $i$, $\forall i \in \text{Sources}, p \in \text{DataTypes}$.
    \item $\text{Requirement}_{j,p}$: Amount of each data type $p$ required at target $j$, $\forall j \in \text{Targets}, p \in \text{DataTypes}$.
    \item $\text{Capacity}_{i,j}$: Maximum total amount of all data types that can be securely transmitted from source $i$ to target $j$, $\forall i \in \text{Sources}, j \in \text{Targets}$.
    \item $\text{TransmissionCost}_{i,j,p}$: Secure transmission cost per unit of data type $p$ from source $i$ to target $j$, $\forall i \in \text{Sources}, j \in \text{Targets}, p \in \text{DataTypes}$.
\end{itemize}

\textbf{Decision Variables:}
\begin{itemize}
    \item $\text{Flow}_{i,j,p}$: Units of each data type $p$ to be securely transmitted from source $i$ to target $j$, $\forall i \in \text{Sources}, j \in \text{Targets}, p \in \text{DataTypes}$.
\end{itemize}

\textbf{Objective:} Minimize the total cost of securely transmitting all data types:
\[
\min \sum_{i \in \text{Sources}} \sum_{j \in \text{Targets}} \sum_{p \in \text{DataTypes}} \text{TransmissionCost}_{i,j,p} \cdot \text{Flow}_{i,j,p}
\]

\textbf{Constraints:}
\begin{enumerate}
    \item The total amount of each data type $p$ transmitted from source $i$ equals its availability:
    \[
    \sum_{j \in \text{Targets}} \text{Flow}_{i,j,p} = \text{Availability}_{i,p}, \quad \forall i \in \text{Sources}, p \in \text{DataTypes}
    \]

    \item The total amount of each data type $p$ received at target $j$ equals its requirement:
    \[
    \sum_{i \in \text{Sources}} \text{Flow}_{i,j,p} = \text{Requirement}_{j,p}, \quad \forall j \in \text{Targets}, p \in \text{DataTypes}
    \]

    \item The total amount of all data types transmitted from source $i$ to target $j$ does not exceed the secure channel capacity:
    \[
    \sum_{p \in \text{DataTypes}} \text{Flow}_{i,j,p} \leq \text{Capacity}_{i,j}, \quad \forall i \in \text{Sources}, j \in \text{Targets}
    \]
\end{enumerate}
\end{tcolorbox}
\end{figure*}

\begin{figure*}
\begin{tcolorbox}[colback=green!5!white, colframe=green!50!black, title = Problem 10]
\textbf{Problem description:} \\
Consider a resource allocation problem in a wireless communication network. Given a set of base stations, BaseStations, and a set of user devices or communication tasks, Tasks. Each base station $i$ has a certain number of available transmission resource units (e.g., time slots, frequency blocks), denoted as Capacity$_{i}$, and each task $j$ requires a certain number of resource units, denoted as Requirement$_{j}$. The cost per unit of resource for serving task $j$ from base station $i$ (e.g., due to signal degradation, energy consumption, or interference) is Cost$_{i, j}$. Each base station $i$ can allocate up to a maximum limit, Limit$_{i, j}$, of its resources to task $j$ (e.g., due to quality-of-service or interference constraints). The problem aims to minimize the total transmission cost of serving all tasks from the base stations. It is constrained that the total number of resource units allocated from each base station $i$ must equal its capacity, and the total number of resource units allocated to each task $j$ must equal its requirement. How to decide the number of resource units to be allocated from each base station $i$ to each task $j$?\\
\textbf{Input data:} \\
 ``Capacity'': [8, 7],
``Requirement'': [5, 10],
``Cost'': [[10, 20], [15, 25]],
``Limit'': [[5, 6], [4, 7]]\\
\textbf{Output:}\\
285\\
\textbf{Ground truth model: }\\
\textbf{Sets:}
\begin{itemize}
    \item $\text{BaseStations}$: A set of base stations.
    \item $\text{Tasks}$: A set of tasks.
\end{itemize}

\textbf{Parameters:}
\begin{itemize}
    \item $\text{Capacity}_i$: Resource units available at base station $i$, $\forall i \in \text{BaseStations}$.
    \item $\text{Requirement}_j$: Resource units required by task $j$, $\forall j \in \text{Tasks}$.
    \item $\text{Cost}_{i,j}$: Cost per resource unit of assigning base station $i$ to task $j$, $\forall i \in \text{BaseStations}, j \in \text{Tasks}$.
    \item $\text{Limit}_{i,j}$: Maximum resource units assignable from base station $i$ to task $j$, $\forall i \in \text{BaseStations}, j \in \text{Tasks}$.
\end{itemize}

\textbf{Decision Variables:}
\begin{itemize}
    \item $\text{Assign}_{i,j}$: Resource units assigned from base station $i$ to task $j$, $\forall i \in \text{BaseStations}, j \in \text{Tasks}$.
\end{itemize}

\textbf{Objective:} Minimize total assignment cost:
\[
\text{minimize} \sum_{i \in \text{BaseStations}} \sum_{j \in \text{Tasks}} \text{Cost}_{i,j} \cdot \text{Assign}_{i,j}
\]

\textbf{Constraints:}
\begin{enumerate}
    \item Total assigned resource units from each base station equals its capacity:
    \[
    \sum_{j \in \text{Tasks}} \text{Assign}_{i,j} = \text{Capacity}_i, \quad \forall i \in \text{BaseStations}
    \]
    
    \item Total assigned resource units to each task equals its requirement:
    \[
    \sum_{i \in \text{BaseStations}} \text{Assign}_{i,j} = \text{Requirement}_j, \quad \forall j \in \text{Tasks}
    \]

    \item Assignment can not exceed the limit for each base station-task pair:
    \[
    \text{Assign}_{i,j} \leq \text{Limit}_{i,j}, \quad \forall i \in \text{BaseStations}, j \in \text{Tasks}
    \]
\end{enumerate}
\end{tcolorbox}
\end{figure*}

\end{document}